\documentclass{article}

\usepackage{microtype}
\usepackage{graphicx}
\usepackage{subfigure}
\usepackage{booktabs} 

\usepackage{hyperref}

\usepackage[accepted]{icml2023}
\usepackage{adjustbox}
\usepackage{makecell}

\usepackage{amsmath}
\usepackage{amssymb}
\usepackage{mathtools}
\usepackage{amsthm}

\usepackage[capitalize,noabbrev]{cleveref}

\theoremstyle{plain}
\newtheorem{theorem}{Theorem}[section]

\newtheorem{lemma}[theorem]{Lemma}
\newtheorem{claim}[theorem]{Claim}

\theoremstyle{definition}
\newtheorem{definition}[theorem]{Definition}

\theoremstyle{remark}

\usepackage[textsize=tiny]{todonotes}

\usepackage[inline]{enumitem}

\icmltitlerunning{AutoCoreset: An Automatic Practical Coreset Construction Framework}

\usepackage[utf8]{inputenc}
\usepackage{mathtools}
\usepackage{amsfonts}
\usepackage{xcolor}
\usepackage{enumitem}

\newcommand{\norm}[1]{\left\lVert#1\right\rVert}

\newcommand{\REAL}{\ensuremath{\mathbb{R}}}

\newcommand{\eps}{\varepsilon}

\newcommand{\infM}{\mathcal{M}^\ast(P,f)}
\newcommand{\subM}{\Tilde{\mathcal{M}}(P,f)}
\newcommand{\ac}{\textsc{AutoCoreset}}
\newcommand{\br}[1]{\left\lbrace #1 \right\rbrace}
\newcommand{\C}{\mathcal{I}}
\newcommand{\abs}[1]{\left| #1 \right|}

\newcommand{\term}[1]{\left( #1 \right)}

\newcommand{\alglinelabel}{%
  \addtocounter{ALC@line}{-1}
  \refstepcounter{ALC@line}
  \label
}

\DeclareMathOperator*{\argmin}{arg\,min}

\newcommand{\algorithmicreturn}{\textbf{return}}
\newcommand{\RETURN}{\item[\algorithmicreturn]}

\newcommand{\say}[1]{``#1"}

\begin{document}
\twocolumn[
\icmltitle{\textsc{\textit{AutoCoreset}}: An Automatic Practical Coreset Construction Framework}



\icmlsetsymbol{equal}{*}

\begin{icmlauthorlist}
\icmlauthor{\href{https://scholar.google.com/citations?user=6r72e-MAAAAJ&hl=en}{Alaa Maalouf$^1$}}{equal}
\icmlauthor{\href{https://scholar.google.com/citations?user=721xaz0AAAAJ&hl=en}{Murad Tukan$^2$}}{equal}
\icmlauthor{\href{https://scholar.google.com/citations?user=DTthB48AAAAJ&hl=en}{Vladimir Braverman$^3$}}{}
\icmlauthor{\href{https://scholar.google.com/citations?user=910z20QAAAAJ&hl=en}{Daniela Rus$^1$}}{}\\
\color{magenta}{$^{1}$Computer Science and Artificial Intelligence Lab (CSAIL),
Massachusetts Institute of Technology (MIT)}\\
{$^{2}$DataHeroes}\\
$^{3}$Department of Computer science, Rice University
 \vspace{2em}
\end{icmlauthorlist}

\icmlcorrespondingauthor{Alaa Maalouf}{alaam@mit.edu}

\icmlkeywords{Machine Learning, ICML}

\vskip 0.3in
]




\printAffiliationsAndNotice{\icmlEqualContribution} 

\begin{abstract}
A coreset is a tiny weighted subset of an input set, that closely resembles the loss function, with respect to a certain set of queries. Coresets became prevalent in machine learning as they have shown to be advantageous for many applications. 
While coreset research is an active research area, unfortunately, coresets are constructed in a problem-dependent manner, where for each problem, a new coreset construction algorithm is usually suggested, a process that may take time or may be hard for new researchers in the field.
Even the generic frameworks require additional (problem-dependent) computations or proofs to be done by the user. Besides, many problems do not have (provable) small coresets, limiting their applicability. 
To this end, we suggest an automatic practical framework for constructing coresets, which requires (only) the input data and the desired cost function from the user, without the need for any other task-related computation to be done by the user. 
To do so, we reduce the problem of approximating a loss function to an instance of vector summation approximation, where the vectors we aim to sum are loss vectors of a specific subset of the queries, such that we aim to approximate the image of the function on this subset. We show that while this set is limited, the coreset is quite general. 
An extensive experimental study on various machine learning applications is also conducted.
Finally, we provide a ``plug and play" style implementation, proposing a user-friendly system that can be easily used to apply coresets for many problems. Full open source code can be found at \href{https://github.com/alaamaalouf/AutoCoreset}{\text{https://github.com/alaamaalouf/AutoCoreset}}. We believe that these contributions enable future research and easier use and applications of coresets.
\end{abstract}


\begin{figure}[h]
    \subfigure[Logistic regression]{
    \includegraphics[width=0.24\textwidth]{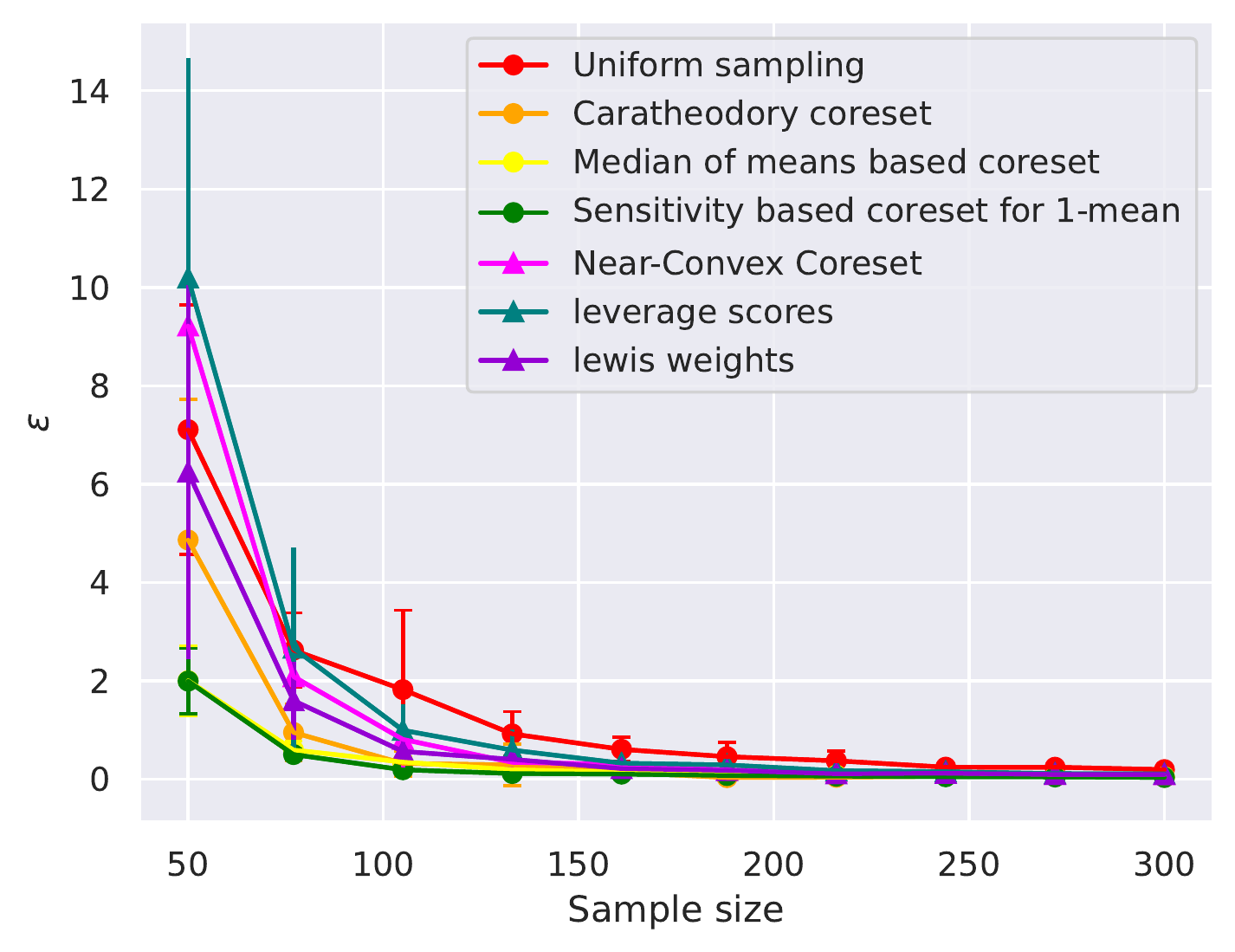}
    \includegraphics[width=0.24\textwidth]{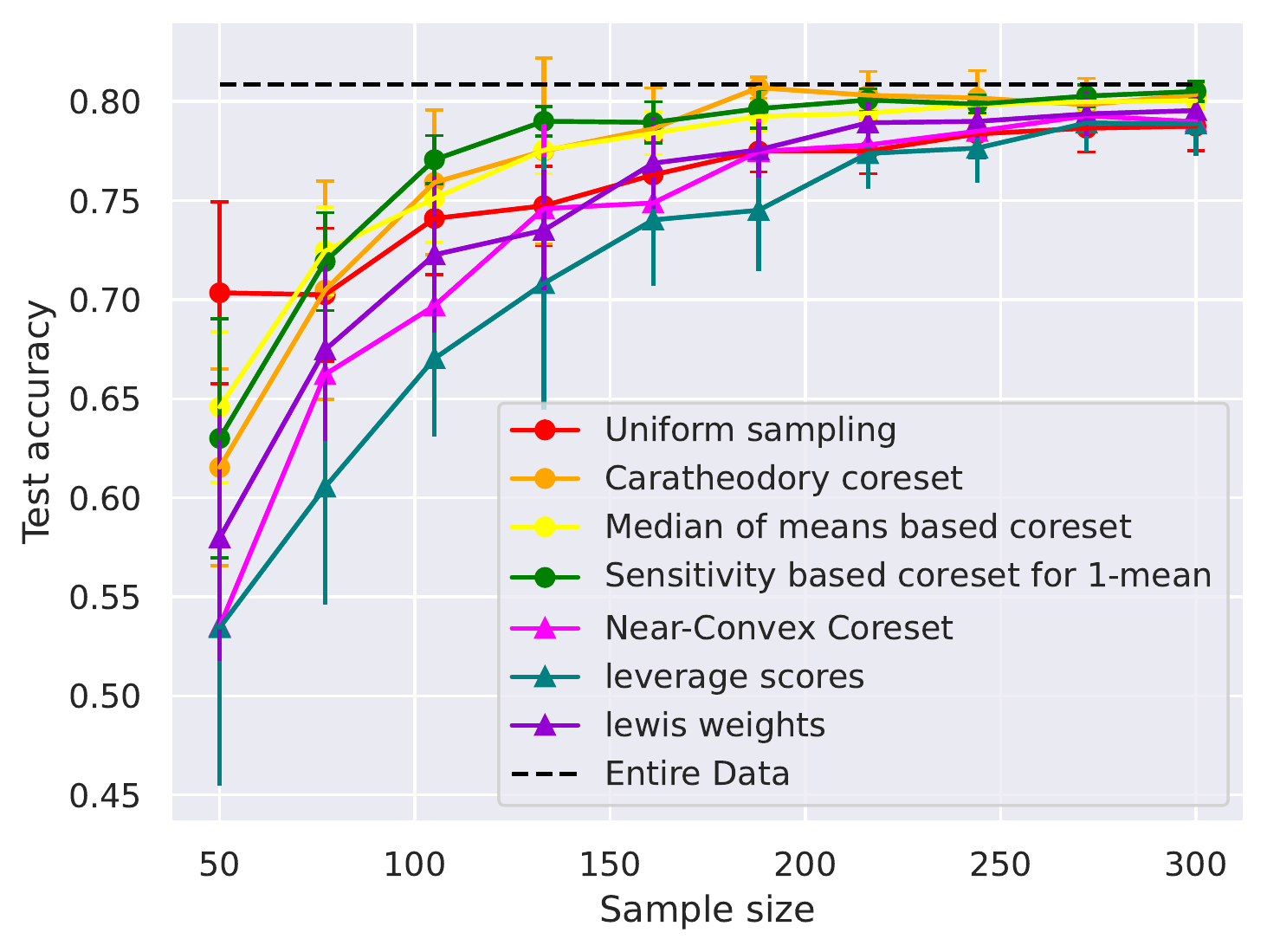}
    \label{fig:credit_logistic_regression}}
    \subfigure[SVMs]{
    \includegraphics[width=0.24\textwidth]{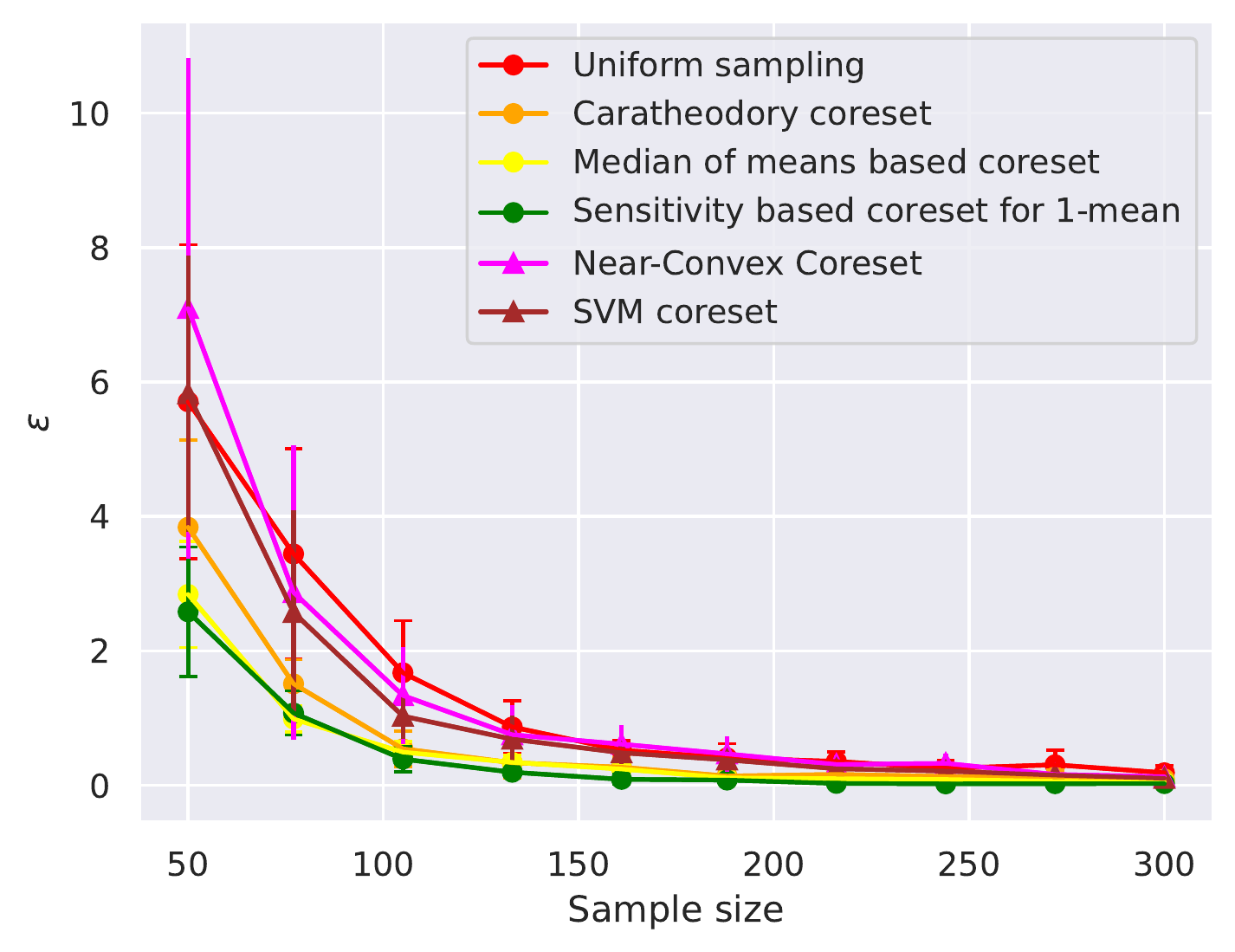}
    \includegraphics[width=0.24\textwidth]{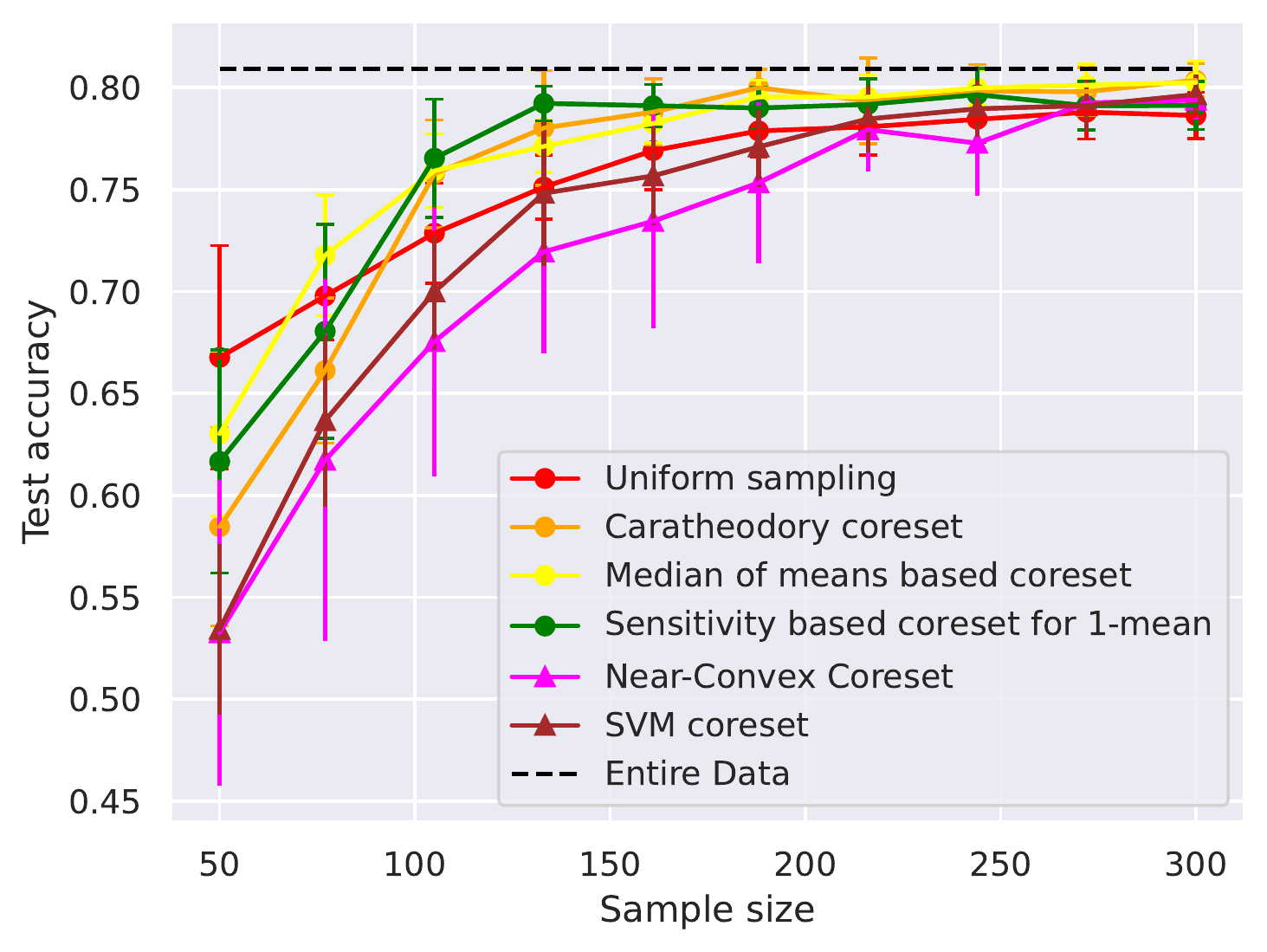}
     \label{fig:credit_svm}}
     \caption{Evaluation of AutoCoreset against other problem dependant coreset construction algorithms for SVM and Logistic regression (on the Dataset~\ref{dataset:credit}). AutoCoreset achieves a much smaller approximation error and a higher test accuracy for the same coreset size while being an automatic and problem-independent framework. Sensitivity-based coreset for 1-mean, Median of means-based coreset, and Caratheordory coreset are all variants of AutoCoreset.}
     \label{fig:credit}
 \end{figure}

\begin{figure*}[h]
    \centering
    \includegraphics[width=0.85\linewidth]{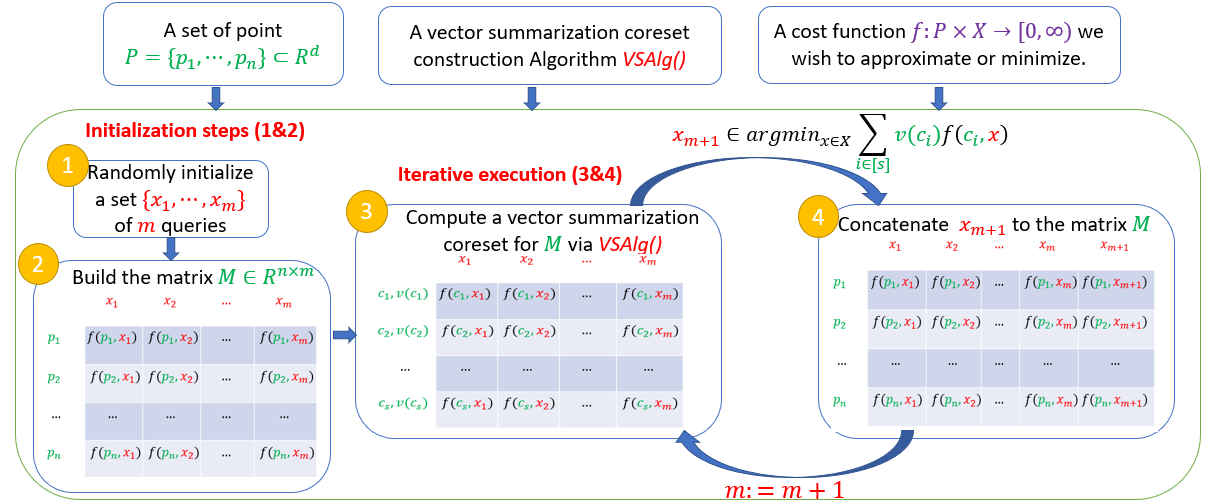}
    \caption{A flowchart illustrating our automatic coreset construction framework. Note that \textit{VSAlg()} can be any algorithm from Table~\ref{table:ourContrib}. }
    \label{fig:flow}
\end{figure*}
\section{Introduction and Motivation}
In many machine learning (ML) problems, the input is usually a set $P=\br{p_1,\cdots, p_n}$ of $n$ items, a (probably infinite) set of candidate solutions $\mathcal{X}$ called query set, and a loss function $f:P \times \mathcal{X} \to [0, \infty])$. 
The goal is to find a query (model, classifier) $x^*$ that minimizes the sum $\sum_{i=1}^n f(p_i,x)$ over every query $x\in \mathcal{X}$. Notably, many of these optimization/learning tasks are typically challenging to approximate when the input is very large. Furthermore, in the era of big data, we usually aim towards maintaining a solution for streaming and/or distributed input data, while consuming small memory. Finally, even well-known problems with a close optimal solution, such as ridge regression and other classes of convex optimization involving Cross-validation methods or hyper-parameter tuning methods, must analyze under many restrictions several queries for various subsets of data, leading to a drastic increase in the running time.


\textbf{Coresets.} A common approach to solve such issues is to use data summarization techniques, namely \emph{Coresets}, which got increasing attention over recent years~\cite{bachem2018scalable,pmlr-v84-bachem18a,buadoiu2008optimal,maalouf2022fast,balcan2013distributed, pmlr-v97-braverman19a,tukan2023provable,curtain2019coresets,jubran2020sets,feldman2014coresets,karnin2019discrepancy,tukan2022pruning,maalouf2021coresets,tukan2022new,tukan2022obstacle,tukan2023efficient}; see surveys in~\cite{feldman2020core,munteanu2018coresets,phillips2016coresets}, and introductions in~\cite{maalouf2021introduction,jubran2019introduction}. 
A coreset, informally, is a tiny weighted subset of the input set $P$, roughly approximating the loss of $P$ for every possible query $x\in \mathcal{X}$, up to a bound of $1\pm \eps$ factor ($0\leq\eps<1$). The size of the coreset is often independent or close to logarithmic in the amount of the input points $n$, but polynomial in $1/\eps$.
Coresets are useful in ML as they significantly increase the efficiency of ML solvers. Specifically, employing conventional methods on the constructed coresets should approximate the optimal solution on the entire dataset, in orders of magnitude less expensive time and memory. Furthermore, by repeatedly running existing heuristics on the coreset in the time it takes to run them once on the original (large) dataset, heuristics that are already quick can be more accurate. Additionally, coresets can be maintained for distributed and streaming data.


\textbf{So what's the problem?} Obtaining non-trivial theoretical guarantees is frequently impossible in many contemporary machine learning problems due to either the target model being highly complex or since every input element $p\in P$ is significant in the sense of high sensitivity; see~\cite{NearConvex}. Hence, generating a coreset becomes a highly challenging process, and the corresponding theoretical analysis occasionally falls short of recommending such approximations. As a result, designing a new coreset and demonstrating its accuracy for a new ML problem might take years, even for simple ones. 

Another crucial issue with current theoretical frameworks is their lack of universality. Even the most general frameworks (e.g.,~\cite{feldman2011unified, Langberg2010universal} replace the problem of generating a coreset for an input set $P$ of $n$ points with $n$ new optimization problems (one problem for each of the $n$ input points $p\in P$) known as sensitivity bounding. Solving these may be more difficult than solving the original problem, where for every $p\in P$ we are required to bound its own sensitivity defined as $s(p)=\sup_{x\in \mathcal{X}}\frac{f(p,x)}{\sum_{q\in P}f(q,x)}$. As a result, distinct approximation strategies are often adapted to each task.
Hence, the main disadvantage of such frameworks is that researchers provide papers solely for bounding the sensitivities with respect to a certain problem or a family of functions~\cite{NearConvex,maalouf2020tight}, limiting the spread of coresets, as non-expert won't be able to suggest coresets for their desired task. These problems raise the following questions:

\textbf{Is it possible to design an automatic and practically robust coreset construction framework (for any desired cost function and input dataset) that does not need sensitivity calculation or any other problem-dependent computation by the user?  Can we provide some provable guarantees with respect to this framework?}
\subsection{Vision}
 
\paragraph{Goal.} Our goal is to provide a single algorithm that only receives the loss function we wish to compute a coreset for and the input dataset, then, it practically outputs a good coreset for the input dataset with respect to the given loss. This algorithm should be generic, efficient, and work practically well for many problems.

\paragraph{motivation.} The main motivation behind this goal is (1) to increase the spread and use of coreset to a larger community that is not limited to coreset researchers or pioneers. (2) Additionally, to ease the use of coresets for many other applications that may be out of the scope of the coreset literature, and finally, to (3) easily apply coresets for new problems that do not have provable coresets. Theoretically speaking, it is indeed very hard to provide a ``theoretical strong coreset’’ to any problem – for example, there exist lower bounds on the coreset sizes for different problems~\cite{munteanu2018logistic,tukan2020coresets}. Thus we aimed at a practical result while providing weaker theoretical guarantees, with an extensive experimental study.

\subsection{Our contribution}
In this paper, we provide a coreset construction mechanism that answers both questions. Specifically:
\begin{enumerate}[label=(\roman*)]
  \item The first automatic practical coreset construction system that only needs to receive the loss function associated with the problem. Our coreset does not require any computation to be done by the user, not mathematical nor technical (without the need for sensitivities or any other task-related computation by the user). To the best of our knowledge, this is the first paper to suggest a \textbf{plug-and-play} style framework/compiler for coreset construction. We also provide a theoretical justification for using our framework. 
  
 \item An extensive empirical study on real-world datasets for various ML solvers of Scikit-Learn~\cite{scikit-learn}, including k-means, logistic regression, linear regression, and support vector machines (SVM), showing the effectiveness of our proposed system.
 
  \item \textbf{AutoCoreset:} An open-source code implementation of our algorithm for reproducing our results and future research. For simplicity and ease of access, to obtain a coreset, the user only needs to plug in his desired loss function and the data into our system. We believe this system will popularize and expose the use of coresets to other scientific fields. 
\end{enumerate}


\section{Setup Details}
\label{sec:setup_details}
Given a set $P = \br{p_1\cdots,p_n} \subseteq \REAL^d$ of $n$ points\footnote{if $P$ is a set of labeled items, then $P = \br{p_i=\term{p'_i, y_i} \middle| p'_i \in \REAL^{d-1}, y_i \in\REAL}_{i=1}^n$} and a loss function $f : P \times \mathcal{X} \to [0, \infty)$ where $\mathcal{X}$ is a (possibly infinite) set of queries. 
In this paper, we develop an automatic coreset construction framework for any problem involving cost functions of the form $\sum_{p \in P} f(p, x)$, here $x\in \mathcal{X}$. 
Formally, we wish to find a small subset $\C \subseteq  [n]$ and a weight function $v: \C \to [0,\infty)$ such that
$
\max_{x \in \mathcal{X}}\frac{\sum\limits_{j \in \C} v(j) f\term{p_j,x}}{\sum\limits_{i = 1}^n f\term{p_i,x}} \in 1 + O(\eps),
$ 
for some small $\eps \geq 0$.


\begin{figure}[t]
    \centering
    \includegraphics[width=\linewidth]{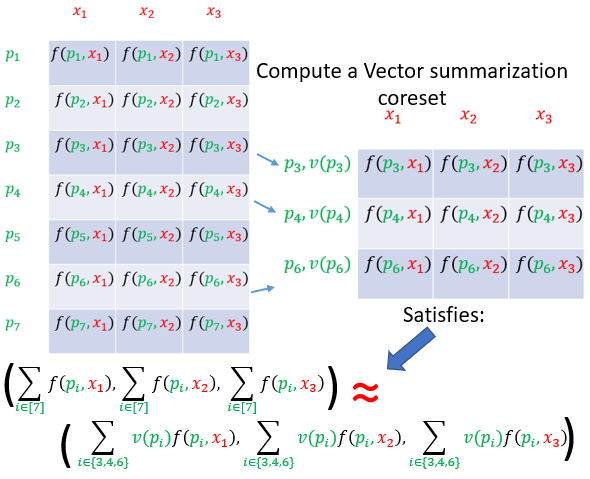}
    \caption{Illustration of a vector summarization coreset for an input matrix of $7$ rows and $3$ columns which represent the loss function concerning a set $P$ of $7$ input points, and set of queries $x_1,x_2,x_3$.}
    \label{fig:vector_sum}
\end{figure}

\subsection{Preliminaries}
We now give our notations and used Definition.

\textbf{Notations.}  For a pair of integers $n,m>0$, we denote by $[n]$ the set $\br{1,\cdots,n}$, and by $\REAL^{n\times m}$ the set of every possible $n \times m$ real matrix. For a matrix $M\in \REAL^{n\times m}$ and a pair of integers $i\in [n],j\in [m]$, we use $M_{i,\ast}$ to denote its $i$th row (vector), $M_{\ast,j}$ to denote its $j$th column, and $M_{i,j}$ to denote the entry in the $i$th row and $j$th columns.

In what follows, we define a crucial component on which our system relies, namely, \emph{vector summarization coreset}.

\begin{definition}[Vector summarization coreset]
\label{def:VSCoreset}
Let $M \in \REAL^{n\times m}$, $\C \subseteq [n]$, $v: \C \to [0, \infty)$ be a weight function, and let $\eps > 0$. The tuple $\term{\C,v}$ is an vector summarization $\eps$-coreset for $M$ if
$
\norm{\sum_{i \in [n]}  M_{i,\ast} - \sum_{j \in \C} v(j) M_{j,\ast}}_2^2 \leq \eps.$
\end{definition}

Many papers suggested different algorithms for computing such coresets; in Table~\ref{table:ourContrib} summarizes some of these results, as we can use them all of them in our method. 

\section{\emph{AutoCoreset}}
A coreset aims to approximate the probability distribution induced upon the input data by the cost function. Hence, in order to approximate a given cost function, the coreset must contain points that can result in an approximated distribution to that of the full data.

\textbf{Key idea. }
Loosely speaking, assume that for a given cost function $f$ and a set $P=\br{p_1,\cdots,p_n} \subset\REAL^d$, we access an infinite matrix $\infM$ where the rows correspond to the $n$ points of $P$, and each column corresponds to a query point from the infinite set of queries $\mathcal{X}$. Specifically, each row $i\in [n]$ is of infinite length representing the loss of each point $p_i \in P$ with respect to the infinite set of queries $\mathcal{X}$. A coreset in this context means finding a subset of the rows $\mathcal{I} \subseteq [n]$, and a weight function $v: \mathcal{I}  \to [0, \infty]$, that satisfies the vector summarization coreset guarantee (see Definition~\ref{def:VSCoreset}), i.e., 
\begin{equation}
\label{eq:dream}
\norm{\sum_{i \in [n]} {\infM}_{i,\ast} - \sum_{j \in \mathcal{I}} v(j){\infM}_{j,\ast}}_2^2 \leq \eps.
\end{equation}

From such a coreset $\mathcal{I} \subseteq [n]$, the cost function can be approximated, since for every query $x\in \mathcal{X}$ (column in the matrix $\infM$), the weighted sum of losses over the coreset $\mathcal{I}$ approximate the original sum of losses of the whole data.  
While such a concept is admirable, having an access to such immense data is rather imaginative. 

Recently~\cite{maaloufsine2022coresets} showed that for an input set of points $P$, and query space $X$ that is defined as a family of sine wave functions, a coreset can be constructed. Specifically, it was shown that if the coreset approximates the loss of every query in a smaller set of queries on the input data, then it will also approximate the losses of the whole set of queries (sine waves). Thus, indeed, the sine wave that fits best the coreset approximates the sine wave that best fits the entire data. 
Inspired by such a result, we aim towards constructing a sub-matrix $\subM$ of $\infM$ ($\subM$ contain a subset of the columns of $\infM$; see Figure~\ref{fig:vector_sum} for illustration) such that constructing the coreset on $\subM$ (a weighted subset of the rows of $\subM$) will also yield a similar coreset to that of~\eqref{eq:dream} on the $\infM$. 
But, how to build the sub-matrix $\subM$? how to choose the query set corresponding to the columns of $\subM$?





\begin{table}[ht]
\centering
\caption{{Summary of known vector summarization coresets and their properties.} 
}
\label{table:ourContrib}
\begin{adjustbox}{width=1\linewidth }
\small
\begin{tabular}{ | c | c | c | c | c |}
\hline \textbf{Method} &
 \textbf{\makecell{Probability\\of failure}} & \textbf{\makecell{Approximation\\error}} & \textbf{Coreset size $|\C|$} & \textbf{\makecell{Construction time}} \\
\hline
\makecell{Caratheodory\\\cite{maalouf2019fast}\\\cite{caratheodory1907variabilitatsbereich}} & 0 & 0 & $m+1$ & $O(\min\{nm+\log^4(m),m^2n^2,nm^3\})$ \\
\hline
\makecell{Frank-Wolfe\\ \cite{feldman2017coresets}\\\cite{clarkson2010coresets}} & 0 & $\eps$ & $O(1/\eps)$ & $O(\min\{nd/\eps\})$ \\
\hline
\makecell{Median of means \\ tournament\\\cite{minsker2015geometric} }& $\delta$ & $\eps$ & $O(1/\eps)$ & $O(m\log^2(1/\delta) + m\log(1/\delta)/\eps)$ \\
\hline
\makecell{Sensitivity sampling\\
\cite{feldman2011unified}} & $\delta$ & $\eps$ & $O(\frac{1}{\eps}(m+\log{\frac{1}{\delta}}))$ & $O(nm)$ \\
\hline
Uniform sampling & $\delta$ & $\eps$ & $O(\frac{1}{\eps\delta})$ & $O(1)$ \\ 
\hline

\end{tabular}
\end{adjustbox}
\end{table}

\begin{algorithm}[!t]
   \caption{$\ac\term{P, f, \tau, m, \zeta}$}
   \label{alg:main}
\begin{algorithmic}[1]
   \INPUT set of $n$ points $P$, a loss function $f$, a coreset size $\tau$, number of initial models $m$, and an stopping criterion $\zeta$
   \OUTPUT A coreset $\term{\C,v}$ such that
   \STATE $\subM \gets \overset{\to}{0}_{n \times m}$ \alglinelabel{line:settingM}
   \FOR{each $i \in [m]$ \alglinelabel{line:pre_for_init_M}}
        \STATE $x_i \gets $ a randomized approximated solution involving $P$ and $f$\alglinelabel{line:RandomSols}
        \FOR{every $j \in [n]$ \alglinelabel{line:for_init_M}}
            \STATE ${\subM}_{j, i} \gets f\term{p_j, x_i}$ \alglinelabel{line:init_M}
        \ENDFOR \alglinelabel{line:ending_for_init_M}
   \ENDFOR \alglinelabel{line:ending_pre_for_init_M}
   \REPEAT \alglinelabel{line:repeat}
        \STATE $\term{\C, v} \gets$ coreset of $m$ indices for vector summarization problem involving $\subM$ \COMMENT{See Definition~\ref{def:VSCoreset}} \alglinelabel{line:generate_VSCoreset}
        \STATE $x^\ast \gets \argmin_{x \in \mathcal{X}} \sum\limits_{i \in \C} v\term{i} f\term{p_i, x}$ \alglinelabel{line:solve_on_coreset}
        \STATE $\subM \gets \left[\subM \middle| \overset{\to}{0}_{n}\right]$ \alglinelabel{line:concat_zero_column}
        \FOR{every $i \in [n]$}
            \STATE ${\subM}_{i, m + 1} \gets f\term{p_i, x_C}$ \alglinelabel{line:add_column_to_M}
        \ENDFOR\alglinelabel{line:end_for_concat_new_column_M}
        \STATE $m \gets m + 1$ \alglinelabel{line:add_column_counter}
   \UNTIL{$\zeta$ is satisfied}
   \RETURN $\term{\C, v}$ \alglinelabel{line:return_Coreset}
\end{algorithmic}
\end{algorithm}

\subsection{A deeper look into \emph{AutoCoreset}}
\label{sec:deeper}

We now give and explain our algorithm $\ac$ (see Algorithm~\ref{alg:main}), which aims to provide a parasitical coreset with similar guarantees.






\textbf{Into the forging of our coresets.} Let $m > 1$ be an integer. 
First, a matrix $\subM$ is generated to contain $n \times m$ zero entries, followed by generating a set $\mathcal{X}^\prime=\br{{x}_1,\ldots,{x}_m }$ of $m$ approximated solutions with respect to $\min\limits_{x \in \mathcal{X}} \sum\limits_{i=1}^n f(p_i,x)$ as depicted at Lines~\ref{line:settingM}--\ref{line:ending_pre_for_init_M}. If no such approximated solution exists, then the initialization may be also completely random. The (sub)matrix $\subM$ is now initialized, where for every $i\in [n]$, and $j\in [m]$, the entry $\subM_{i,j}$ in the $i$th row and $j$th column is equal to $f\term{p_i,{x}_j}$. 
While the properties associated with generated solutions at Line~\ref{line:RandomSols} hold with some probability, our framework is always guaranteed practically to generate a good coreset. This is due to the fact that these solutions are merely used as an initialization mechanism. 

From this point on, a loop is invoked. First, using the current state of $\subM$, a vector summarization coreset $\term{\C,v}$ (see Definition~\ref{def:VSCoreset}) is generated with respect to the rows of $\subM$. 

A coherent claim of our system is that any vector summarization coreset $\C$ for the rows of $\subM$, is directly mapped to coreset for $P$ (using the same set of indexes and the same weight function) with respect to the query set $\mathcal{X}^\prime \subset \mathcal{X}$ and the function $f$, where $\mathcal{X}^\prime$ is the set of all queries that brought about the columns of $\subM$. More preciously, 
$
\max_{x \in \mathcal{X}^\prime}\frac{\sum\limits_{j \in \C} v(j) f\term{p_j,x}}{\sum\limits_{i = 1}^n f\term{p_i,x}} \in 1 + O(\eps); 
$
see Lemma~\ref{lem:VSC2FC}.

Since the computed vector summarization coreset $\C$ is also a coreset with respect to $f, P$, and $\mathcal{X}^\prime$, we can optimize $f$ over the small coreset $\C$ to obtain a new query $x^*\in \mathcal{X}$ that gives an approximated solution to the full data (see Line~\ref{line:solve_on_coreset}). We then apply the loss $f$ function and the new solution $x^*$ on $p_1,\cdots, p_n$ to obtain the vector of losses $l = (f(p_1,x^*),\cdots,f(p_n,x^*))^T$, and concatenate such vector of loss values to $\subM$ as its last column. This aids in expanding the exposure of generated coreset to a wider spectrum of queries, leading towards a \emph{strong coreset}. Observe that in the next iteration, when we compute a new coreset for the given set of queries, the coreset will approximate all of the previous ones (set of queries) and the new computed query/solution $x^*$. 

This procedure is repeated until some stopping criterion $\zeta$ is invoked -- we provide more details on the used $\zeta$ in Section~\ref{sec:exp}. We refer the reader to Lines~\ref{line:solve_on_coreset}--\ref{line:add_column_counter}. Note that if we were able to run the above procedure infinitely while ensuring that at each iteration a new solution is computed, $\infM$ would have been generated, resulting in the \say{strong coreset} this system is leaning towards. To better grasp the idea of the framework, we provide a flowchart illustration at Figure~\ref{fig:flow}.

\textbf{The parameters $\tau, m, \zeta$.} Our Algorithm initializes its matrix $\subM$ with respect to the losses of $m>1$ different queries, and outputs a coreset of size $\tau>1$, hence, the larger the $m$ and $\tau$ the better the approximation, but the slower the time; See section~\ref{sec:exp} for more details. Regarding $\zeta$, it is the used stopping criterion, we provide full details regarding the used $\zeta$ in Section~\ref{sec:exp}.


\subsection{Weaker coresets are fine too}
Our \emph{AutoCoreset} system, while ambitiously aims towards holding a grasp over $\infM$, it finds a weaker version of the \say{strong coresets}. Specifically speaking, it finds a coreset that attains approximation guarantees with respect to a subset of the query set $\mathcal{X}$. Theoretically speaking, the following lemma summarizes one aspect of the theoretical properties guaranteed by \emph{AutoCoreset}.

\begin{lemma}[Vector summarization coreset $\to$ \say{a weak coreset for any loss}]
\label{lem:VSC2FC}
Let $P=\br{p_1,\cdots,p_n}\subseteq \REAL^{d}$ be a set of $n$ points as defined in Section~\ref{sec:deeper}, $\mathcal{X}^\prime \subset \mathcal{X}$ be a set of queries, $f : P \times \mathcal{X} \to [0, \infty)$ be a loss function, and let  $\subM \in \REAL^{n \times \abs{\mathcal{X}^\prime}}$ be the loss matrix defined with respect to $P, f, \mathcal{X}^\prime$ as in Algorithm~\ref{alg:main}. Let $\tau \geq 1$ be an integer, and let $\term{\C, v}$ be a $\eps$-vector summarization coreset concerning $\subM$ of size $\abs{\C} = \tau$.
Then, for every $x \in \mathcal{X}^\prime$,
\[
\abs{\sum_{i \in [n]} f\term{p_i,x} - \sum_{j \in \C} v(j) f\term{p_j,x}}^2 \leq \eps. 
\]
\end{lemma}

\textbf{Implications of Lemma~\ref{lem:VSC2FC}.} \emph{AutoCoreset} guarantees theoretically that for a finite set of queries $\mathcal{X}^\prime$, a coreset can be constructed supporting $\mathcal{X}^\prime$. A key advantage here would be the ability to represent any query $x$ such that its loss vector $(f(p_1,x),\cdots,f(p_n,x))$ lies inside the \say{convex hull} of the loss vectors of the query set $\mathcal{X}^\prime$. Luckily, such a trait is supported by our system. Specifically speaking, for any query such that its corresponding loss vector $\ell$ with respect to $f$ and $P$ can be formulated as a convex combination of the columns of $\subM$, then a vector summarization coreset for the rows of $\subM$ is also a vector summarization to the rows of concatenating $\subM$ and the column vector $\ell$. In what follows, we give the theoretical justification for the above claim.

\begin{claim}[Weak Coreset with hidden abilities]
\label{clm:hidden_talent}
Let $P=\br{p_1,\cdots,p_n}\subseteq \REAL^d $ be a set of $n$ points as in Section~\ref{sec:deeper}, $f$ be a loss function supported by \emph{AutoCoreset}, and let $m, \tau, \zeta$ be the defined number of initial solutions, sample size, and stopping criterion, respectively. Let $z \geq m$, $\term{\C,v}$ be the output of a call to $\ac\term{P, f, \tau, m, \zeta}$, and let $\subM \in \REAL^{d \times z}$ be the matrix of losses that was constructed throughout the running time of $\ac$; see Lines~\ref{line:settingM},~\ref{line:init_M},~\ref{line:concat_zero_column}~\ref{line:add_column_to_M} at Algorithm~\ref{alg:main}. 
Then for any weight function $\alpha : [z] \to [0,1]$ where $\sum\limits_{i =1}^z \alpha\term{i} = 1$, and any $x \in \mathcal{X}$ satisfying that for every $i \in [n]$, $f\term{p_i,x} = \sum\limits_{k=1}^z \alpha\term{k} \subM_{i, k}$ , we have
\[
\abs{\sum\limits_{i =1}^n f\term{p_i, x} - \sum\limits_{j \in \C} v(j) f\term{p_j, x}}^2 \leq \eps,
\]
where $\eps \geq 0$ is the approximation factor associated with generating a vector summarization coreset of $m$ points.
\end{claim}


\textbf{The best of both worlds.} Claim~\ref{clm:hidden_talent} states that even if it seems that our generated coreset only supports a handful of queries from $\mathcal{X}$, our coreset basically supports many more queries. The highlight of such a claim is that if the optimal solution for the objective function involves $f$ and $P$, then our coreset becomes stronger in the sense of ensuring better quality even during the training/optimization process which involves both $f$ and $P$. Such a claim is usually targeted via \say{Strong coresets} and mainly by \say{Weak coresets}. \emph{AutoCoreset} ensures a coreset that resides on the spectrum involving these coresets at its ends, i.e., generating a coreset from the best of both worlds -- a coreset supporting the optimal solution that the user is aiming to solve using accelerated training via coresets while maintaining the provable approximation guarantees of strong coresets to some extent. 








\section{Size, Space, and Time Analysis}

\textbf{Time complexity. }
Let $\texttt{VAlg}$ be the vector summarization algorithm used at Line 9 of Algorithm~\ref{alg:main} (pick one from Table~\ref{table:ourContrib}). Let $\varepsilon, \delta \in (0,1)$ be the desired vector summarization approximation error, and probability of failure, respectively. Now denote by
\begin{itemize}
    \item  $T(n, i, \varepsilon, \delta )$: the running time of $\texttt{VAlg}$ on a matrix of $n$ rows and $i$ columns with respect to $\varepsilon$ and $\delta$.
    \item  $S(n, i, \varepsilon, \delta )$: the size of the coreset computed by VAlg on a matrix of $n$ rows and $i$ columns with respect to $\varepsilon$ and $\delta$. 
    \item $T_{sol}(n,d)$: the time required to compute a solution vector $x^*$ for $n$ points in the $d$ dimensional space with respect to the problem at hand (e.g., the time required to compute the solution of linear regression is $O(nd^2)$).
    \item $T_{cost}(n,d)$: the time required to calculate the cost for $n$ points in the $d$ dimensional space on a single query with respect to the problem at hand (e.g., the time required to compute the cost of linear regression for $n$ points in the $d$ dimensional space given a solution vector $x$ is $O(nd)$.
    “t”: be the number of iterations of the algorithm.
\end{itemize}
  
At each iteration “i”, Algorithm~\ref{alg:main} 
\begin{enumerate}
    \item  applies $\texttt{VAlg}$ on a matrix of $n$ rows and $i$ columns to obtain a coreset of size $S(n, i, \varepsilon, \delta )$. This step requires $T(n, i, \varepsilon, \delta )$ time.
    \item Solves the problem on the coreset to obtain a new solution $x^*$. Requiring $T_{sol}(S(n, i, \varepsilon, \delta ) , d)$ time.
    \item  Calculates the cost of the $n$ points with respect to $x^*$. Requiring $T_{cost}(n,d)$ time
\end{enumerate}
Thus, for a single step $i$ the running time is
    $T(n, i, \varepsilon, \delta ) + T_{sol}(S(n, i, \varepsilon, \delta ), d) + T_{cost}(n,d)$. 
Summing for $t$ iterations:
    $$\sum_{i=1}^t (T(n, i, \varepsilon, \delta ) + T_{sol}(S(n, i, \varepsilon, \delta ) , d)) +t T_{cost}(n,d).$$

For example, in Linear regression and using the Sensitivity sampling as \textit{VAlg}, an immediate bound for the running time is $O(t (nt + (t/\varepsilon+log(1/\delta)/\varepsilon) d^2 + nd))$.

\textbf{Space complexity. } First, note that the input data and the matrix of losses take $O(n(d+t))$ where $t$ here denotes the number of iterations our coreset generation has taken. Recall the definitions of $\textit{VAlg}, \varepsilon, \delta$ and $S(n, i, \varepsilon, \delta )$. We now denote by
\begin{itemize}
    \item  $Mem(\textit{VAlg}, \varepsilon, \delta, i)$ the amount of space needed by Valg to generate an $\varepsilon$-coreset with a success probability of at least $ 1 - \delta$.
    \item  $Mem_{sol}(n,d)$ the space required to compute a solution vector $x^*$ for $n$ points in the $d$ dimensional space with respect to the problem at hand (e.g., the space required to compute the solution of SVM is $O(n^2 + d)$.
\end{itemize}
   
The total space complexity is thus bounded by 
$O(n(d+t) + \max_{i \in [t]} (Mem(\textit{VAlg}, \varepsilon, \delta, i) + Mem_{sol}(S(n, i, \varepsilon, \delta )),d).$

For example for SVM and using the Sensitivity sampling vector summarization, an immediate bound for the space complexity is   $O(n(d+t) + (1/\varepsilon(t + \log(1/\delta)))^2)$.

\textbf{Coreset size.}  The size of the constructed coreset is equal to the used vector summarization coreset size (See Table~\ref{table:ourContrib}), and it depends on the approximation error $\varepsilon$, the probability of failure $\delta$ we wish to have, and the final number of approximated queries – columns of the query matrix.

In short – let $\varepsilon$ be the desired approximation error and let $\delta$ be the probability of failure. Let $t$ be the number of iterations required Algorithm~\ref{alg:main}. Denote by $S(n, i, \varepsilon, \delta )$ the size of the set computed by the used vector summarization algorithm on a matrix of $n$ rows and $i$ columns with respect to $\varepsilon$ and $\delta$ (see Table~\ref{table:ourContrib} for examples). Then, the size of the coreset is $S(n, t, \varepsilon, \delta )$.

For example, using the Sensitivity sampling method (as the used vector summarization coreset), to approximate the currently given $t$ queries after $t$ iterations, with $\varepsilon$ approximation error, and $\delta$ probability of failure, we get a coreset of size $O(t/\varepsilon + log(1/\delta)/\varepsilon)$.

\textbf{From additive to multiplicative approximation error. } Algorithm~\ref{alg:main} can immediately be modified to compute a coreset that yields a multiplicative approximation as follows. Given the set $P$, the current set of queries $\mathcal{X}^\prime$, and the loss $f$, define a new function $g(p,x):= \frac{f(p,x)}{\sqrt{\sum_{p\in P}f(p,x)}}$ for every pair of a query $x \in \mathcal{X}^\prime$ and input data $p\in P$. 

Now build the corresponding matrix $\tilde{\mathcal{M}}(P,g)$ (as done in Algorithm~\ref{alg:main} for $f(p,x)$) instead of $\tilde{\mathcal{M}}(P,f)$, and run the exact same vector summarization coreset algorithm on it. Then, by Lemma~\ref{lem:VSC2FC}, for every $x \in \mathcal{X}^\prime$,
$
\abs{\sum_{i \in [n]} g\term{p_i,x} - \sum_{j \in \C} v(j) g\term{p_j,x}}^2 \leq \eps,$
 and by the definition of $g$ we get  that the result is a multiplicative coreset for the given set of queries as for every $x \in \mathcal{X}^\prime$
\[
\abs{\sum_{i \in [n]} f\term{p_i,x} - \sum_{j \in \C} v(j) f\term{p_j,x}}^2 \leq \eps \sum_{p\in P}f(p,x).
\]

\section{Experimental Study}
\label{sec:exp}

 \begin{figure*}[!t]
    \subfigure[Logistic regression]{
    \includegraphics[width=0.24\textwidth]{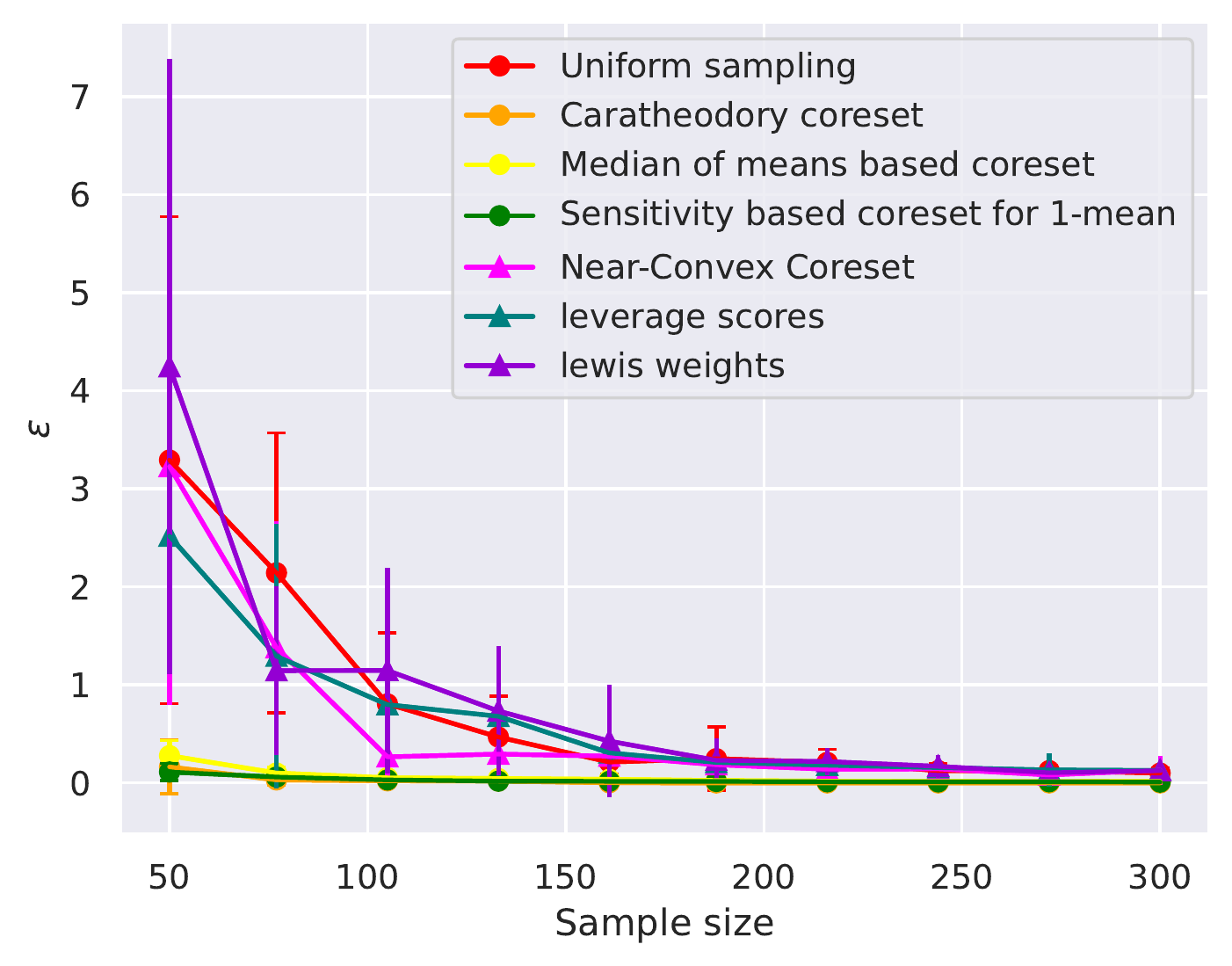}
    \includegraphics[width=0.24\textwidth]{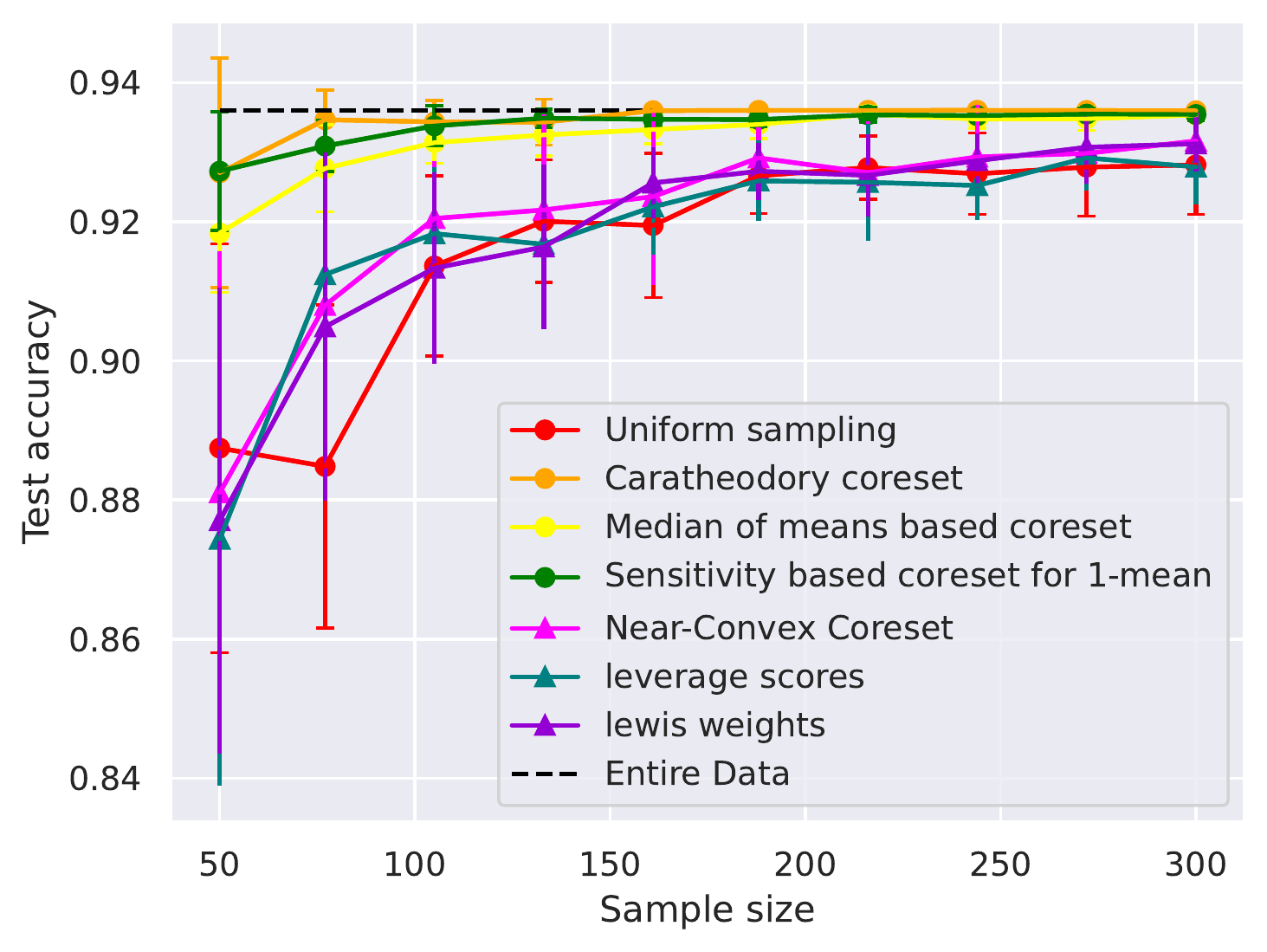}
    \label{fig:cod-rna_logistic_regression}}
    \subfigure[SVMs]{
    \includegraphics[width=0.24\textwidth]{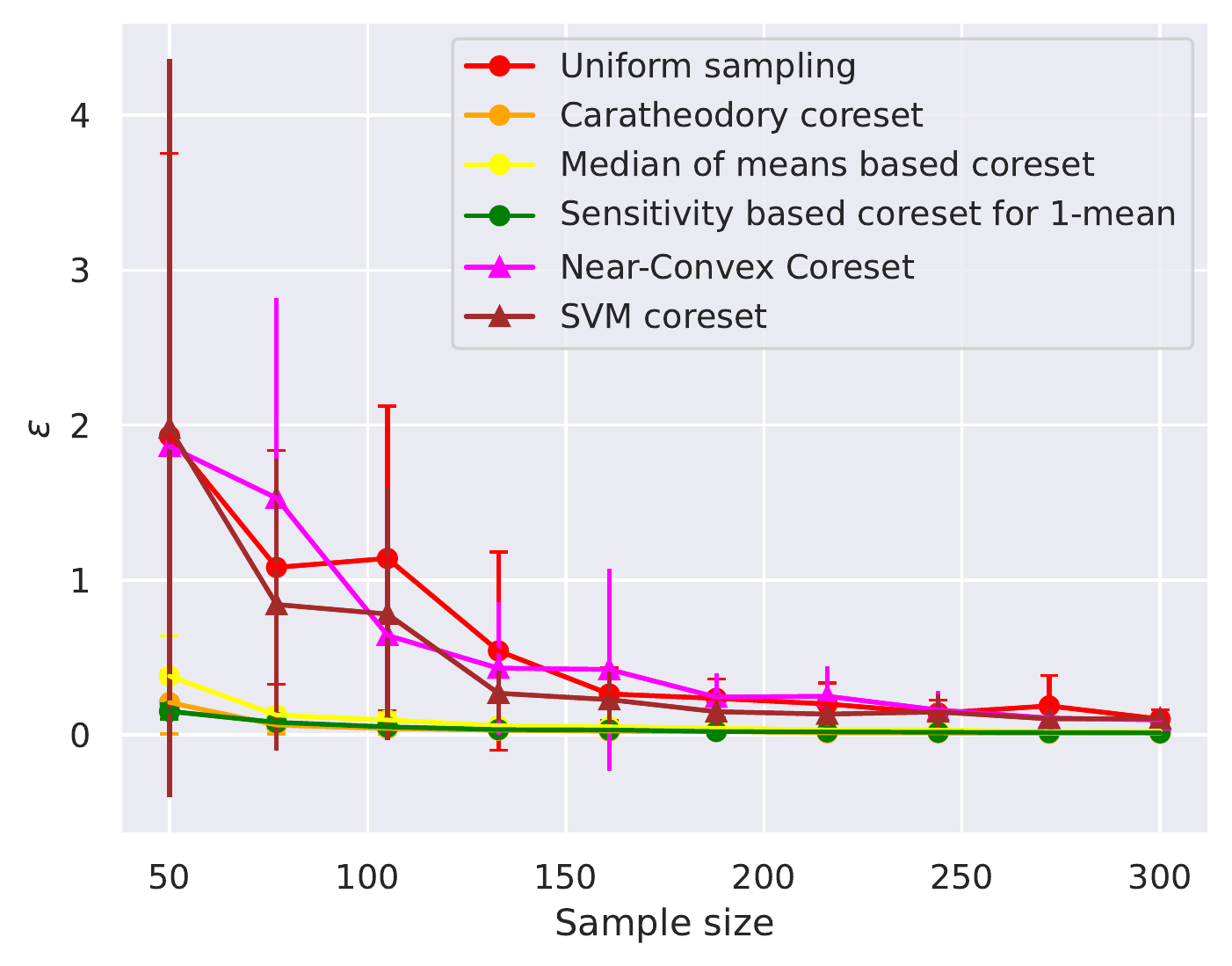}
    \includegraphics[width=0.24\textwidth]{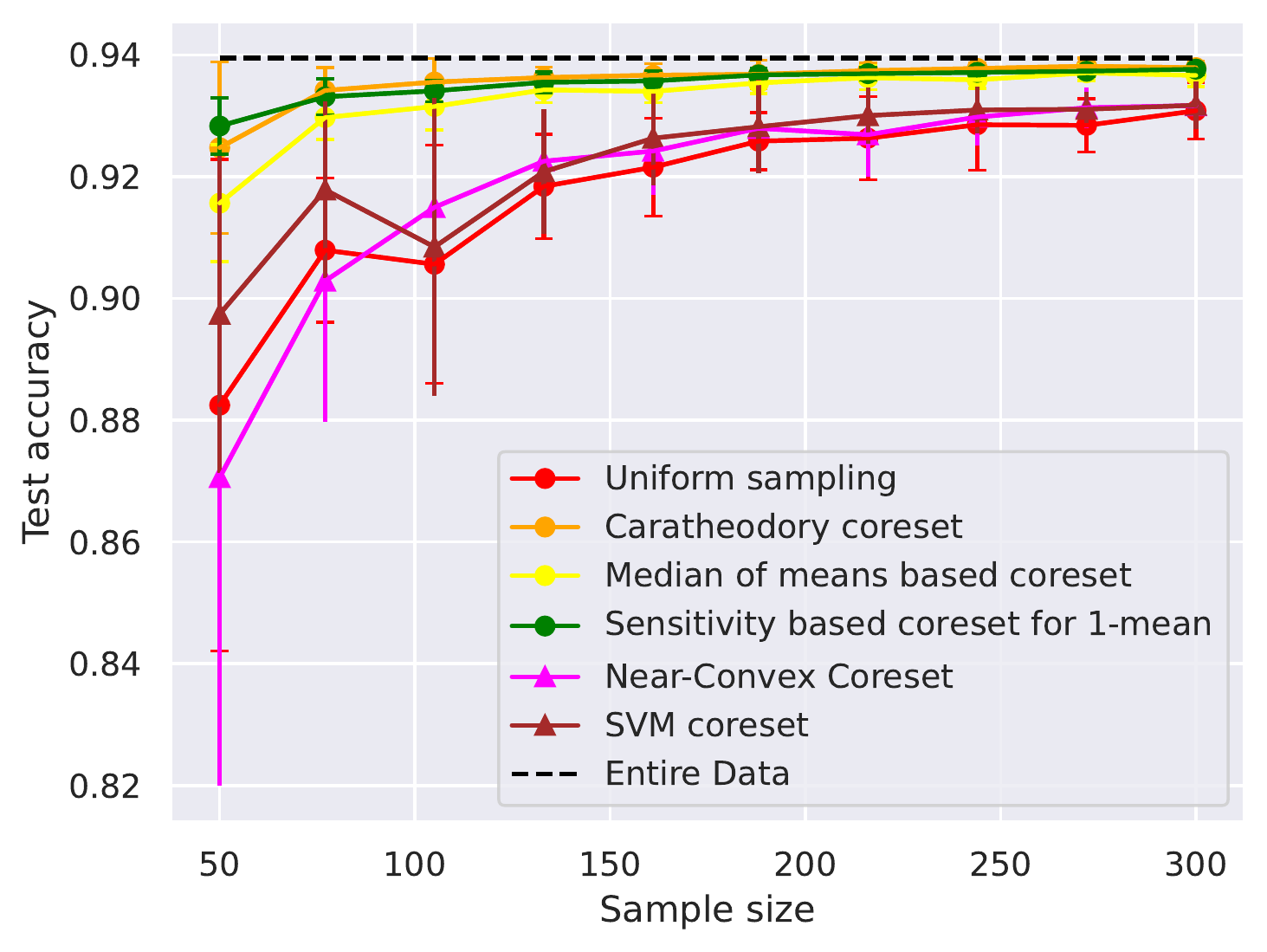}
     \label{fig:cod-rna_svm}}
     \caption{Evaluation of our coresets against other competing methods on the Dataset~\ref{dataset:cod_rna}.}
     \label{fig:cod}
 \end{figure*}

 \begin{figure*}[!htb]
    \subfigure[Logistic regression]{
    \includegraphics[width=0.24\textwidth]{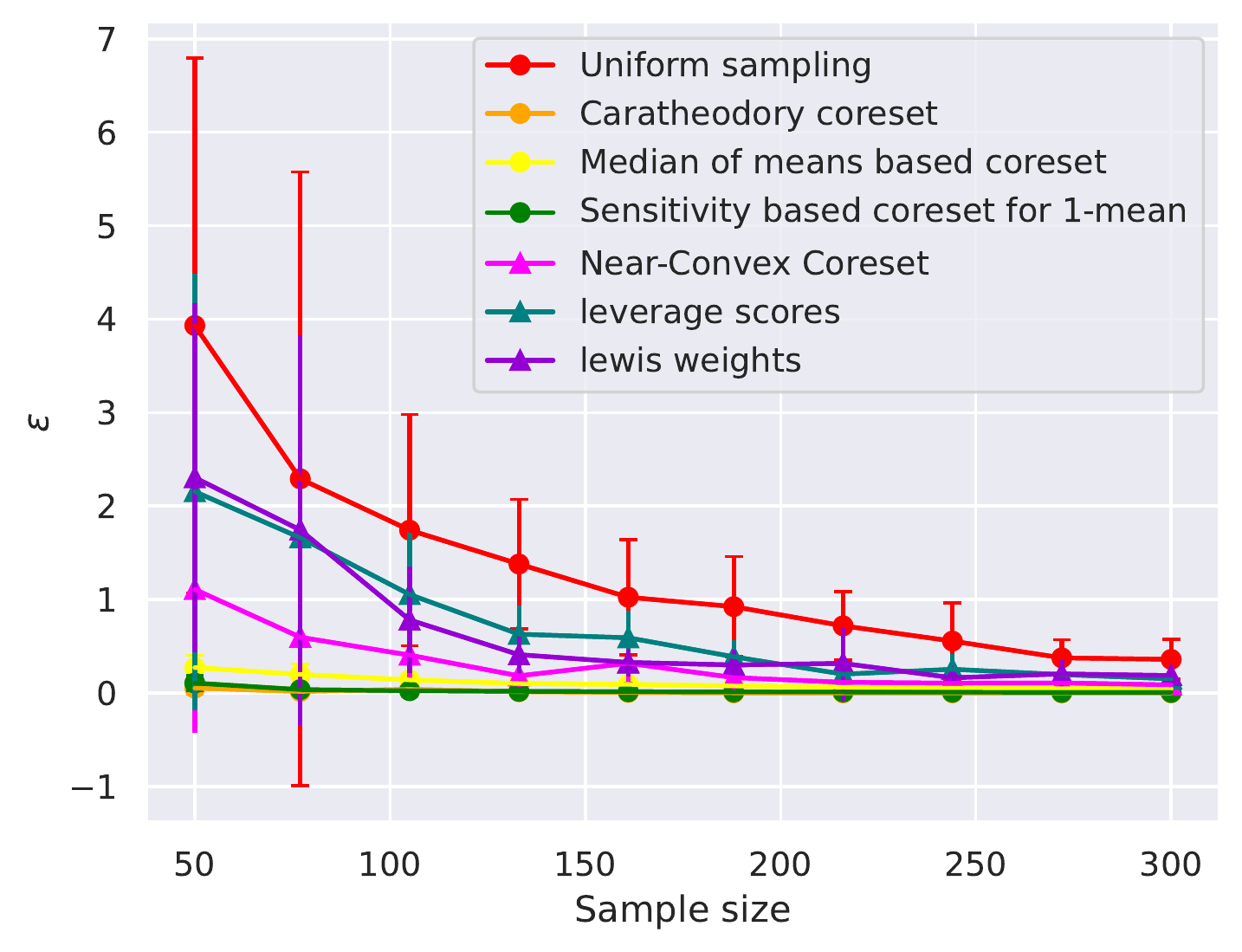}
    \includegraphics[width=0.24\textwidth]{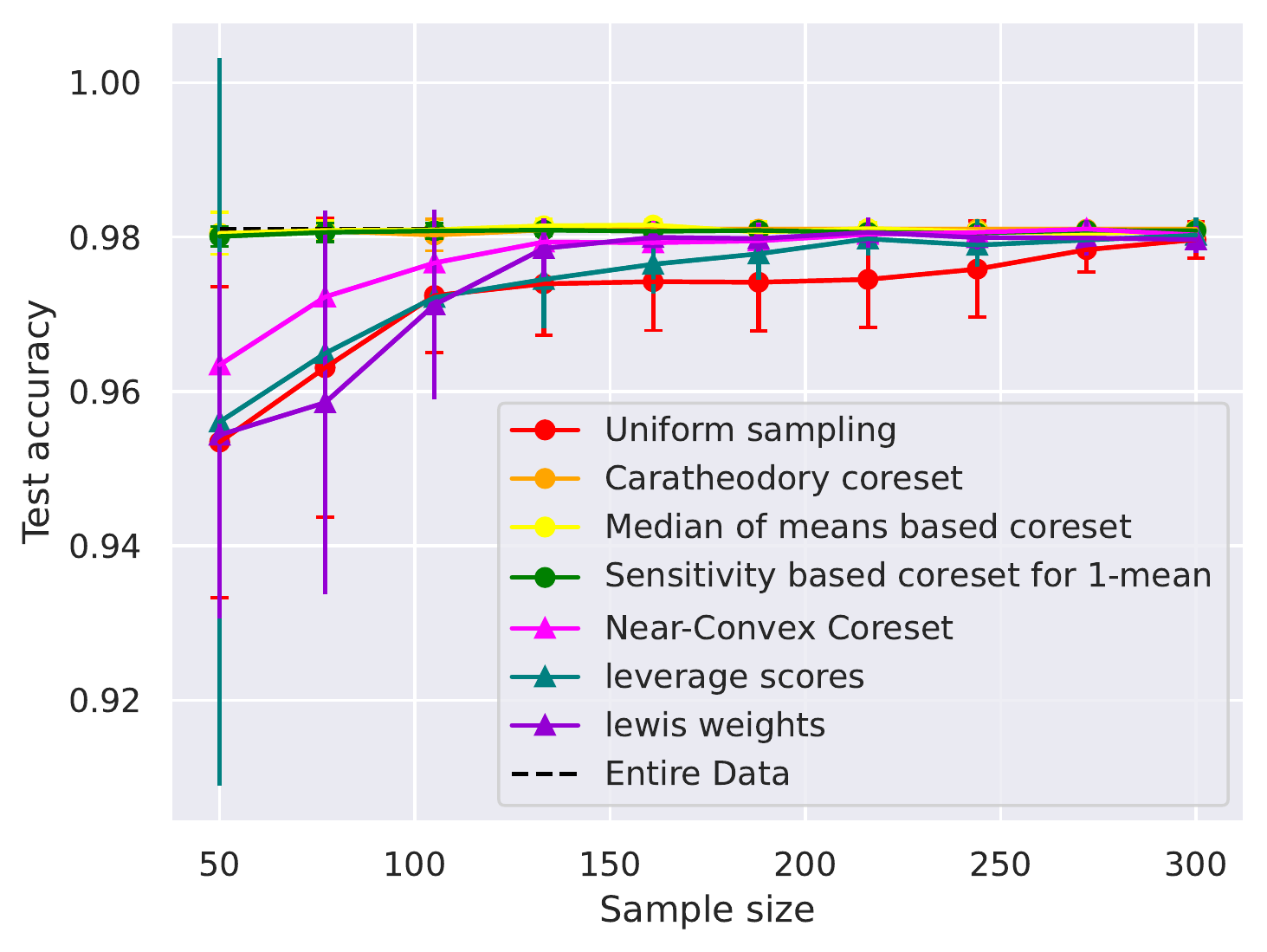}
    \label{fig:HTRU_2_logistic_regression}}
    \subfigure[SVMs]{
    \includegraphics[width=0.24\textwidth]{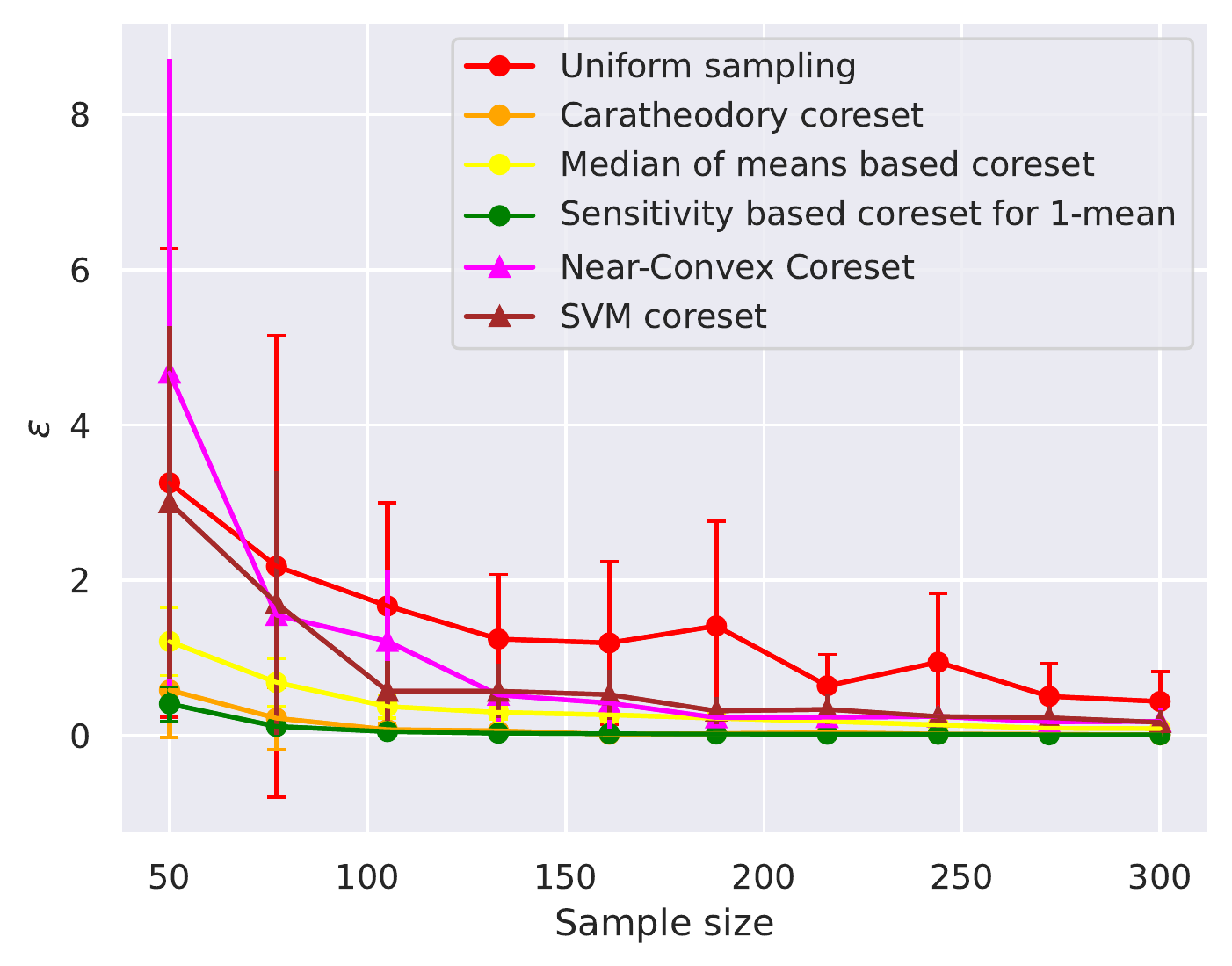}
    \includegraphics[width=0.24\textwidth]{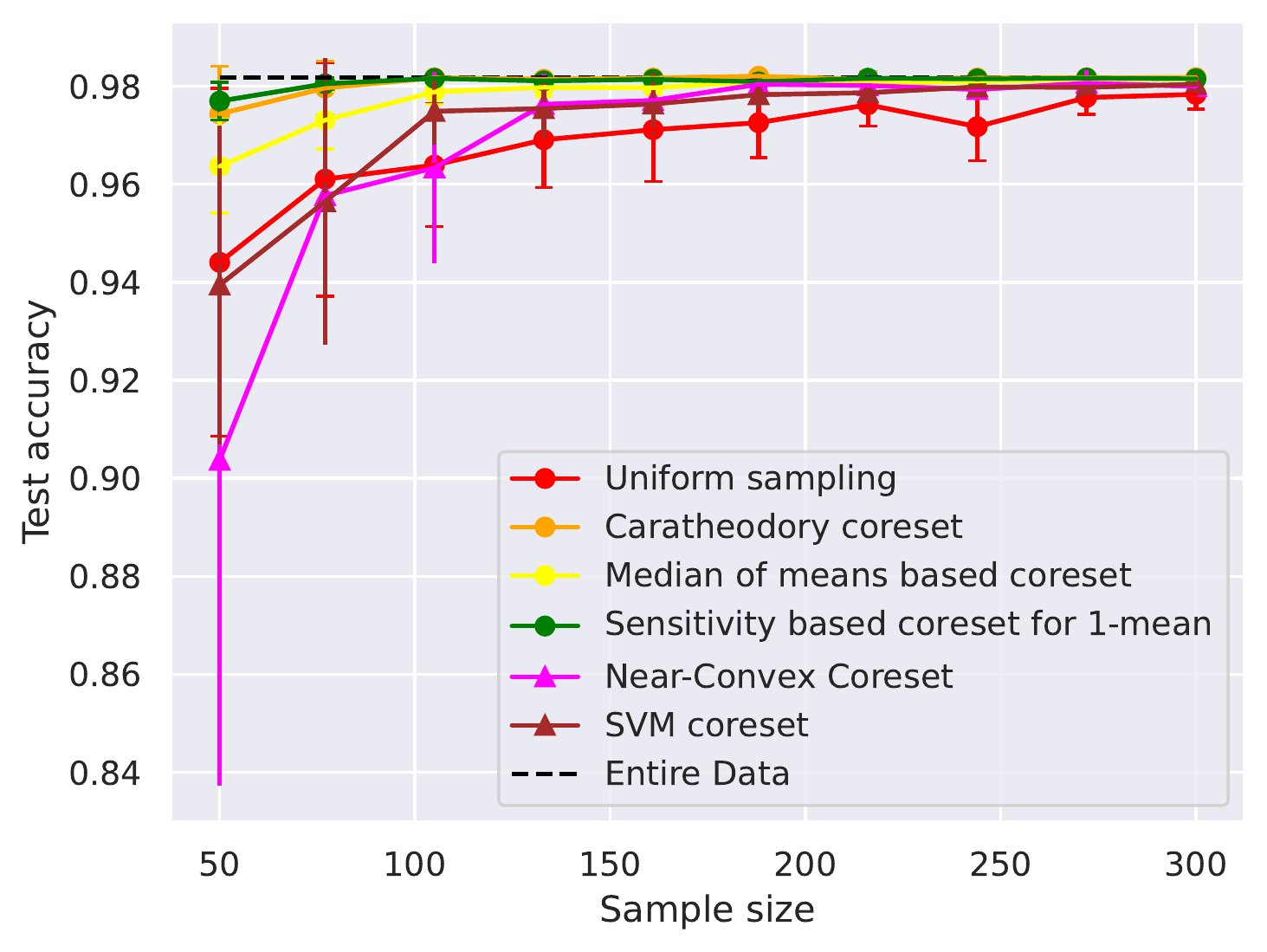}
     \label{fig:HTRU_2t_svm}}
     \caption{Evaluation of our coresets against other competing methods on the Dataset~\ref{dataset:HTRU}.}
     \label{fig:HTRU}
 \end{figure*}

In what follows, we first discuss the choices of different vector summarization coresets, and the used parameters in our experiments, followed by evaluating our coreset on real-world datasets, against other famous competing methods: Near Convex Coreset~\cite{NearConvex}, Lewis weights~\cite{munteanu2018logistic} and leverage scores~\cite{munteanu2018logistic} for logistic regression, Near Convex Coreset~\cite{NearConvex} and optimization based coreset~\cite{tukan2020coresets} for support vector machines (SVM), SVD-based coreset~\cite{maalouf2020tight} for linear regression, Bi criteria coreset~\cite{braverman2021efficient} for $k$-means, and uniform sampling in all of the experiments. We note that each experiment was conducted for $16$ trials, we report both the mean and std for all of the presented metrics. 

\textbf{Software/Hardware. } Our algorithms were implemented in Python 3.9~\cite{10.5555/1593511} using \say{Numpy}~\cite{oliphant2006guide}, \say{Scipy}~\cite{2020SciPy-NMeth} and \say{Scikit-learn}~\cite{scikit-learn}. Tests were performed on $2.59$GHz i$7$-$6500$U ($2$ cores total) machine with $16$GB RAM.

\subsection{\emph{AutoCoreset} parameters}
\textbf{Vector summarization coresets.} There are many methods for computing such coresets, some of them are deterministic, i.e., with no probability of failure, and others work with some probability $1-\delta$. On the other hand, some are accurate, i.e., $\eps=0$, and others yield an approximation error $\eps>0$. In Table~\ref{table:ourContrib} we summarize some of the common methods for computing such coresets, and their properties, such as size, running time, approximation error, and probability of failure. In our system, we implemented all of the given methods and compared them via extensive experiments.

\textbf{Setting the number of initial solutions $m$.} Throughout our experiments, we have set the number of initial solutions to $10$. The idea behind this is to expose $\ac$ to a number of  solutions that is not too high nor too low. Hence, we ensure that the coreset is not too weak nor too dependent on the initial solutions.

\begin{figure*}[!htb]
 \centering
    \subfigure[Full data]{
    \includegraphics[width=0.19\textwidth]{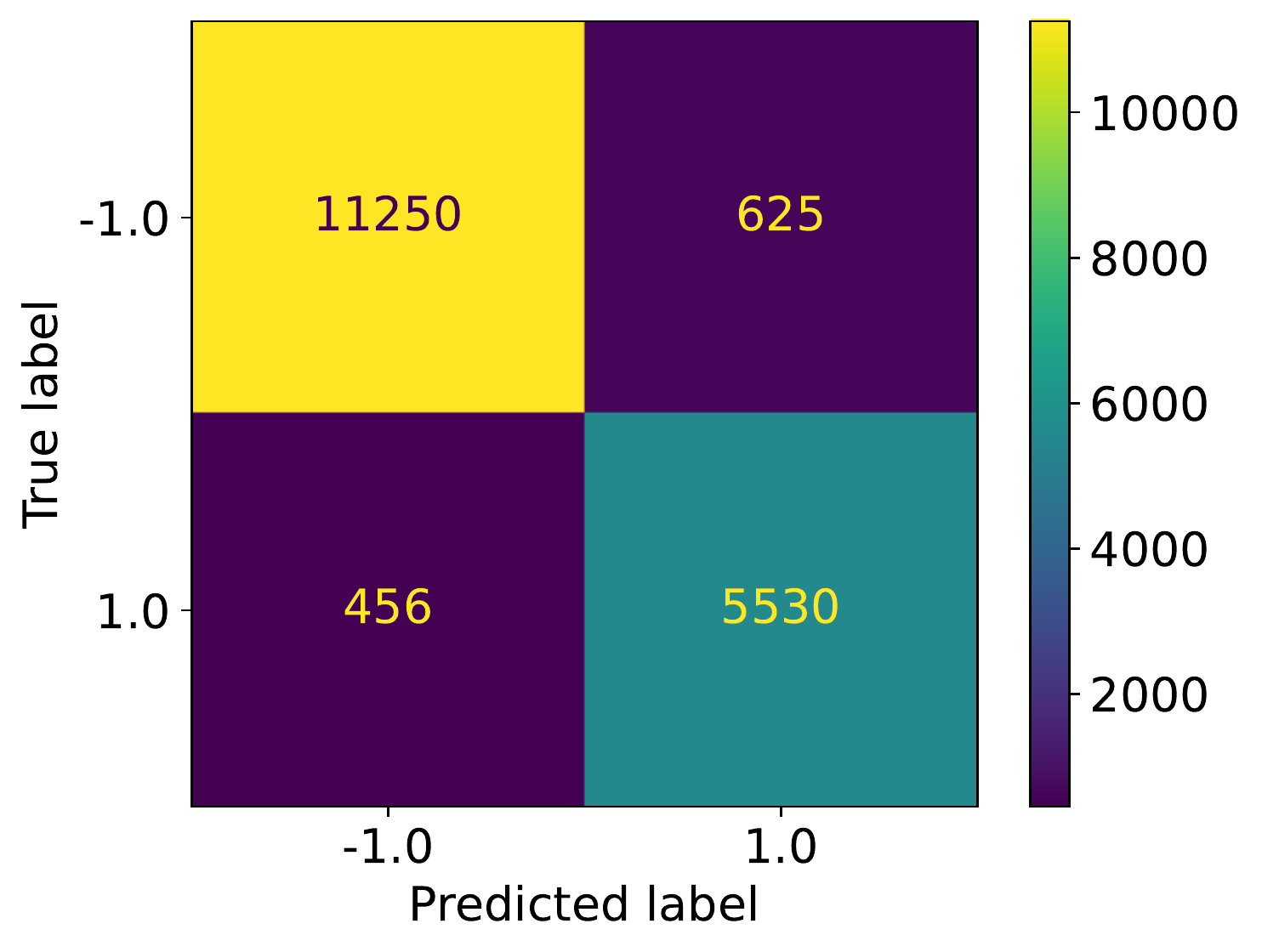}}
    \subfigure[Uniform sampling]{
     \includegraphics[width=0.19\textwidth]{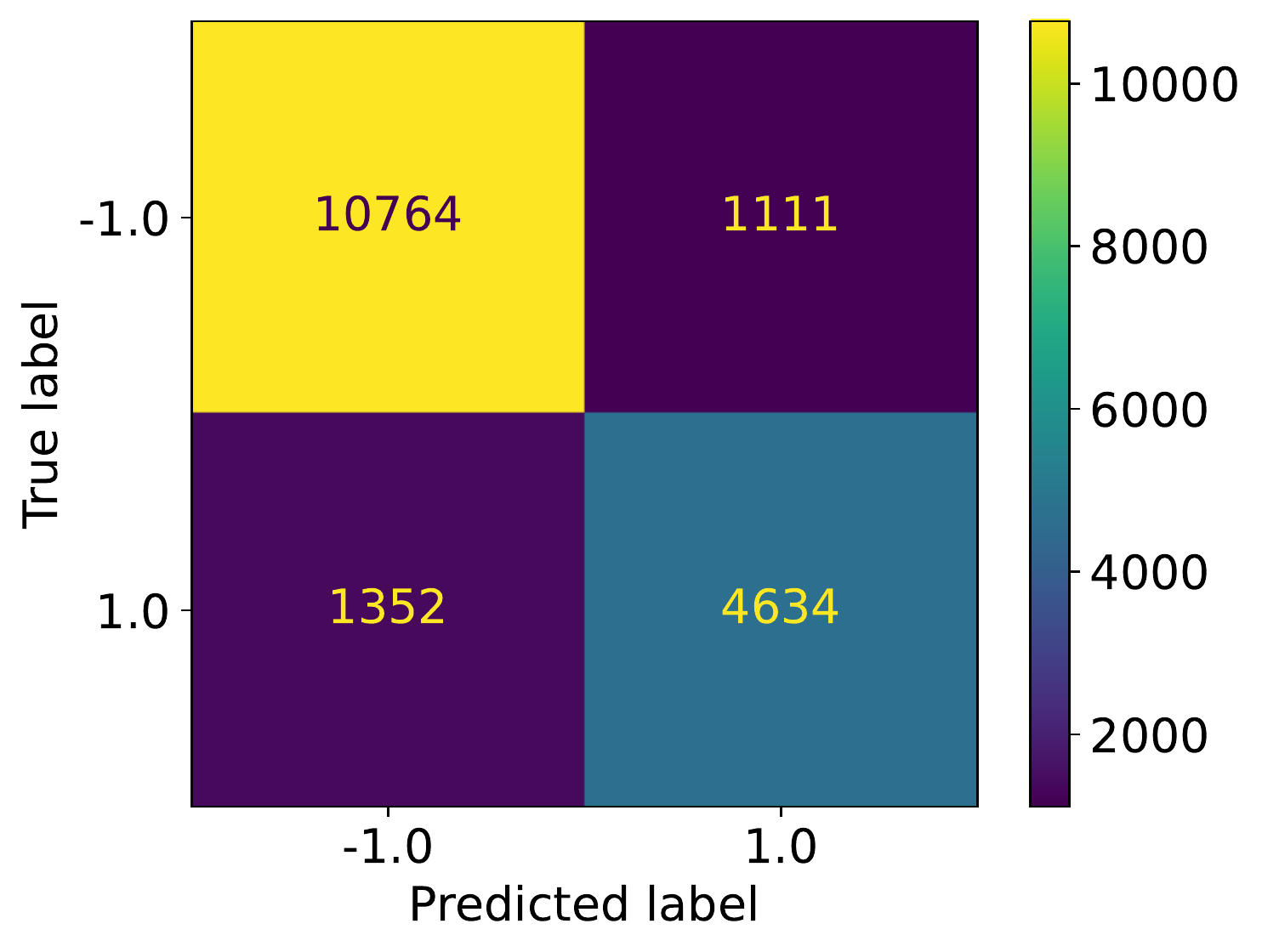}}
     \subfigure[Carath\'{e}dory coreset]{
     \includegraphics[width=0.19\textwidth]{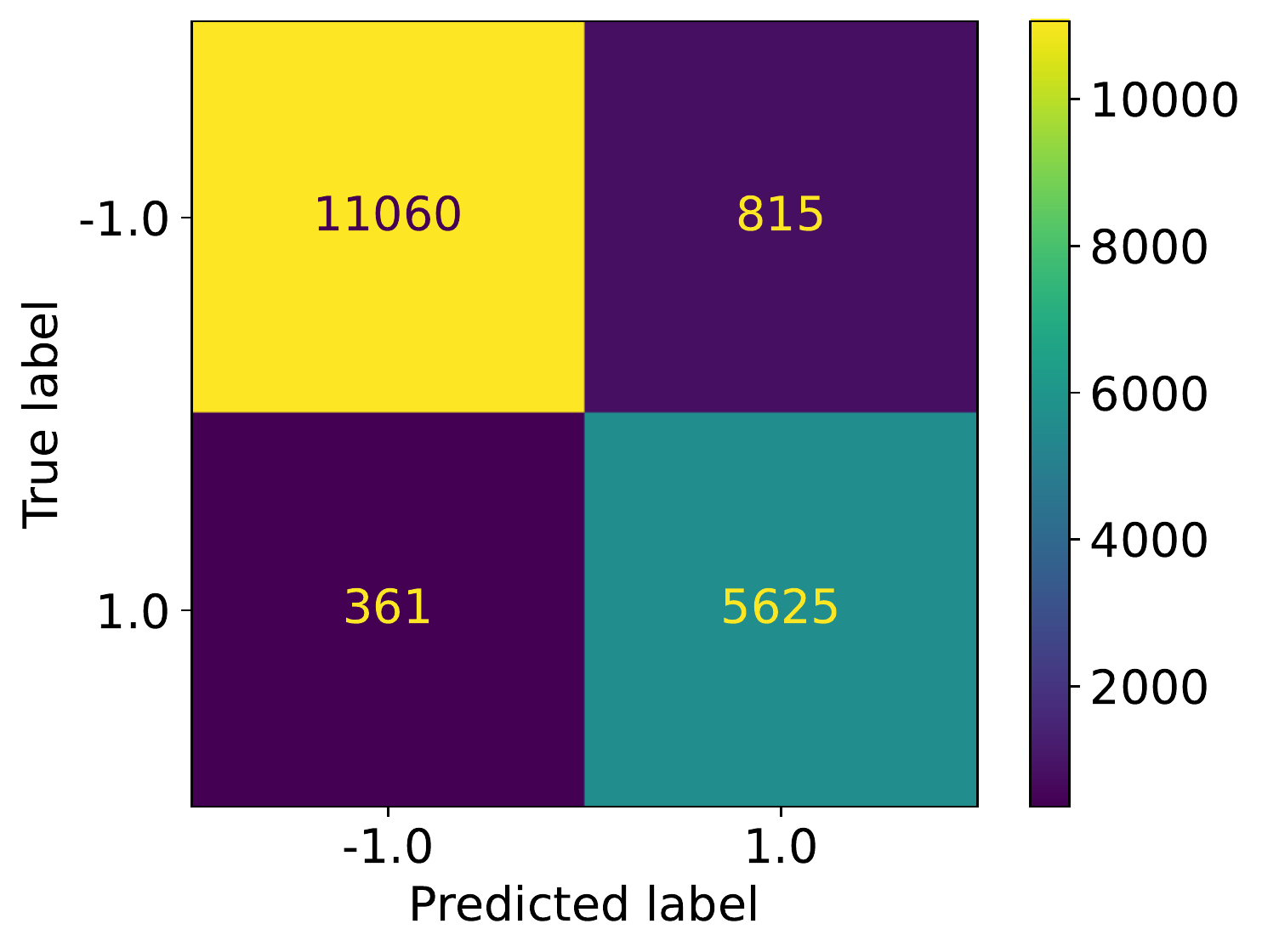}}
    \subfigure[Median of means \newline coreset]{
     \includegraphics[width=0.19\textwidth]{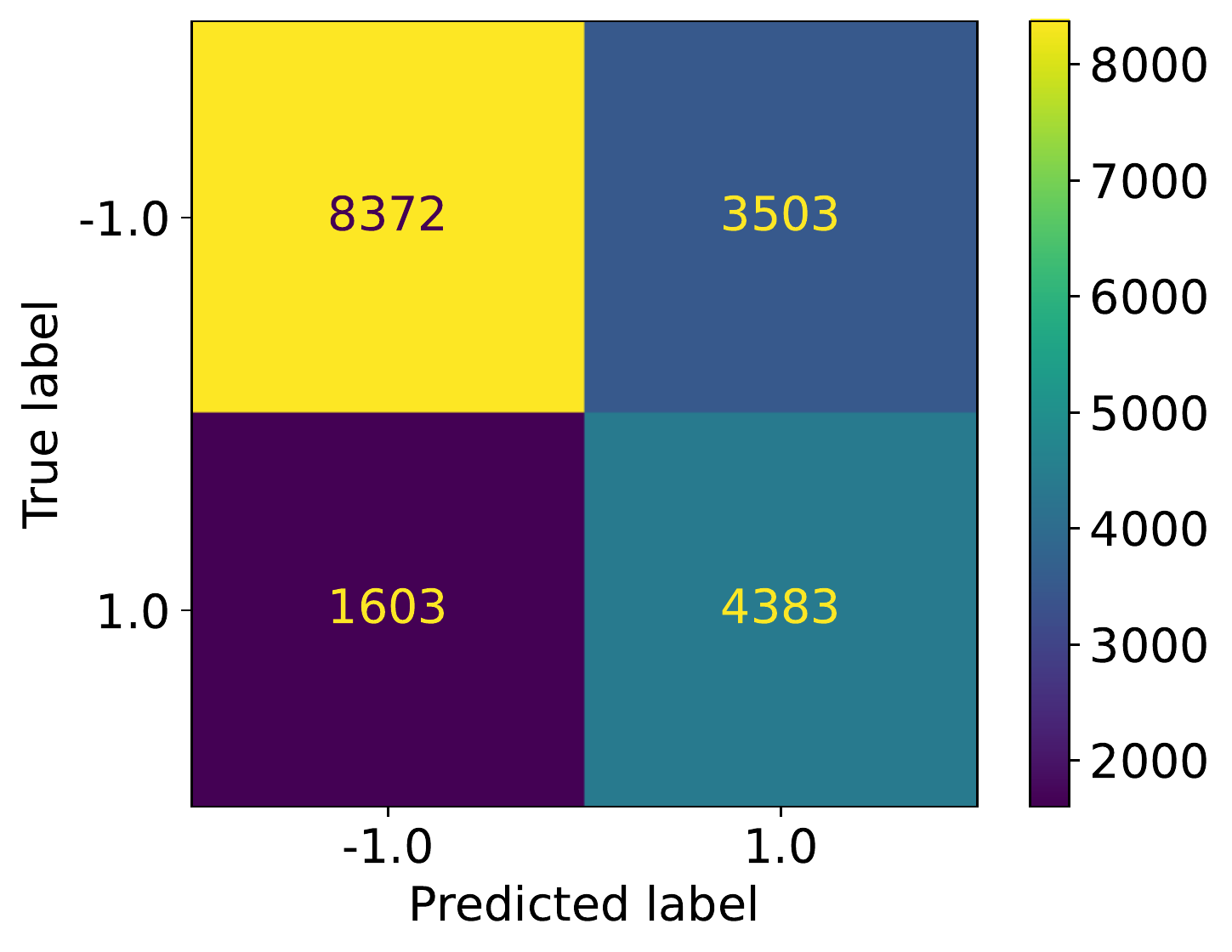}}
     \subfigure[Importance sampling \newline based coreset]{
     \includegraphics[width=0.19\textwidth]{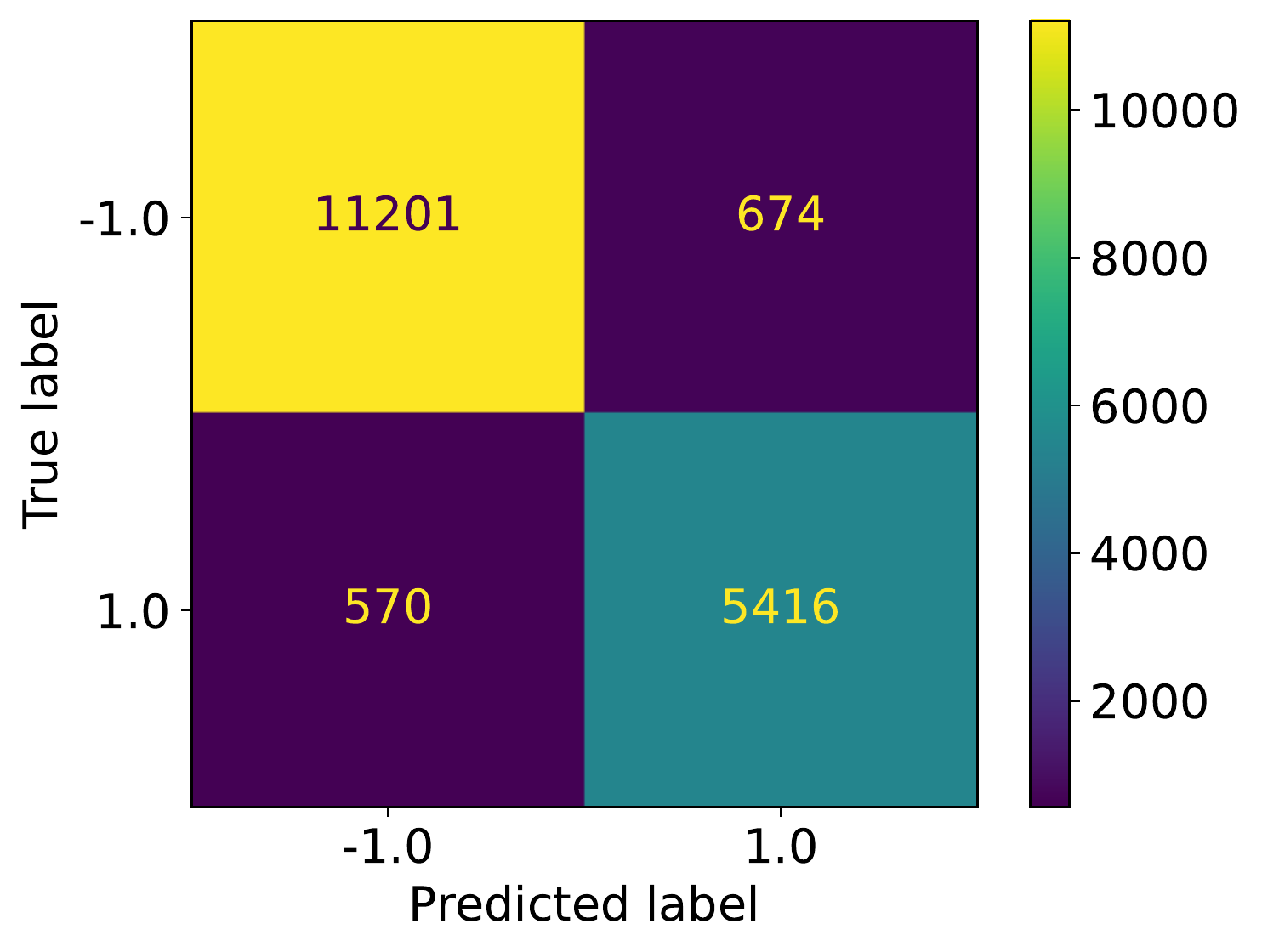}}
     \caption{SVM confusion matrices with respect to our coresets against Uniform sampling and the entire data of Dataset~\ref{dataset:cod_rna}.}
     \label{fig:logistic_cod_rna_svm}
 \end{figure*}

 \begin{figure*}[!t]
 \centering
    \subfigure[Full data]{
    \includegraphics[width=0.19\textwidth]{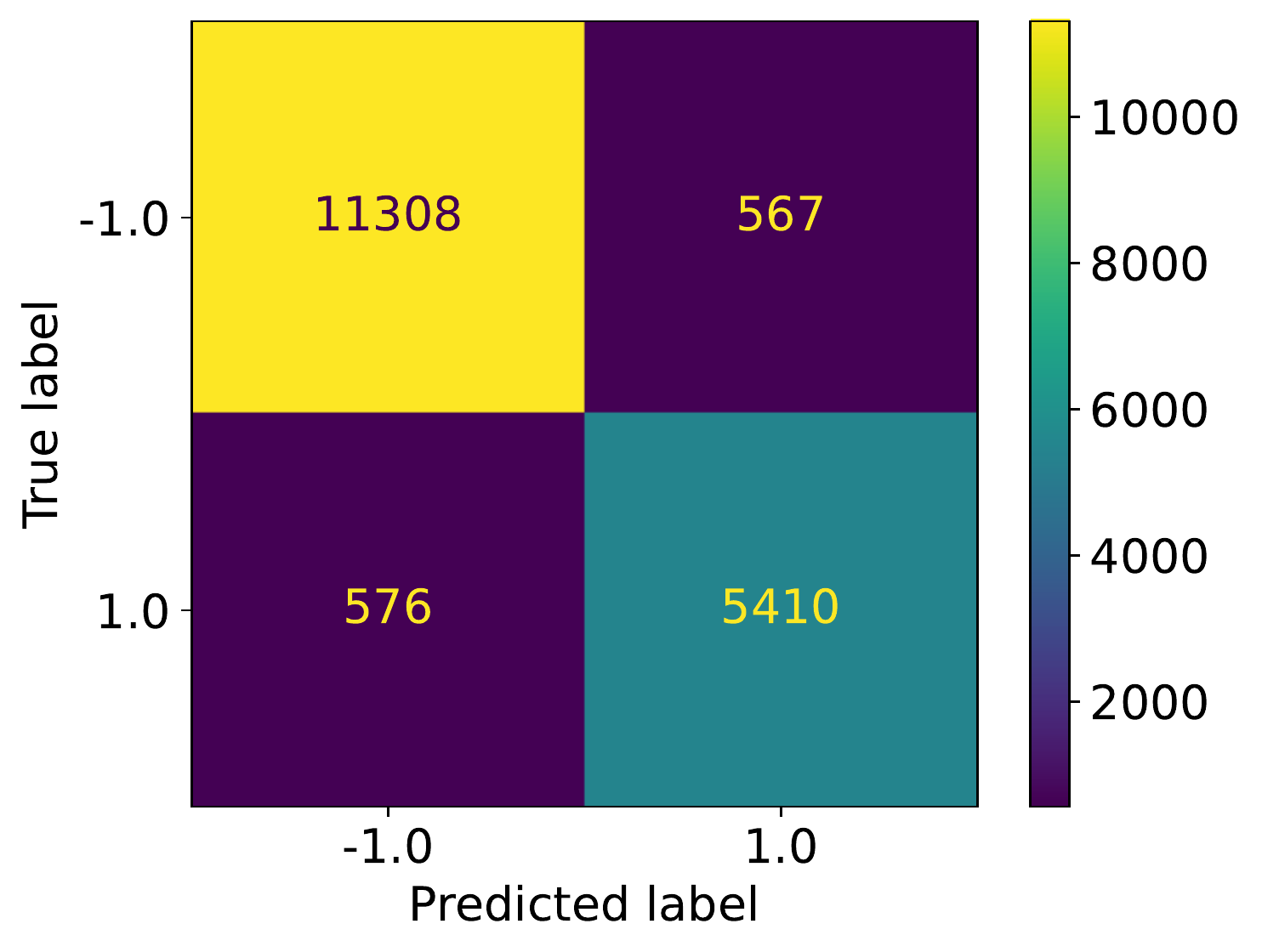}}
    \subfigure[Uniform sampling]{
     \includegraphics[width=0.19\textwidth]{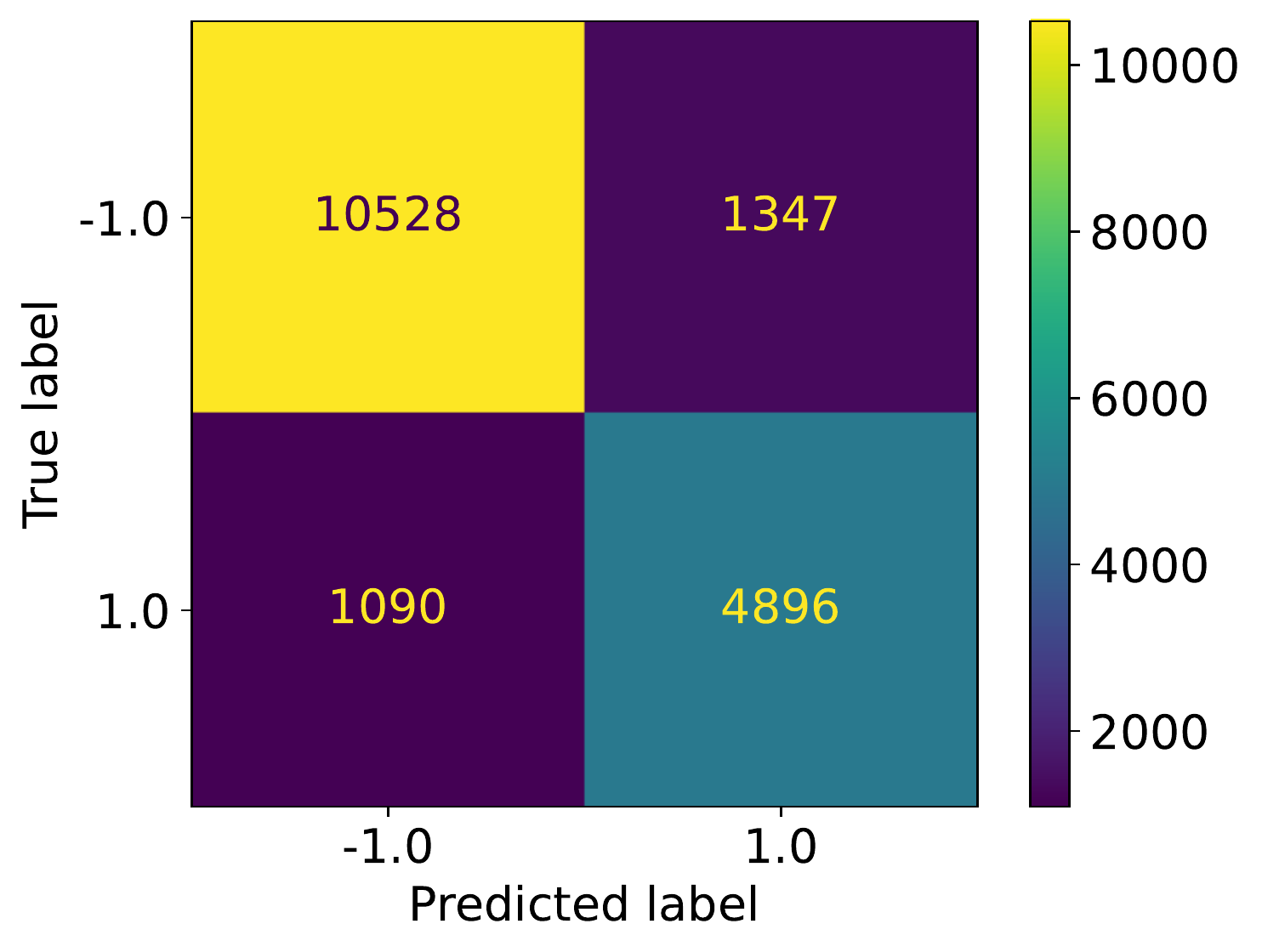}}
     \subfigure[Carath\'{e}dory coreset]{
     \includegraphics[width=0.19\textwidth]{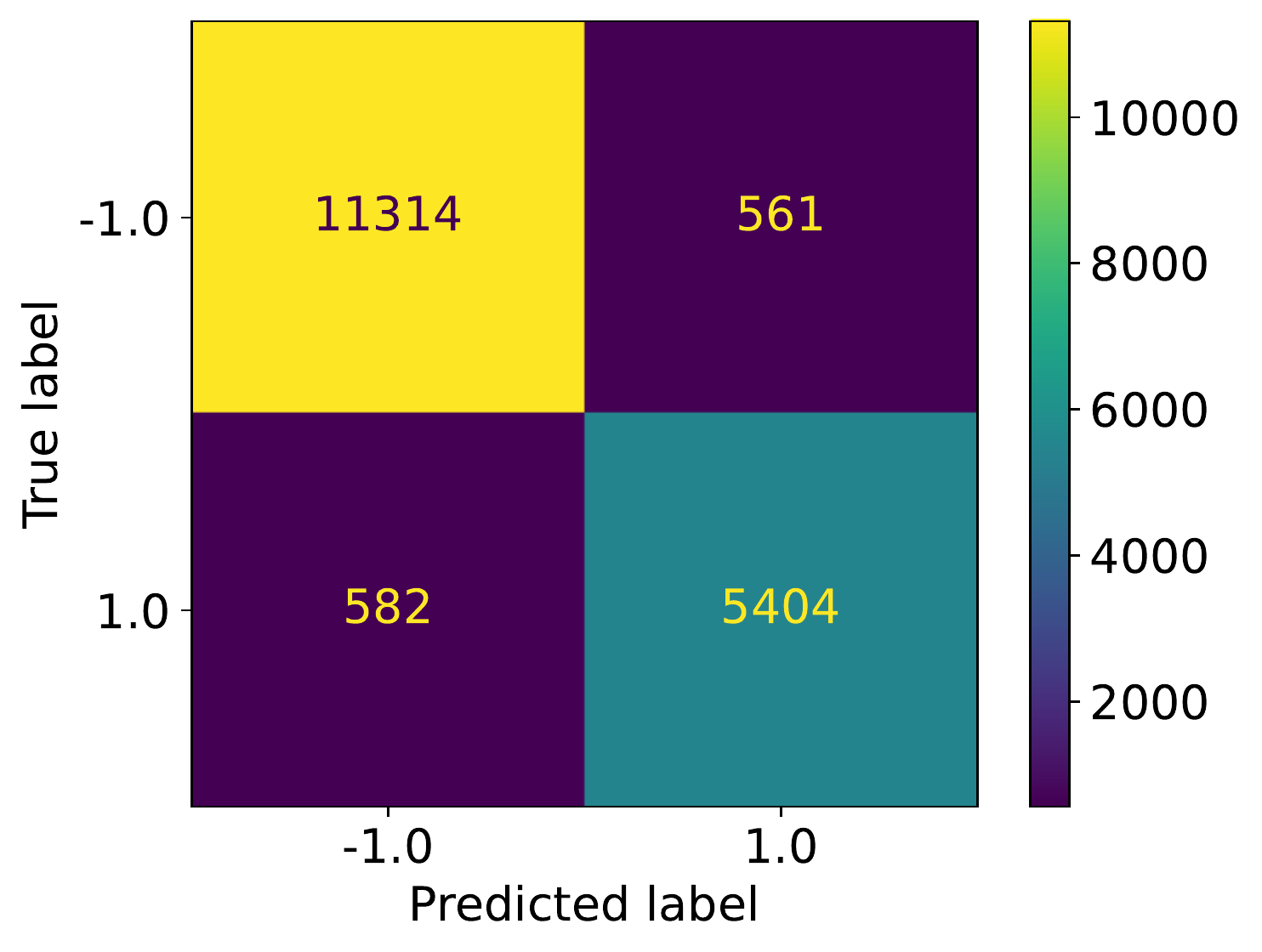}}
    \subfigure[Median of means \newline coreset]{
     \includegraphics[width=0.19\textwidth]{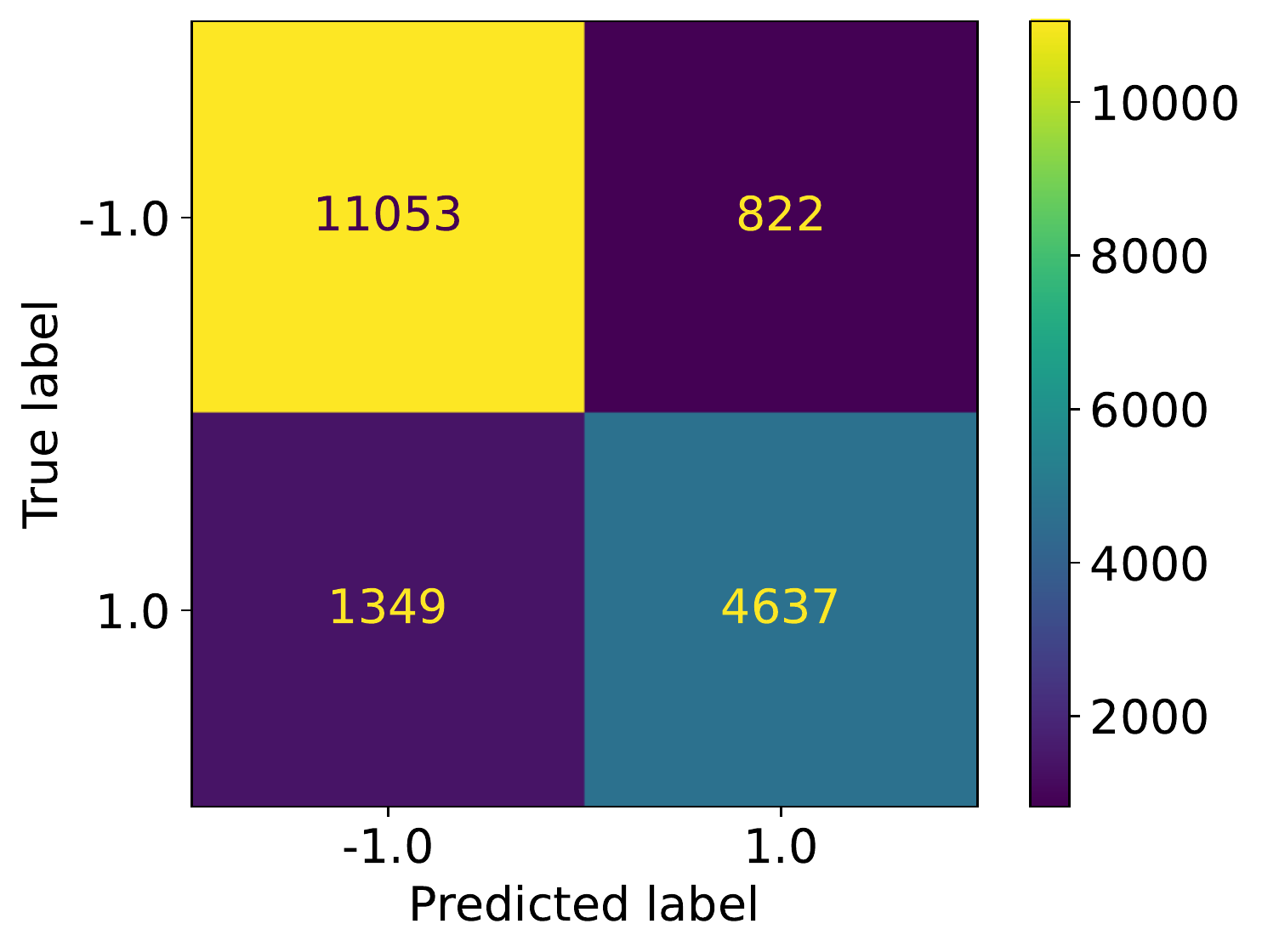}}
     \subfigure[Importance sampling \newline based coreset]{
     \includegraphics[width=0.19\textwidth]{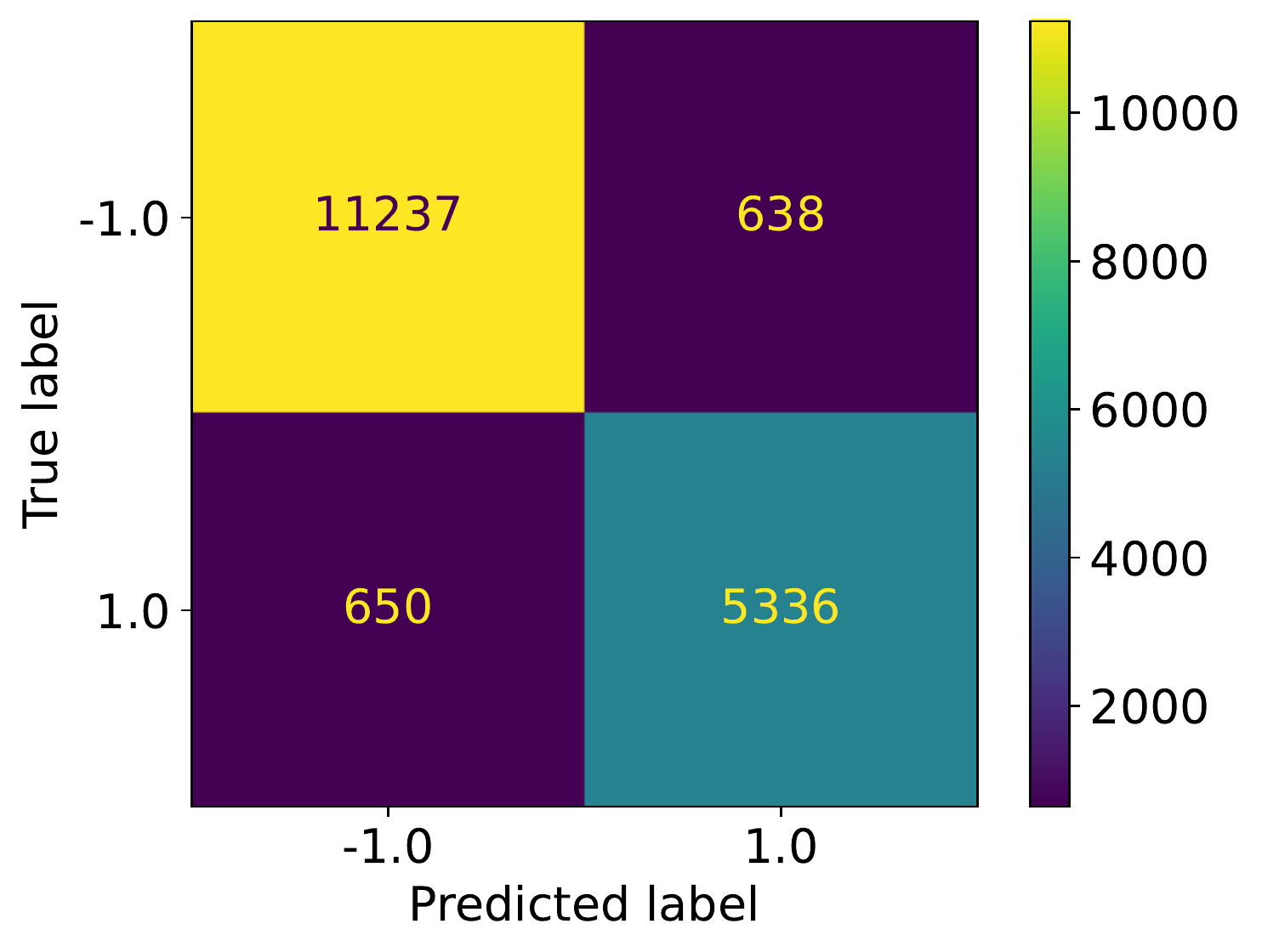}}
     \caption{Logistic regression confusion matrices with respect to our coresets against Uniform sampling and the entire data of Dataset~\ref{dataset:cod_rna}.}
     \label{fig:confusion_cod_rna_logistic}
 \end{figure*}

\textbf{Choosing a stopping criterion $\zeta$.} Inspired by the early-stopping mechanism of~\cite{prechelt1998early}, we adopt a similar idea. We make use of a parameter, namely \say{patience}, which was set to $7$, to attempt an indication of the occurrence of saturation with respect to the exposure of our coreset paradigm to new queries; see more details at Section~\ref{sec:moredetails}. 
To correctly use this parameter, we use additional two parameters, one of which is a counter, while the other holds the optimal coreset that resulted in the smallest sum of the entries of the concatenated columns (see Line~\ref{line:add_column_to_M} at Algorithm~\ref{alg:main}). The counter will be reset to $0$ once a new column is added such that its sum is lower than the smallest sum so far, and the optimal coreset will be updated. Otherwise, the counter will be increased. $\ac$ will keep running until the above counter reaches the \say{patience} parameter. In our experiments, we returned the optimal coreset since it led to better results. For completeness, we refer the reader to the appendix where we conduct an ablation study and check our results without taking the optimal coreset, i.e., in those results, we take the last coreset. Note that, in both sets of experiments, we outperform the competing methods.


\textbf{Datasets.} The following datasets were used throughout our experimentation. These datasets were taken from~\cite{Dua:2019} and~\cite{CC01a}: \begin{enumerate*}[label=(\roman*)]
    \item Credit card dataset~\cite{yeh2009comparisons} composed of $30000$ points with $24$ features representing customers' default payments in Taiwan, \label{dataset:credit}
    \item Cod-RNA dataset~\cite{uzilov2006detection}: dataset containing $59535$ points with $8$ features, \label{dataset:cod_rna}
    \item HTRU dataset~\cite{lyon2016fifty}: Pulsar candidates collected during the HTRU survey containing $17898$ each with $9$ features, \label{dataset:HTRU}
    \item $3D$ Road Network~\cite{guo2012ecomark}: $3D$ road network with highly accurate elevation information from Denmark containing $434874$ points each with $4$ features, \label{dataset:3D}
    \item Accelerometer dataset~\cite{ScalabriniSampaio2019}: an accelerometer data from vibrations of a cooler fan with weights on its blades containing $153000$ points consisting each of $5$ features, \label{dataset:accemolator} and 
    \item Energy efficiency Data Set~\cite{tsanas2012accurate}: a dataset containing $768$ points each of $8$ features. \label{dataset:EB}
\end{enumerate*}

\textbf{ML models.} Throughout our entire set of experiments, we have relied on \say{Scikit-Learn} ML models.

\textbf{Reported results.} 
First, for each coreset $(\C,v)$ of an input data $P$ and a loss function $f$, we compute the optimal solution on the coreset $x^*_{\C}\in\argmin_{X\in \mathcal{X}}\sum_{i\in \C}v(i)f(p_i,x) $, and on the real data $x^*_{P}\in\argmin_{x\in \mathcal{X}}\sum_{i\in [n]}f(p_i,x) $, and we report the optimal solution approximation error $\eps=\abs{ \sum_{i\in [n]} f(p_i,x^*_{\C}) -  \sum_{i\in [n]} f(p_i,x^*_{P})}$. Secondly, we show for classification problems the test accuracy obtained when training on the coreset, while on regression problems we show an estimate of the coefficient of determination of the prediction $R^2$~\cite{ozer1985correlation}. Additional measures are reported for some problems; we discuss them in the following sections. The bars in our graphs reflect the standard deviation. 

\subsection{Traditional ML classification problems}

\begin{figure*}[!t]
    \subfigure[Dataset~\ref{dataset:3D}]{
    \includegraphics[width=0.245\textwidth]{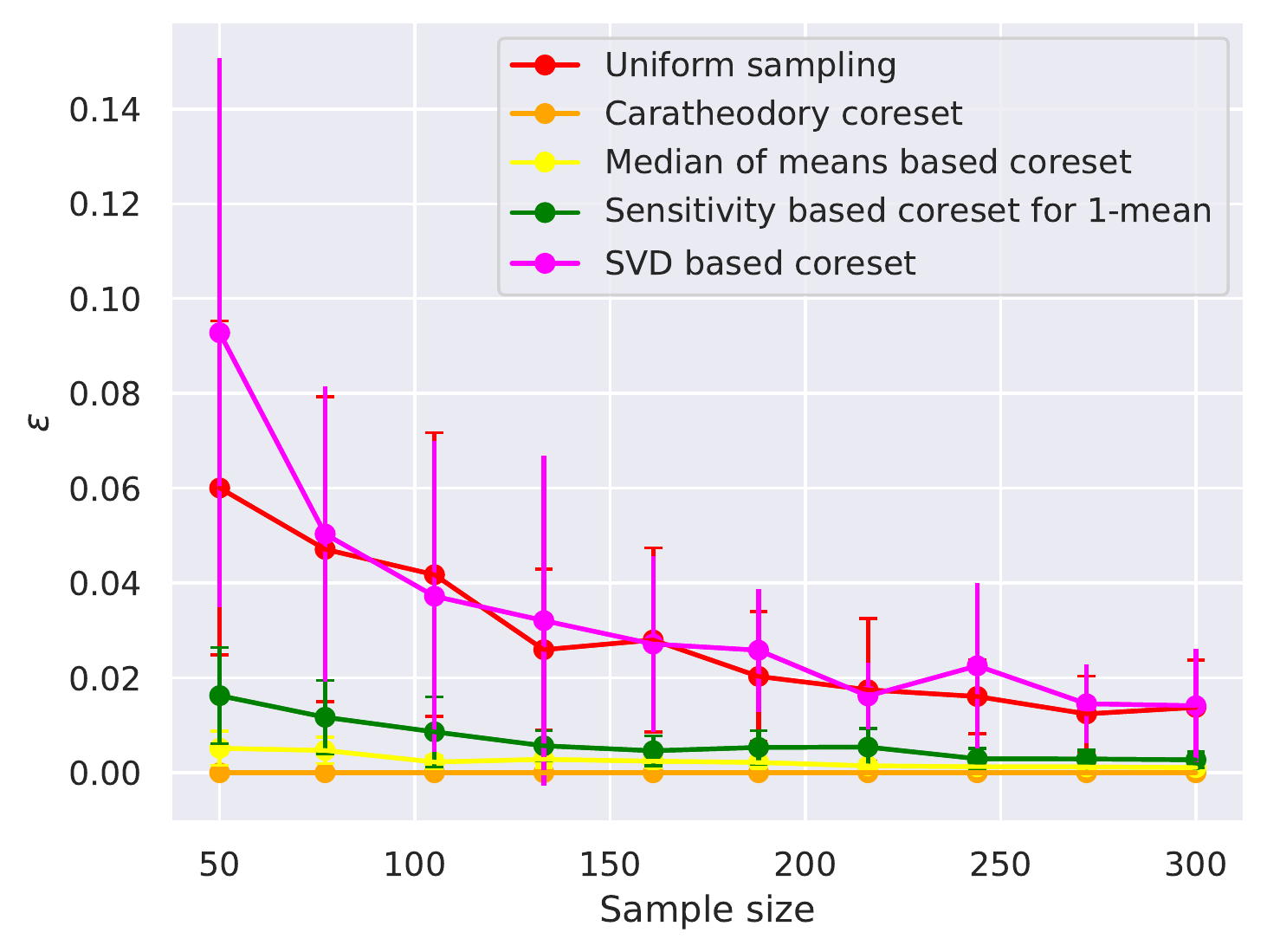}
    \includegraphics[width=0.245\textwidth]{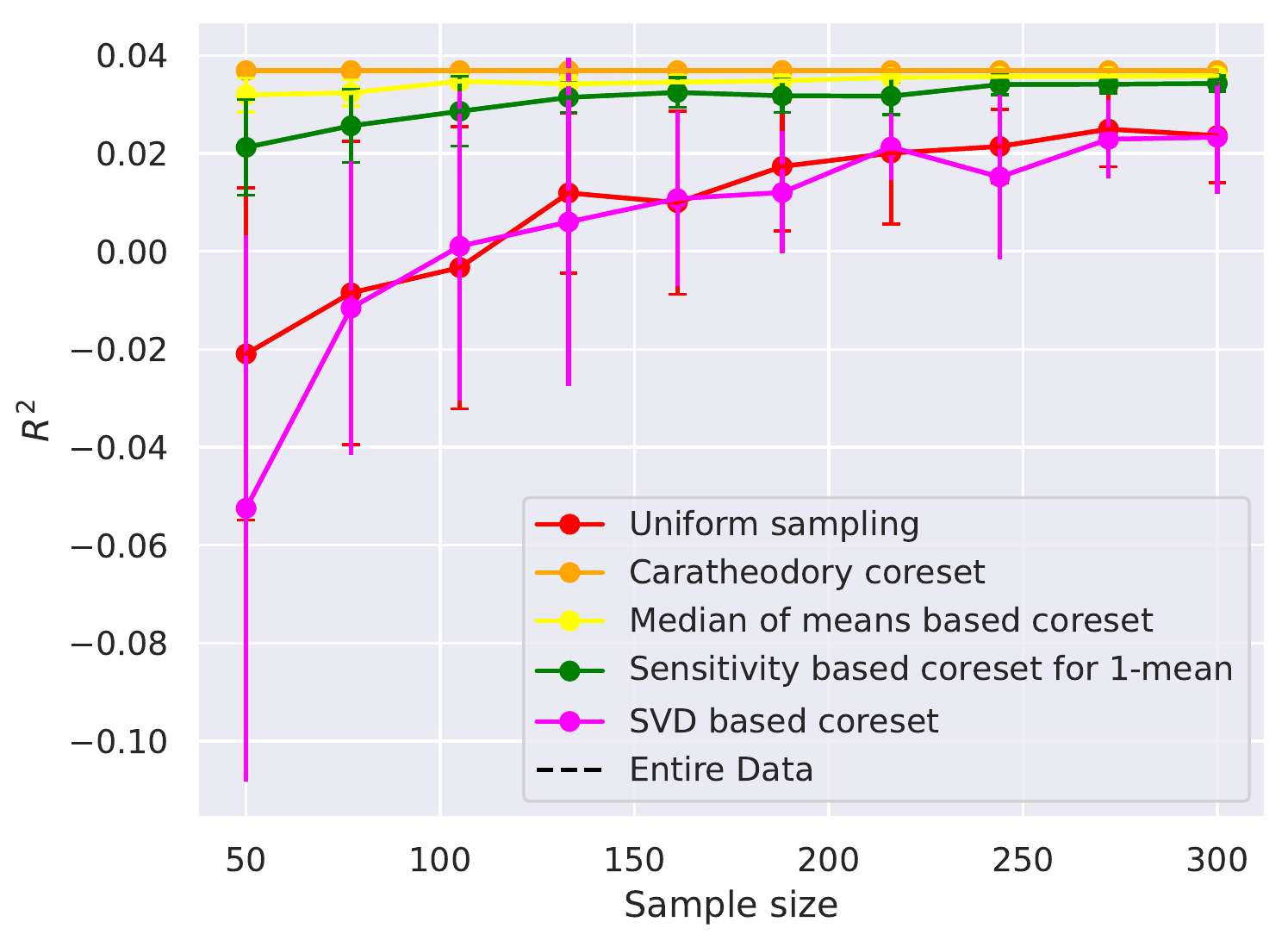}
    \label{fig:3D_spatial_network_linear_regression}}
    \subfigure[Dataset~\ref{dataset:accemolator}]{
    \includegraphics[width=.245\linewidth]{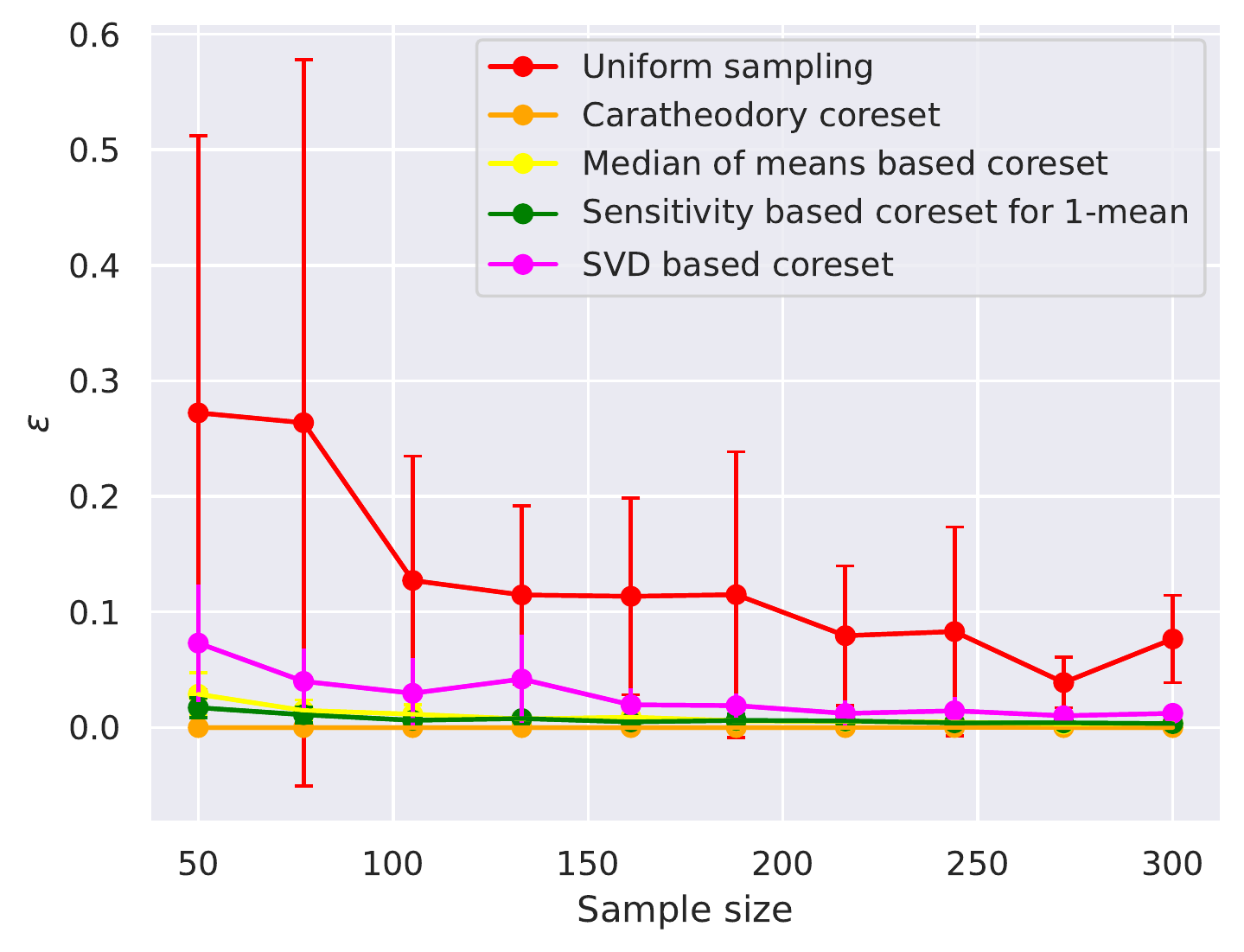}
    \includegraphics[width=.245\linewidth]{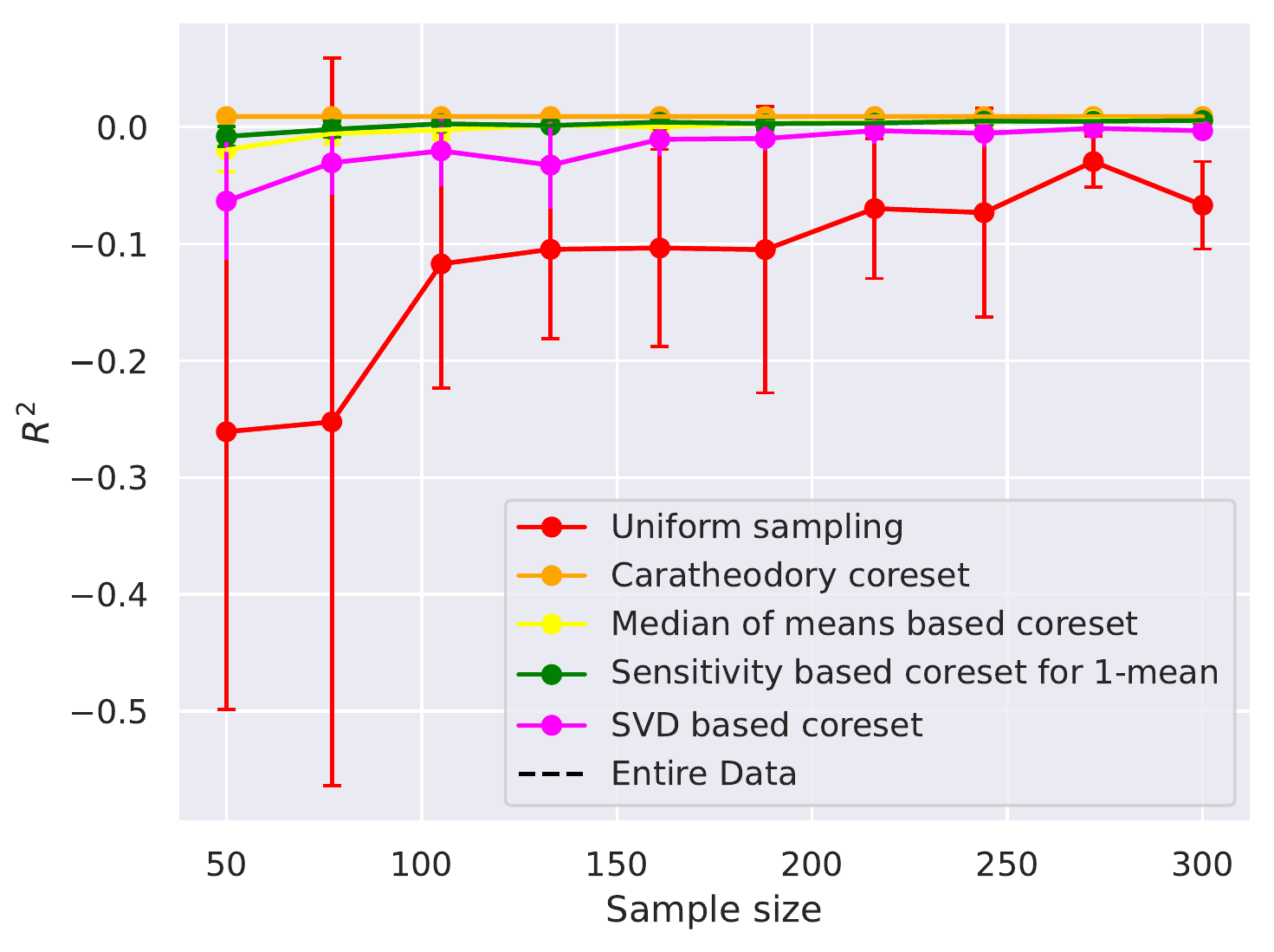}
    }
    
     \caption{Evaluation of our coresets against other competitors concerning the linear regression problem.}
     \label{fig:linear_regression_problem}
 \end{figure*}


In what follows, we show our results when setting $f$ to be the loss function of either the \emph{Logistic} regression problem or the \emph{SVMs} problem. In both experiments, since, some of the datasets were unbalanced, each sample coreset size has been split -- small classes get a slightly larger portion of the sample size than simply taking $\eta \times $ sample size where $\eta$ represents the class size percentage with respect to the total number of points, while larger classes get a portion of the sample size smaller than $\eta \times $ sample size.

\textbf{Logistic regression.} We have set the maximal number of iterations to $1000$ (for the Scikit-Learn solver) while setting the regularization parameter to $1$. Our system's approximation error was smaller by orders of magnitude, and the accuracy associated with the models trained using our coreset was better than the model trained on the competing methods; see Figure~\ref{fig:credit_logistic_regression} and Figure~\ref{fig:HTRU_2_logistic_regression}. On the other hand, Figure~\ref{fig:cod-rna_logistic_regression} depicts a multiplicative gap of  $30$ with respect to the approximation error in comparison to the competing methods while simultaneously acceding by $5\%$ accuracy gap over them.  In addition, we present the confusion matrix for each of our coresets using \emph{AutoCoreset}, and compare it to the confusion matrices with respect to the entire data and the uniform sampling coreset; See Figure~\ref{fig:confusion_cod_rna_logistic}. The confusion matrices aim towards explaining our advantage as our system outputs coresets that approximately maintain the structural properties of the confusion matrix of the entire data better than simply using uniform sampling, as our recall and accuracy are closer to their corresponding values when using the entire dataset.

 \begin{figure}[t]
\centering
    \subfigure[Dataset~\ref{dataset:HTRU}, $k:=5$]{
    \includegraphics[width=0.47\linewidth]{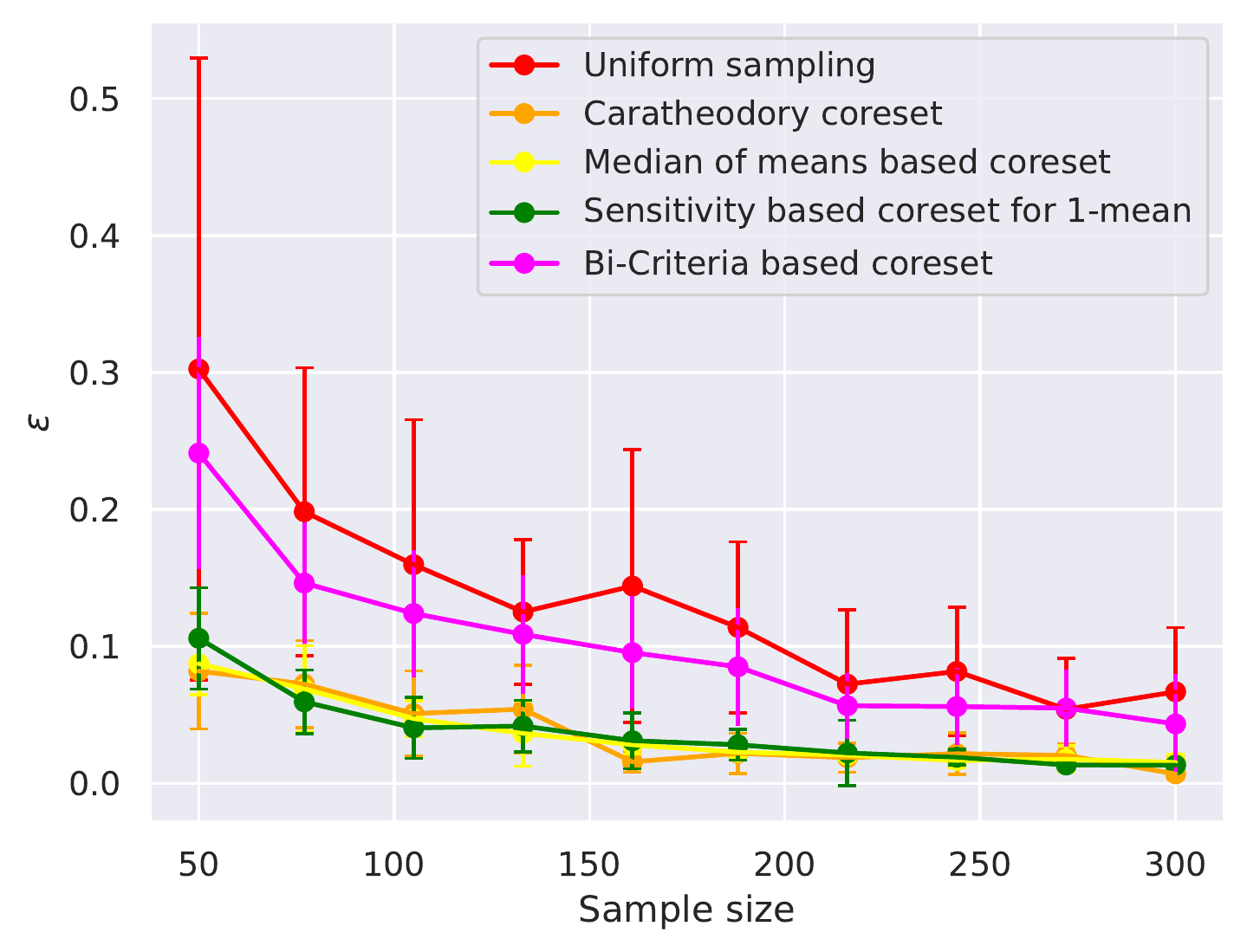}
    \label{fig:k_means_HTRU}}
    \subfigure[Dataset~\ref{dataset:EB}, $k:=2$]{
    \includegraphics[width=.47\linewidth]{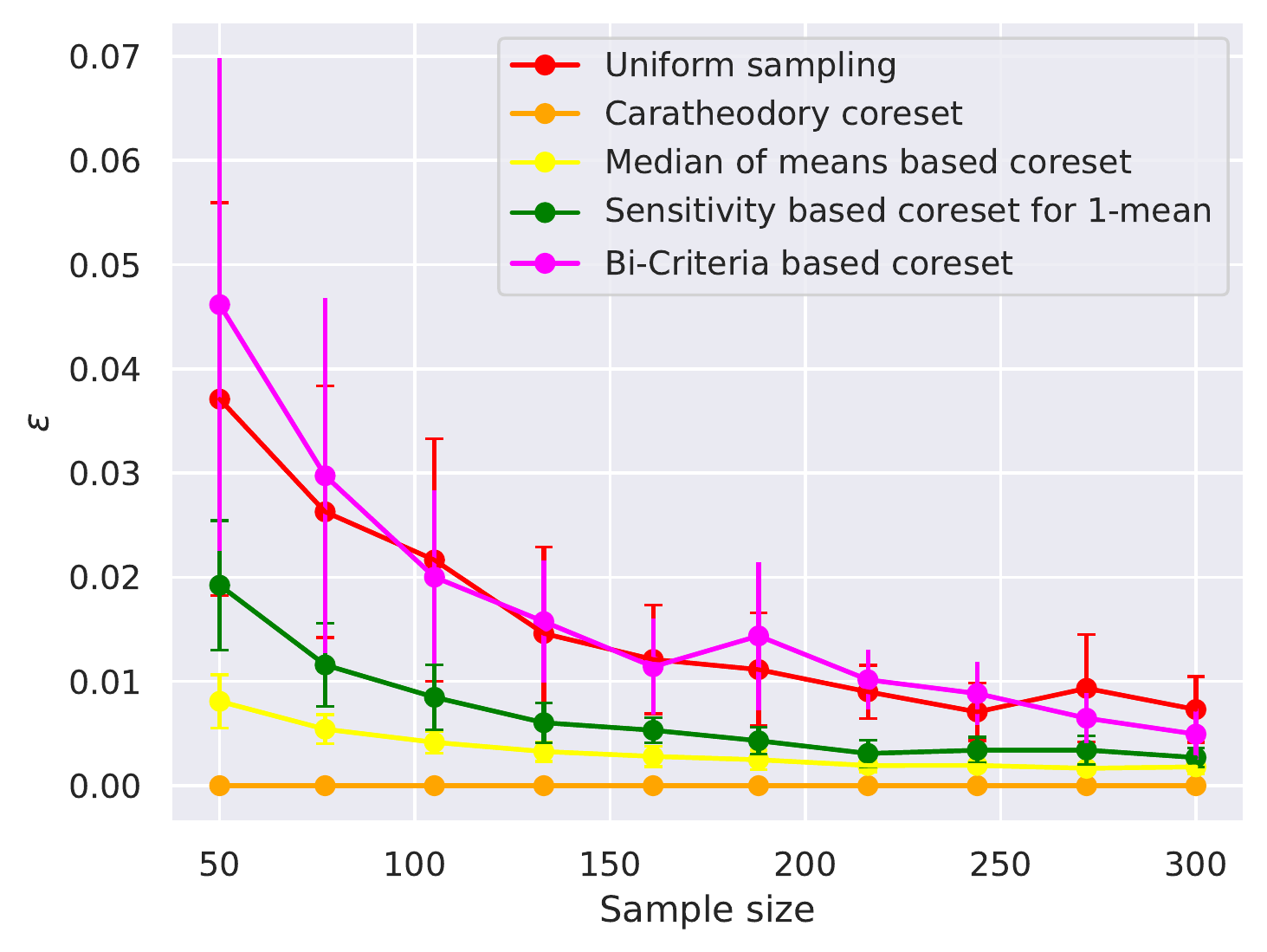}}
     \caption{Evaluation of our coresets against other competitors concerning the $k$-means problem.}
     \label{fig:Kmeans}
 \end{figure}

\textbf{SVMs.}
As for SVMs, we mainly focused on the linear kernel, while setting the regularization parameter to $1$. Similarly to logistic regression, we outperform the competing methods both in accuracy and approximation error; see Figure~\ref{fig:credit_svm} and Figure~\ref{fig:HTRU_2t_svm}.


\textbf{Discussion} 
These results show that general frameworks that aim to handle a large family of functions without embedding some crucial information concerning the properties of the problem, usually tend to lose through the race towards smaller coresets sizes with small approximation errors. We thus show that while \emph{AutoCoreset} is general in the reach of its applications, it also embeds the functional properties of the problem into higher consideration than that of~\cite{NearConvex}, and practically achieves robust results (smaller std).

\subsection{Linear regression and $k$-means clustering.}
In our experiments for linear regression, we observe a clear gap between each of our vector summarization coresets and the competing methods, leading towards outperforming the competing coreset for the task of fitting linear regression. In addition, we observe that the determination coefficient $R^2$ for our method is much closer to the determination coefficient $R^2$ when using the entire data. This indicates that our coresets lead to better learning and correlation between the input data and the corresponding outputs of the regression problem; see Figure~\ref{fig:linear_regression_problem}. In addition, for $k$-means, our coresets outperform the competitors (see Figure~\ref{fig:Kmeans}), justifying their robustness across a wide range of applications.

\section{Conclusions and Future Work}
In this work, we proposed an automatic practical coreset construction framework that requires only two parameters: the input data and the loss function. 
Our system, namely \emph{AutoCoreset}, results in small coresets with multiplicative approximation errors significantly smaller than traditional coreset constructions for various machine learning problems, as well as showing that the model learned on our coresets gained more information than the other coresets. While \emph{AutoCoreset} is practical, we also show some desirable theoretical guarantees. We believe that \emph{AutoCoreset} can be further enhanced and tuned to work in the context of Deep learning, e.g., subset selection for boosting training of deep neural networks. We leave this as future work.

Finally, we hope \emph{AutoCoreset} will lay the foundation of practical frameworks for coresets, and hope it reaches the vast scientific community, aiding to achieve faster training with provable guarantees due to training on our coresets.

\section{Acknowledgements}
This research was supported in part by the AI2050 program at Schmidt Futures (Grant G-96422-63172), the United States Air Force Research Laboratory, and the United States Air Force Artificial Intelligence Accelerator and was accomplished under Cooperative Agreement Number FA8750-19-2-1000.

\bibliographystyle{icml2023}
\bibliography{main}

\newpage
\appendix
\onecolumn
\section{More details}
\label{sec:moredetails}

\textbf{More on the initialization technique. } 
Initialization using different approximated solutions is a technique commonly used in optimization algorithms, including kmeans~\cite{arthur2007k}. The idea behind this technique is to start the optimization process from different starting points, or initializations, and to use the resulting approximate solutions to improve the overall optimization performance. This is because different initializations may result in different local optima, and by considering multiple initializations, the optimization algorithm may be able to find a better overall solution. In our method, we do not optimize the given approximate solution, but approximating several approximated solutions using our coreset practically moves the coresets towards approximating ``good’’ various regions of the query set, where each of these regions contains a good solution on the dataset.  While there is a possibility that the solutions found may be very similar, in practice, the technique tends to provide benefits in terms of improved optimization performance. Practically, we saw that uniform sampling is also sufficient to achieve very good coresets which approximate the optimal solution very well.

\textbf{More on  the stopping criteria. } First of all, the intuition behind setting stopping criteria is derived from the theory of training models in deep learning. Specifically speaking, the early stopping technique in deep learning. While we could have set the number of iterations to a hard-coded scalar (e.g., $400$), we would have either made a very weak coreset that has been exposed to not enough queries, or we would have extended the running time of the algorithm beyond the limits of being practical.  The idea that we have used in the paper is to put a threshold on the number of times the minimal cost so far has not changed thus implying some sort of convergence. Notably and most importantly, the usage of such criteria is intensively justified practically in many experimental papers (see for example,~\cite{prechelt2002early,zhou2020bert,gu2018recent}) in deep/machine learning.

We also note that the user can use any stopping criterion and of course, the results will change depending on such a choice.

\textbf{The construction of the query set. } We aimed to obtain a coreset that supports a query set that can span a meaningful part of the entire query space. Intuitively speaking, we aim to have a coreset that approximates the loss of a query set containing (i) the optimal solution of the entire data or some fine approximation to it (see next paragraph for an intuitive explanation of how this should intuitively hold) and (ii) the optimal solution on this computed coreset, given a desired problem (e.g., logistic regression). With this in mind, solving the desired problem on our generated coreset will yield a coreset approximating the solution of the entire data up to $O(\varepsilon)$.

Hence, in the $i$th iteration of our algorithm, we add the solution optimizing the current coreset to the supported set of queries (e.g., optimal logistic regression solution for the current coreset).

Since the coreset is biased towards this solution, we have evaluated the quality of such a solution on the entire data and concatenated such a vector of losses to our matrix of losses (denoted by the matrix $\tilde{\mathcal{M}}$).

This, in turn, means that each time a new query is added to the supported set of queries, the coreset in the next iteration will be adapted to approximate every query in the query set and it will become more generalized, or in a sense a “stronger coreset”.

With this in mind, we can initialize our support query set with approximated solutions to the problem (e.g. $\varepsilon$-approximations), so as to ensure a good initial coreset.


\section{Proof of Our Theoretical Results}
\subsection{Proof of Lemma~\ref{lem:VSC2FC}}
\begin{proof}
First, observe that by construction of $\subM$, it holds that for every $x \in \mathcal{X}^\prime$, and $j\in [n]$, there exists an integer $i \in [\abs{\mathcal{X}^\prime}]$ such that 

\begin{equation}
\label{eq:property_M_to_f}
\subM_{j,i} = f\term{p_j,x}.
\end{equation}

By Definition~\ref{def:VSCoreset}, the pair $\term{\C, v}$ satisfies that 
\begin{equation}
\label{eq:VSCCoresetOnM}
\norm{\sum_{j= 1}^n \subM_{j,\ast} - \sum_{\ell \in \C} v\term{\ell} \subM_{\ell,\ast}}_2^2 \leq \eps.
\end{equation}

Note that~\eqref{eq:VSCCoresetOnM} dictates that for every $k \in \left[\abs{X^\prime}\right]$, it holds that
\begin{equation}
\label{eq:VSCCoresetOnM_per_columns}
\abs{\sum_{j \in [n]} \subM_{j,k} -\sum_{\ell \in \C} v\term{\ell} \subM_{\ell,k} }^2 \leq \eps.
\end{equation}

Finally, combining~\eqref{eq:property_M_to_f} and \eqref{eq:VSCCoresetOnM_per_columns} yields Lemma~\ref{lem:VSC2FC}.
\end{proof}
 \subsection{Proof of Claim~\ref{clm:hidden_talent}}

 \begin{proof}
For every $k\in [z]$, denote by $x_k$ the query which corresponds to the $k$th column of $\subM$. 
The claim holds by the following derivations:
\begin{align*}
&\abs{\sum_{i =1}^n f\term{p_i, x} - \sum_{j \in \C} v(j) f\term{p_j, x}}^2 \\ 
 &=  \abs{\sum_{i =1}^n \sum_{k=1}^z \alpha\term{k} f(p_i,x_k) - \sum_{j \in \C} v(j) \sum_{k=1}^z \alpha\term{k} f(p_i,x_k)}^2
 \\&=
 \abs{\sum_{k=1}^z \alpha\term{k}\sum_{i =1}^n  f(p_i,x_k) - \sum_{k=1}^z \alpha\term{k}\sum_{j \in \C} v(j)  f(p_i,x_k)}^2
  \\&=
 \abs{\sum_{k=1}^z \alpha\term{k}( \sum_{i =1}^n  f(p_i,x_k) - \sum_{j \in \C} v(j)  f(p_i,x_k))}^2
 \\&\leq
  \abs{\sum_{k=1}^z \alpha\term{k} \sqrt{\eps}}^2 = \abs{\sqrt{\eps}}^2 \leq \eps ,
\end{align*}

where the first equality hold by the definition of $x$, the second and thirds are simple rearrangements, the first inequality holds by Claim~\ref{clm:hidden_talent}.
\end{proof}
  
\section{Experimental Results}
\label{sec:appendix_res}
In this section, we dive into exploring the effect of the actions/parameters used in \emph{AutoCore}.



\subsection{Taking the last coreset}
In what follows, we show the results of using the last coresets \emph{AutoCore} has devised, i.e., as Algorithm~\ref{alg:main} suggests.

\begin{figure*}[!htb]
    \subfigure[Logistic regression]{
    \includegraphics[width=0.24\textwidth]{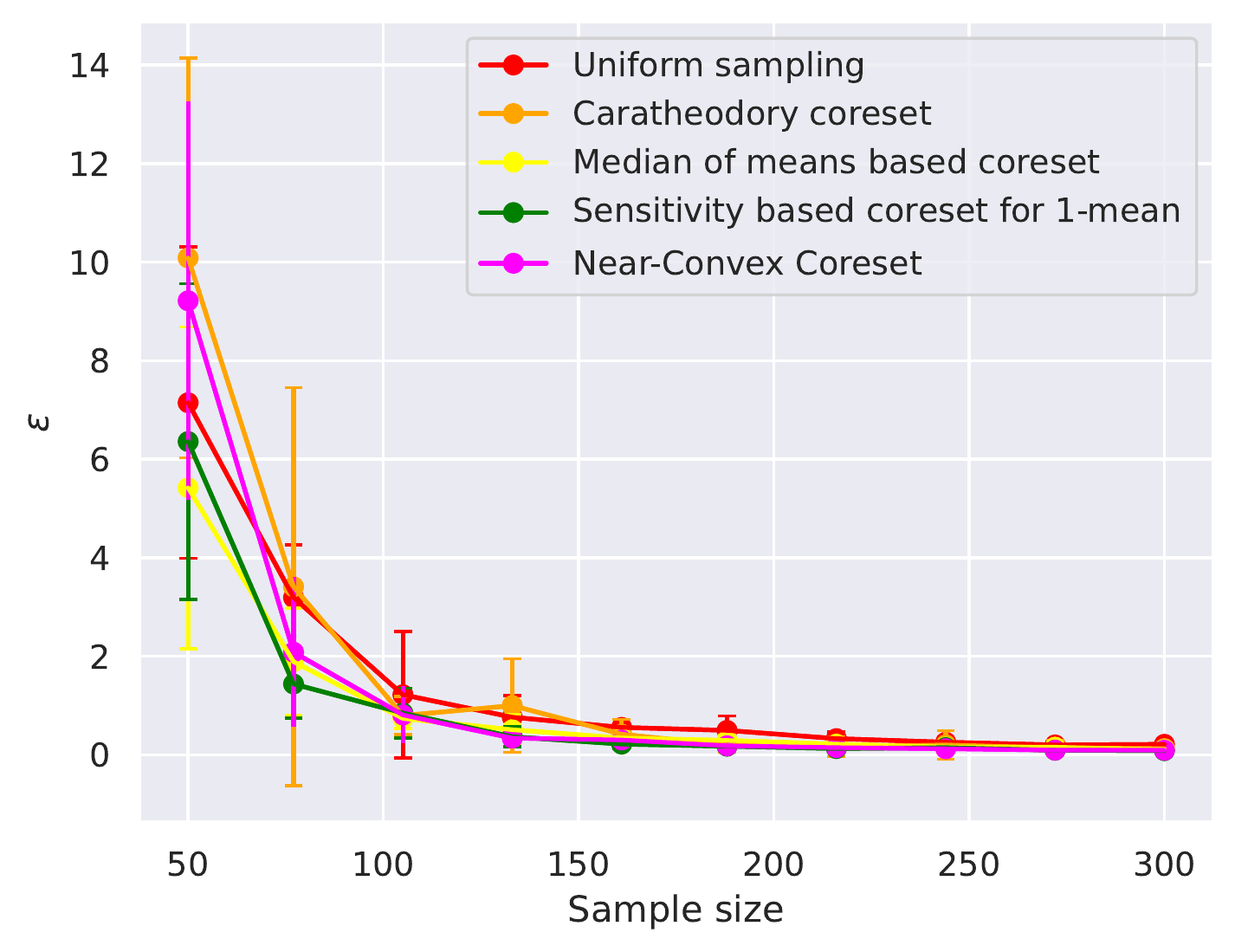}
    \includegraphics[width=0.24\textwidth]{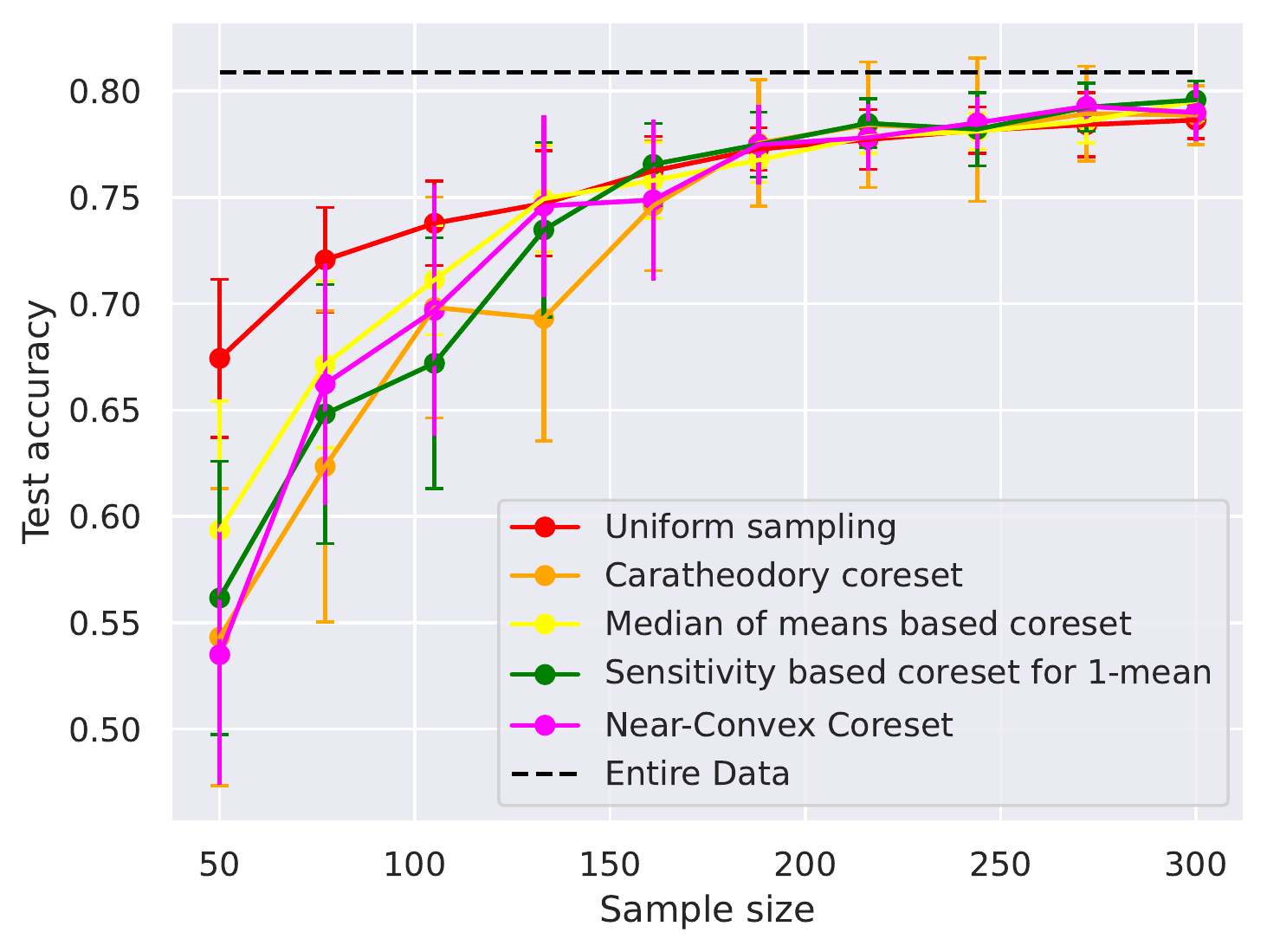}
    \label{fig:credit_logistic_regression_normal}}
    \subfigure[SVMs]{
    \includegraphics[width=0.24\textwidth]{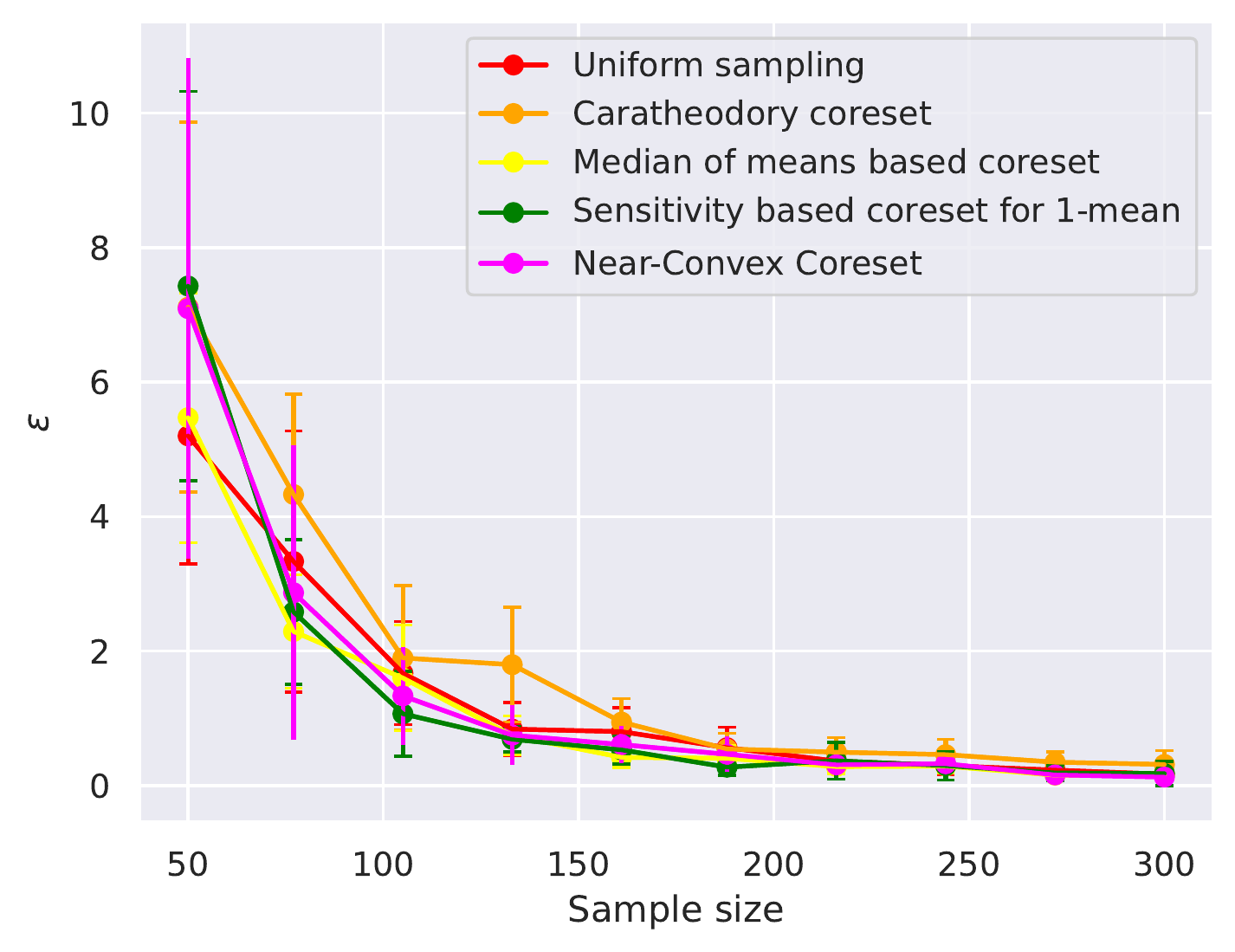}
    \includegraphics[width=0.24\textwidth]{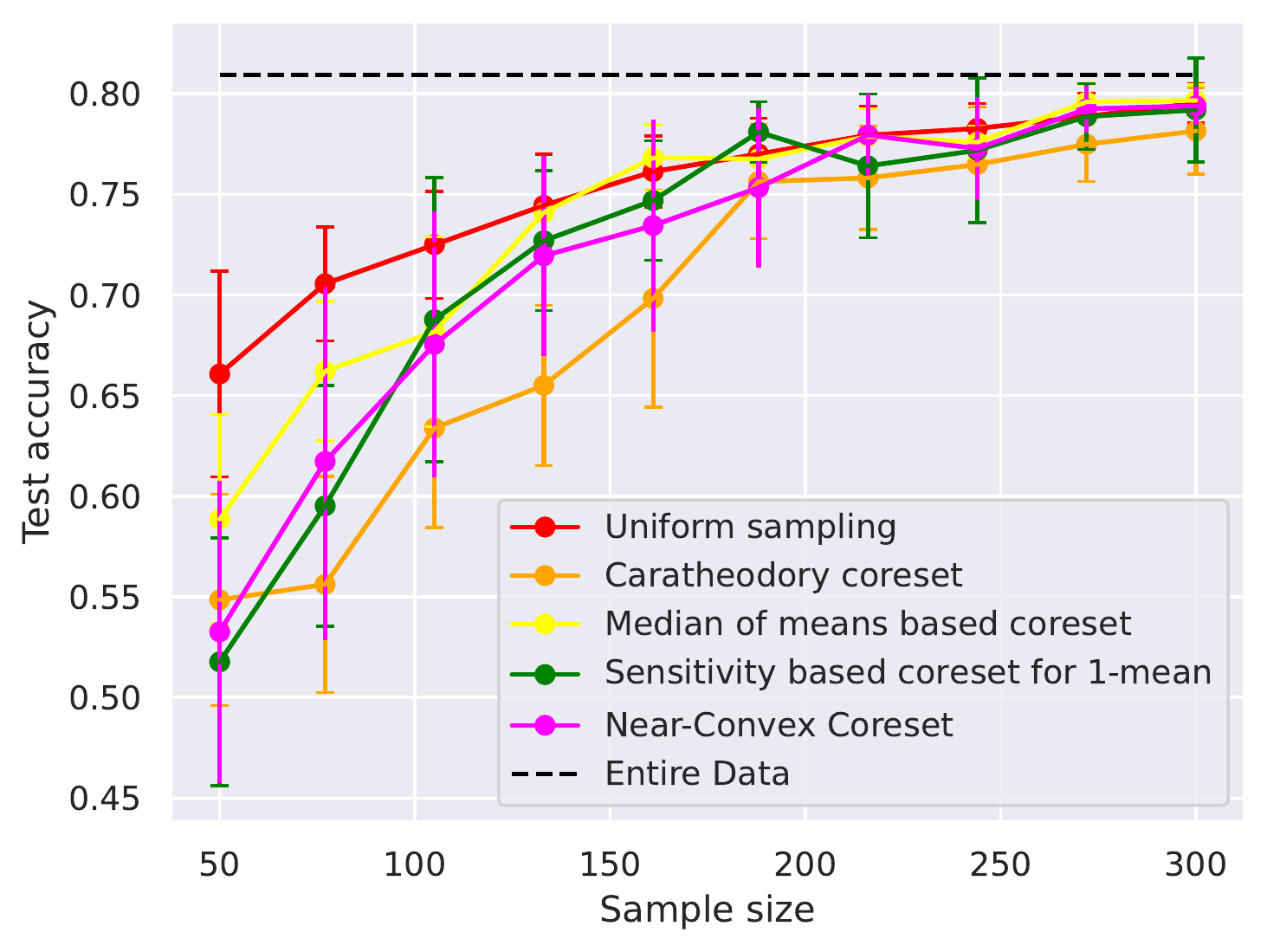}
     \label{fig:credit_svm_normal}}
     \caption{Evaluation of our coresets against Uniform sampling on the Dataset~\ref{dataset:credit}.}
     \label{fig:credit_normal}
 \end{figure*}

 \begin{figure*}[!htb]
    \subfigure[Logistic regression]{
    \includegraphics[width=0.24\textwidth]{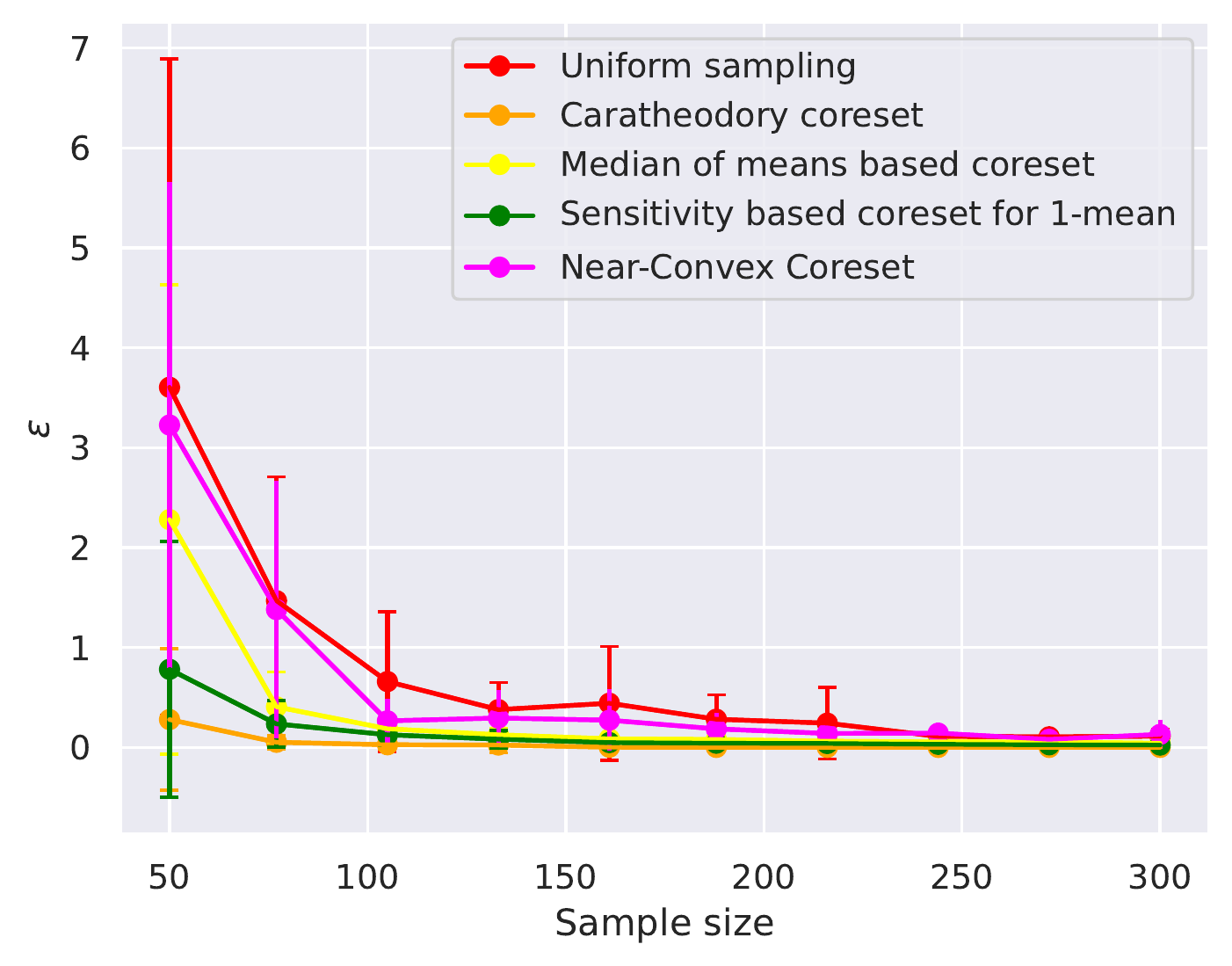}
    \includegraphics[width=0.24\textwidth]{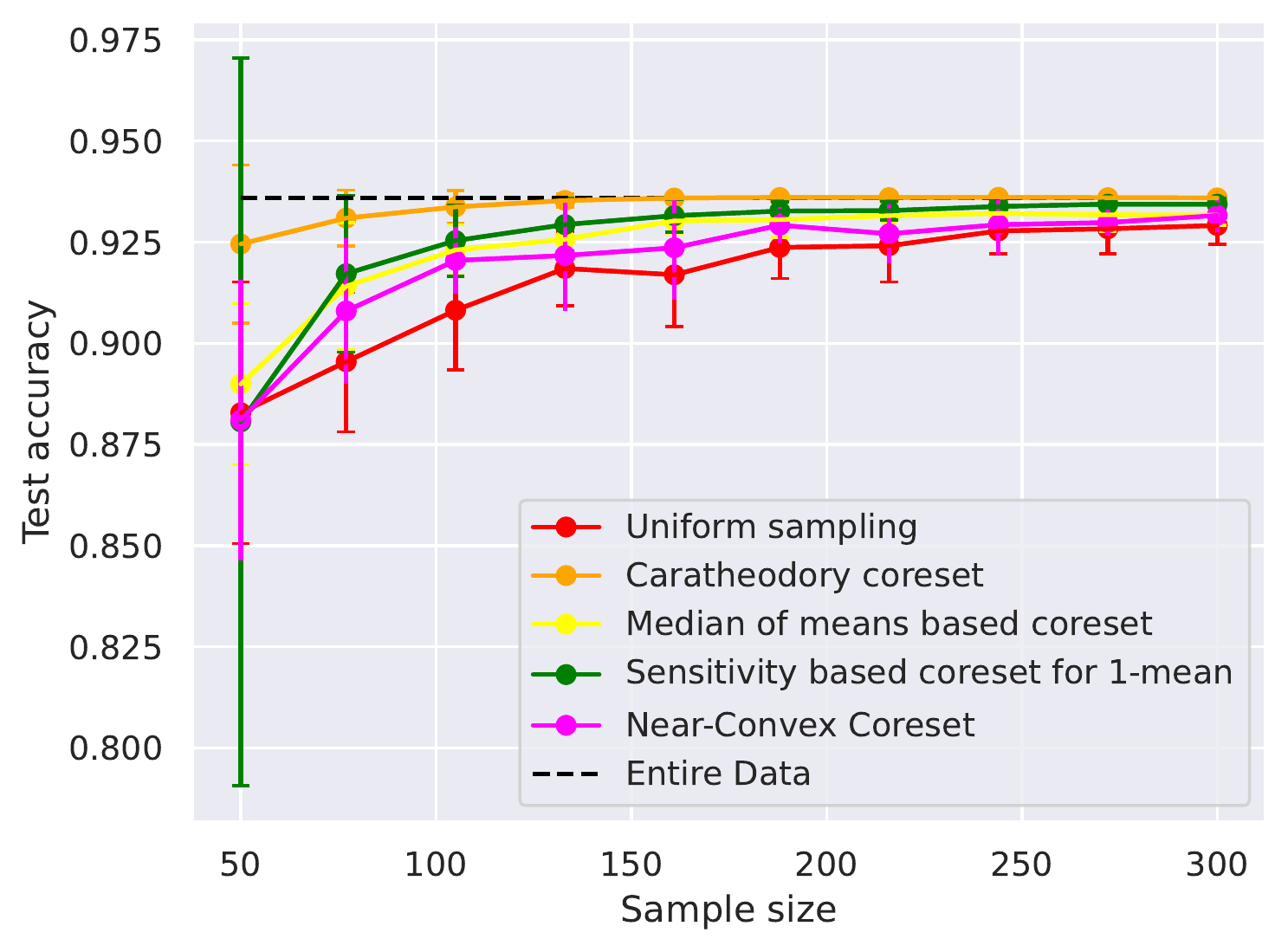}
    \label{fig:cod-rna_logistic_regression_normal}}
    \subfigure[SVMs]{
    \includegraphics[width=0.24\textwidth]{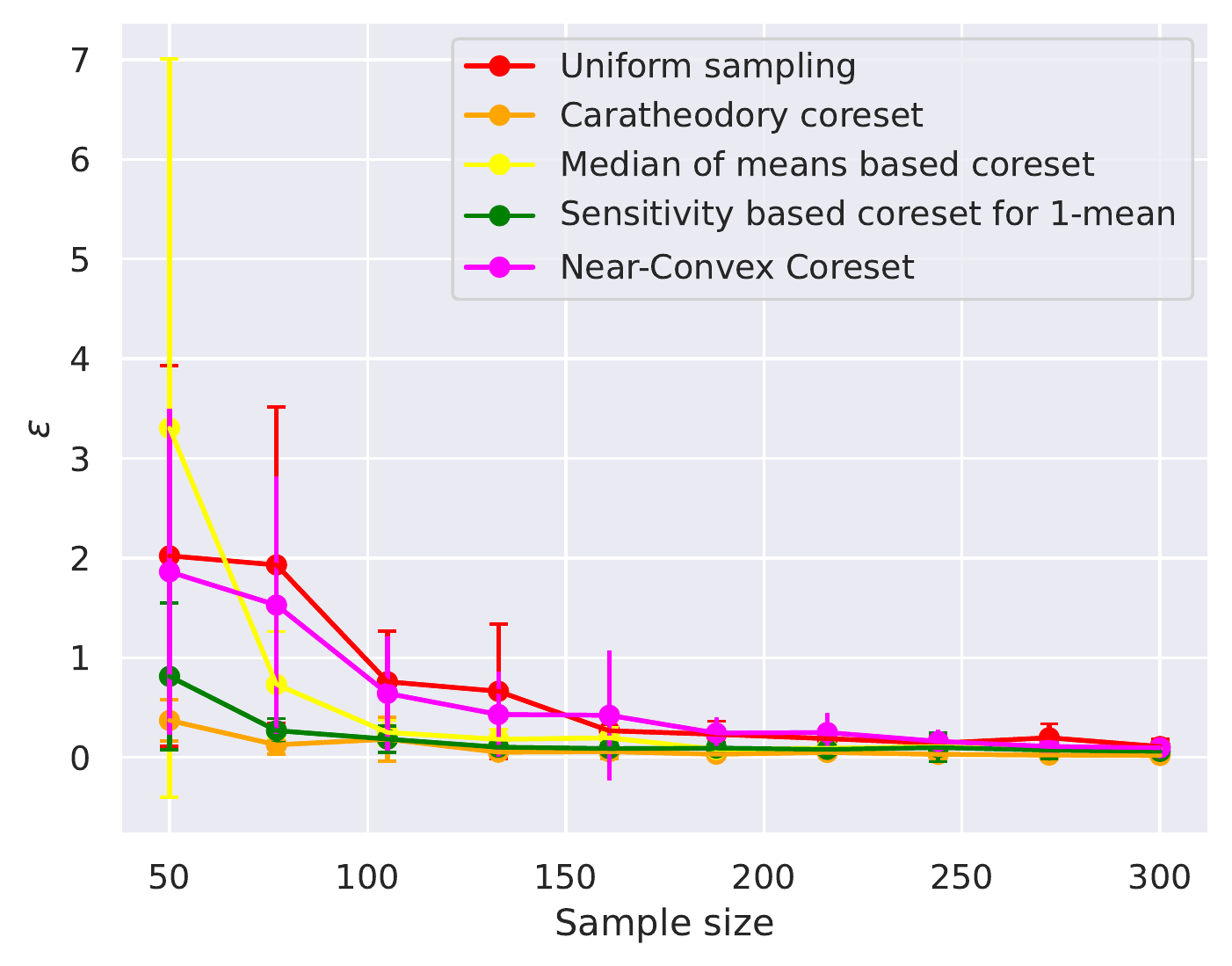}
    \includegraphics[width=0.24\textwidth]{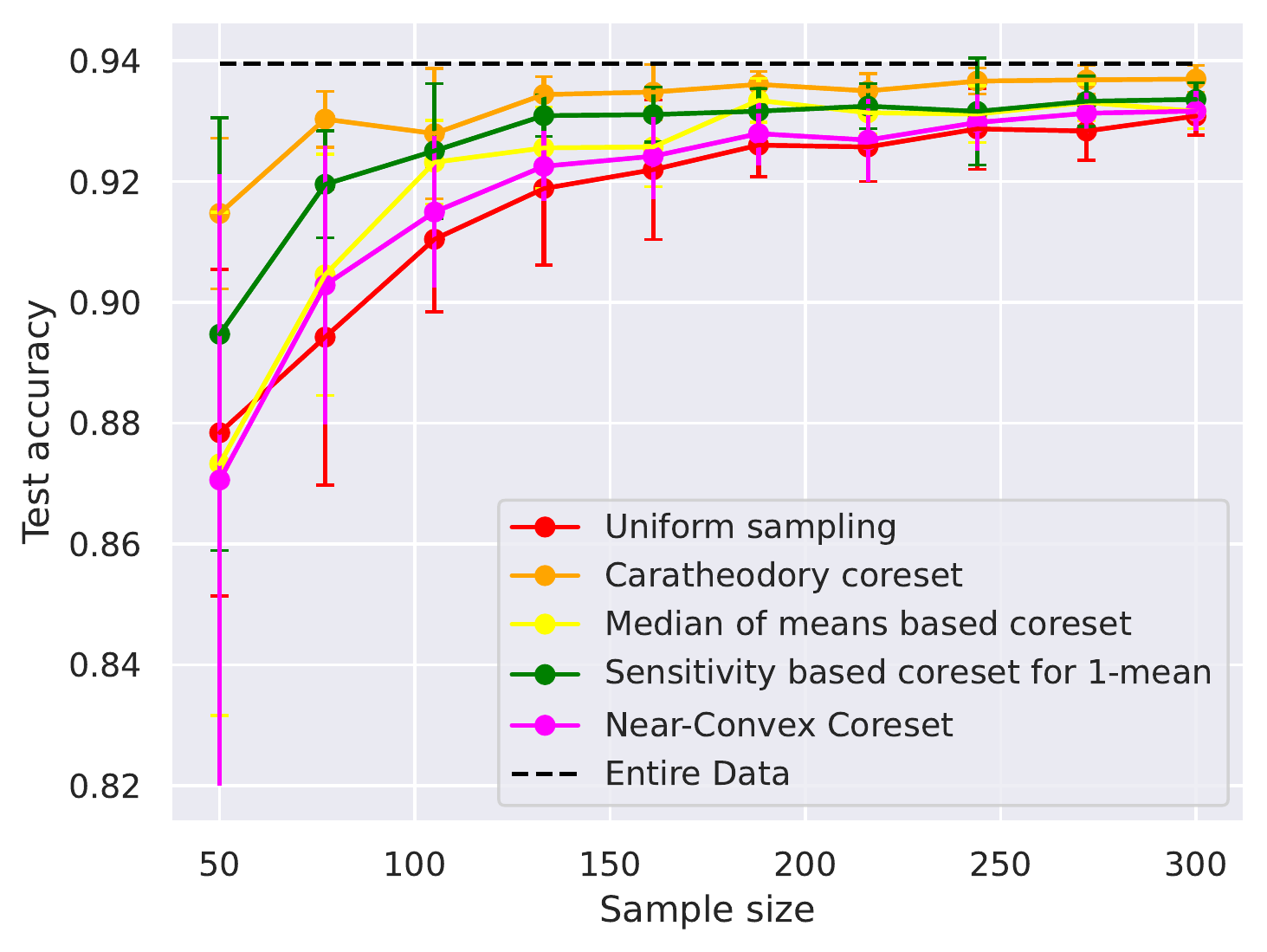}
     \label{fig:cod-rna_svm_normal}}
     \caption{Evaluation of our coresets against Uniform sampling on the Dataset~\ref{dataset:cod_rna}.}
     \label{fig:cod_normal}
 \end{figure*}

 \begin{figure*}[!htb]
    \subfigure[Logistic regression]{
    \includegraphics[width=0.24\textwidth]{figs/compete/HTRU_2_logistic_regression_vals.pdf}
    \includegraphics[width=0.24\textwidth]{figs/compete/HTRU_2_logistic_regression_acc.pdf}
    \label{fig:HTRU_2_logistic_regression_normal}}
    \subfigure[SVMs]{
    \includegraphics[width=0.24\textwidth]{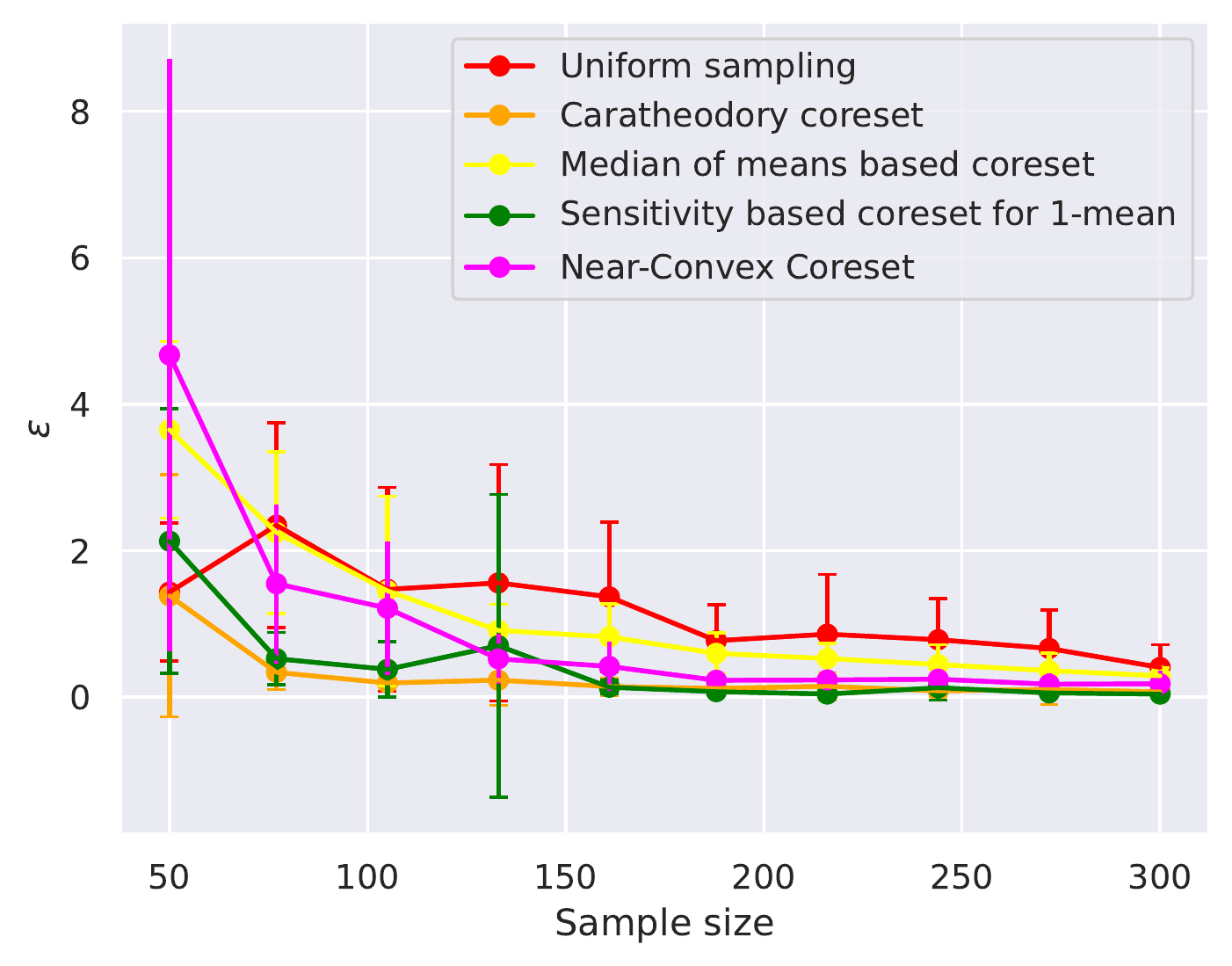}
    \includegraphics[width=0.24\textwidth]{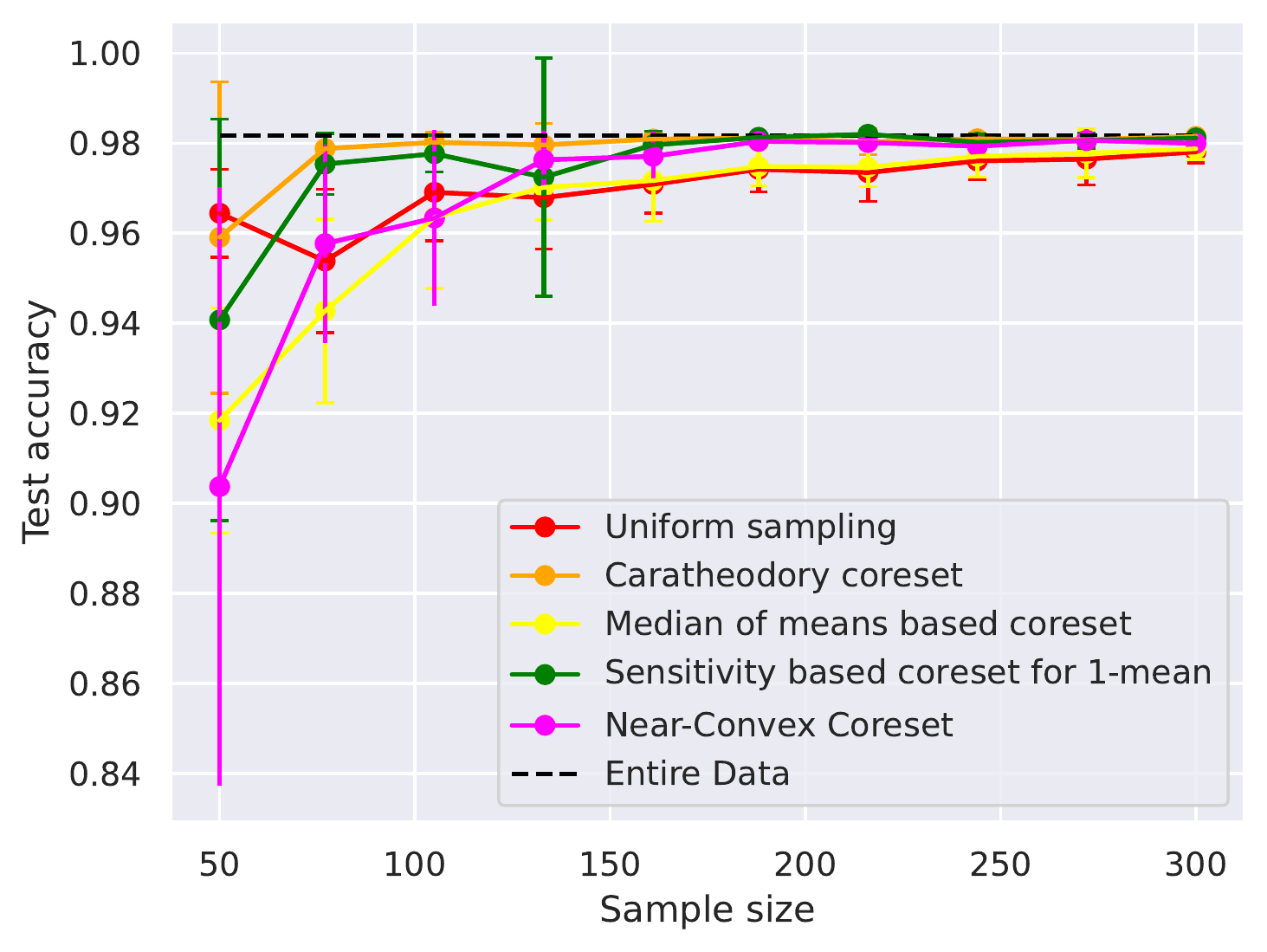}
     \label{fig:HTRU_2t_svm_normal}}
     \caption{Evaluation of our coresets against Uniform sampling on the Dataset~\ref{dataset:HTRU}.}
     \label{fig:HTRU_normal}
 \end{figure*}

 As depicted throughout Figures~\ref{fig:credit_normal}--\ref{fig:HTRU_normal}, we observe that~\emph{AutoCore} output coresets that outperform the competing methods almost in all of our experiments. In some, we observe that the desired behavior of our coreset gets delayed (takes 2 to 3 samples to outperform the rest of the competitors). This is due to the fact that taking such coresets means that the coreset is becoming more general, thus requiring a larger sample size to guarantee better approximation, one needs to sample more. Such behavior does not appear in our \say{optimal coresets} where we have taken the coreset with the optimal cost; see Figures~\ref{fig:credit}--~\ref{fig:HTRU}. The reason for this is that the optimal coreset has been exposed to fewer models/queries than the coreset that would be output by the plain \emph{AutoCore}, and thus the need for a larger sample size for smaller approximation error becomes less demanding.





\subsection{Exploration of different algorithms for choosing queries}

In what follows, we show the effect of different methods for choosing the next query for our practical coreset paradigm with respect to the logistic regression problem.

\begin{figure*}[!htb]
    \subfigure[Results when using Carath\'{e}odory as our coreset inner construction]{
    \includegraphics[width=0.33\textwidth]{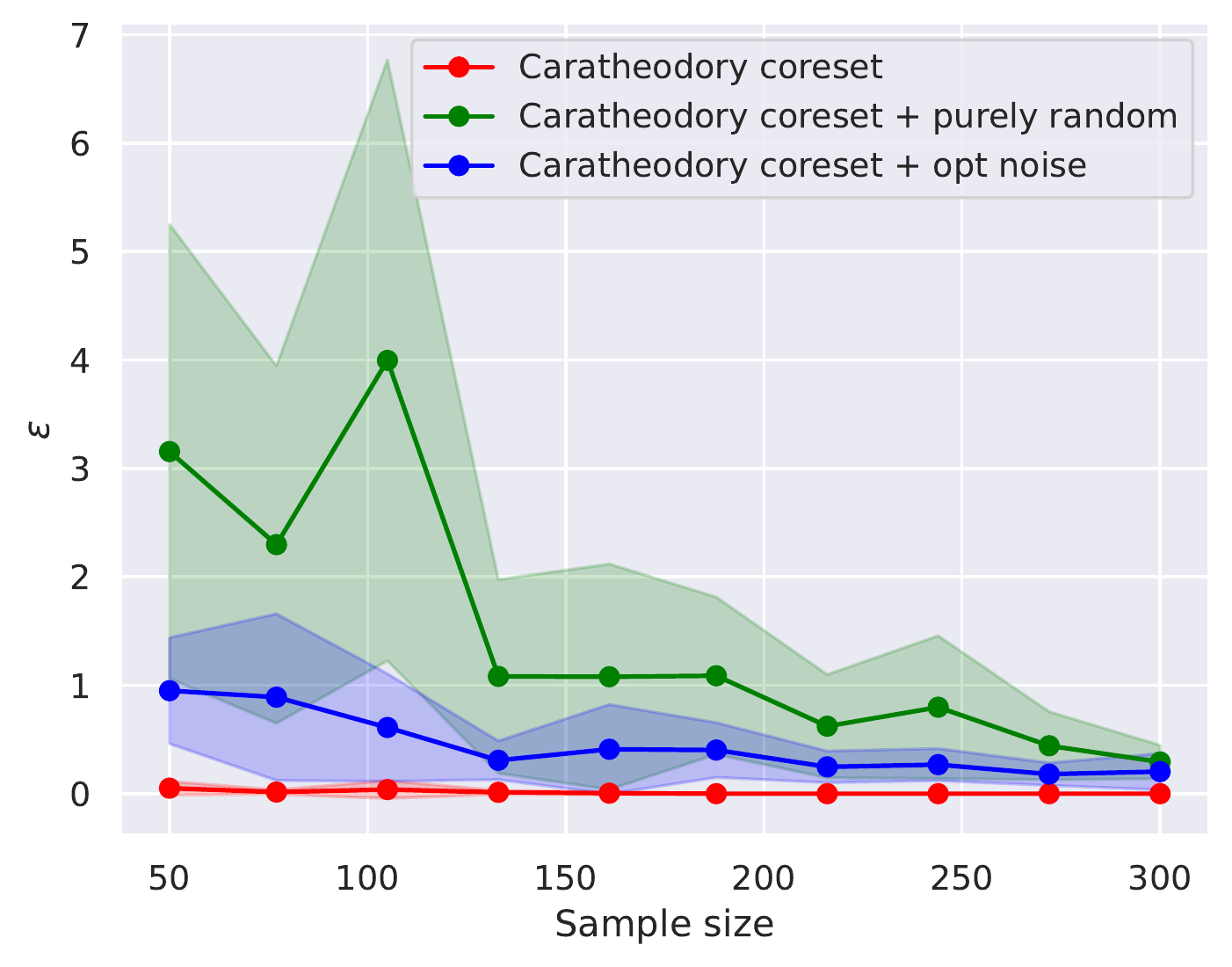}
    \includegraphics[width=0.33\textwidth]{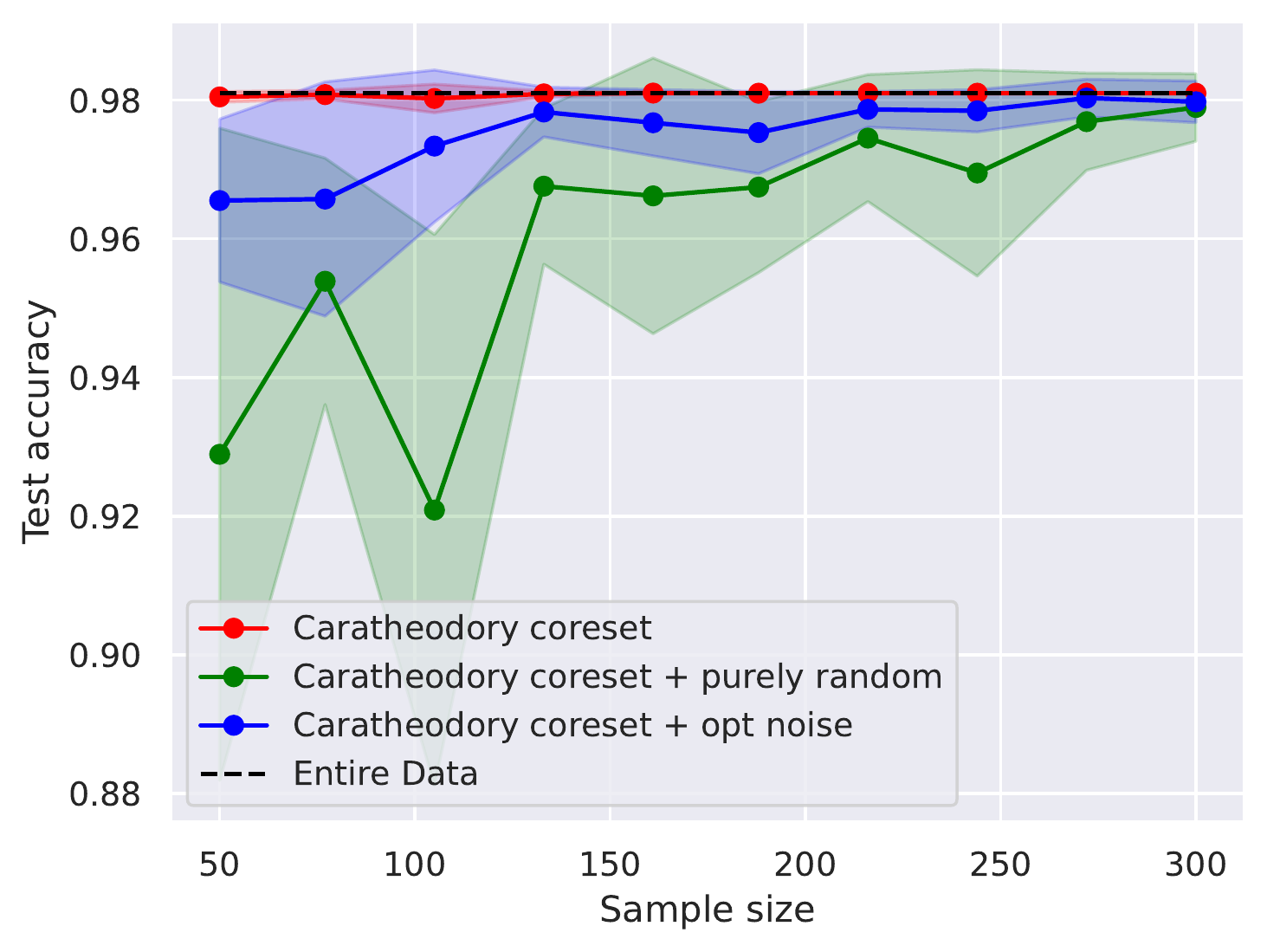}
     \includegraphics[width=0.33\textwidth]{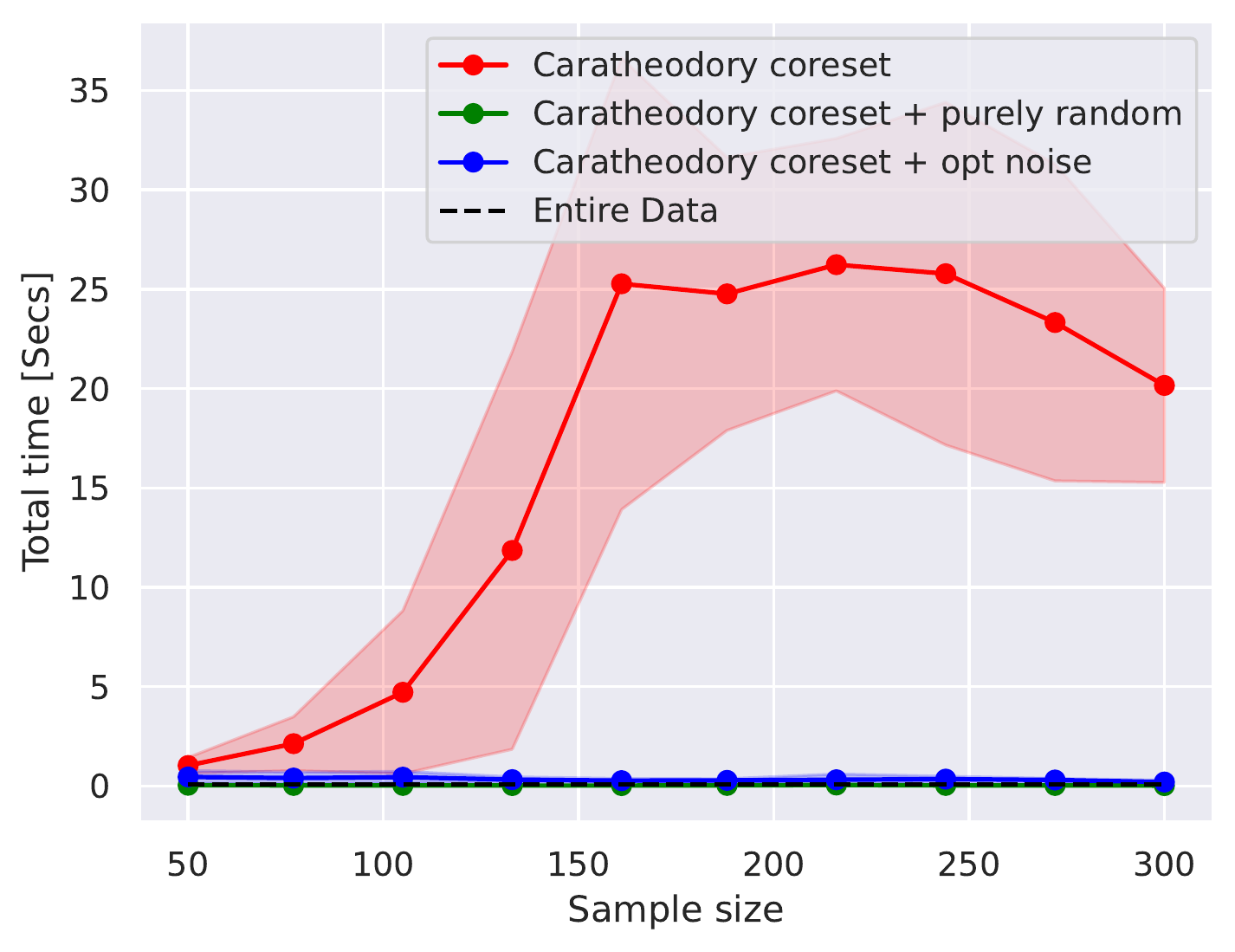}
     \label{fig:HTRU_logistic_init_caratheodoy}}
    \subfigure[Median of means]{
    \includegraphics[width=0.33\textwidth]{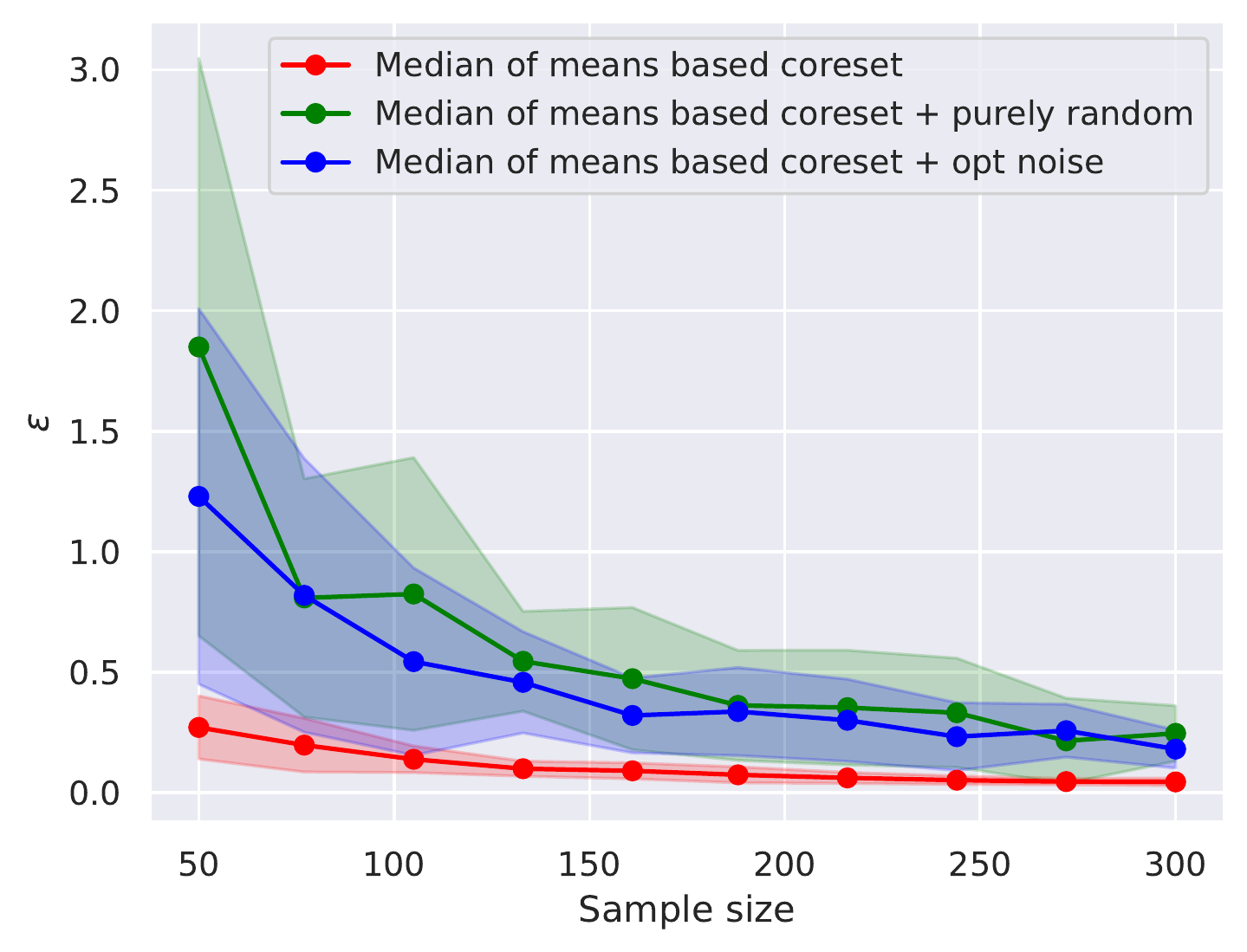}
    \includegraphics[width=0.33\textwidth]{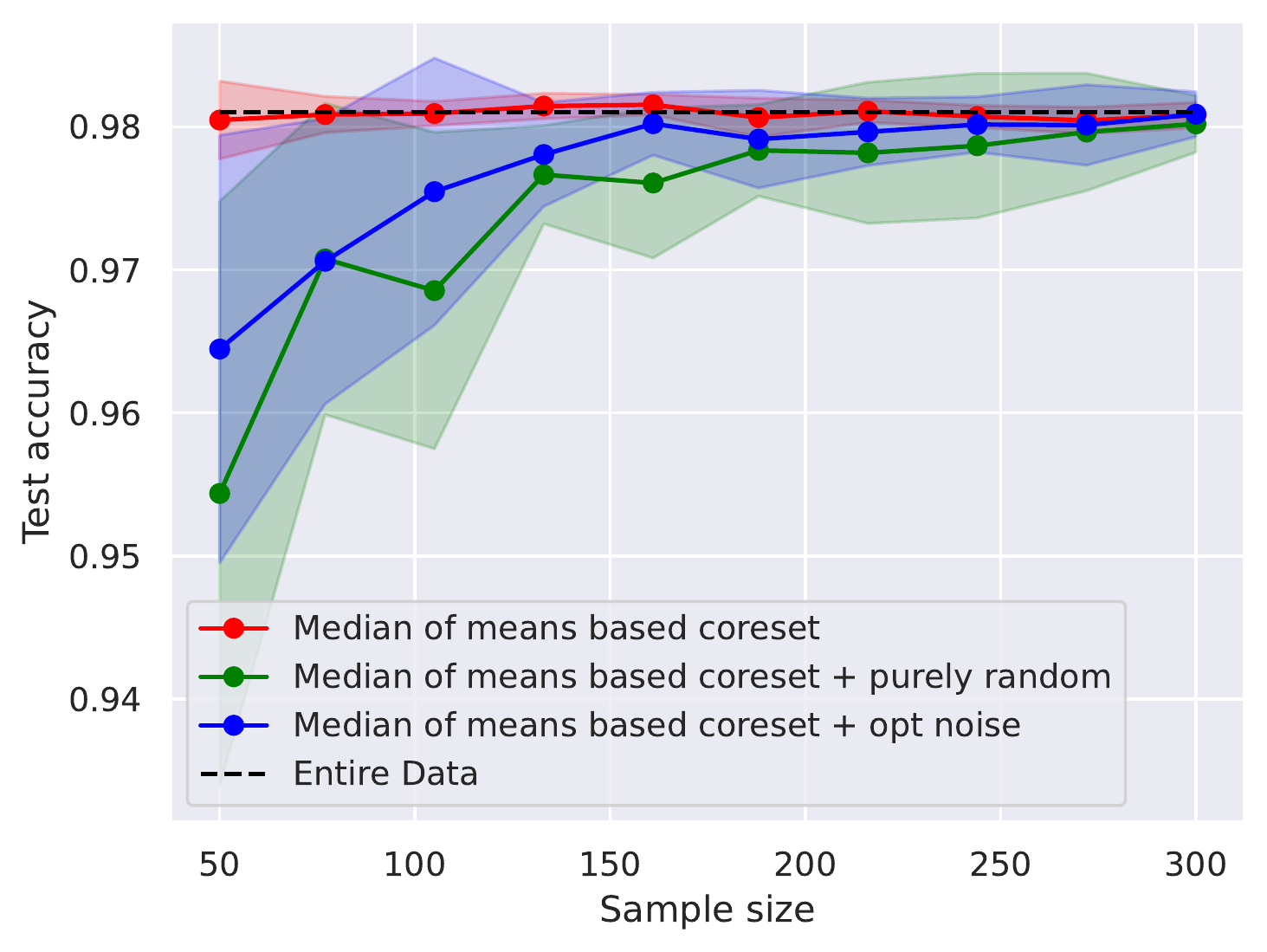}
     \includegraphics[width=0.33\textwidth]{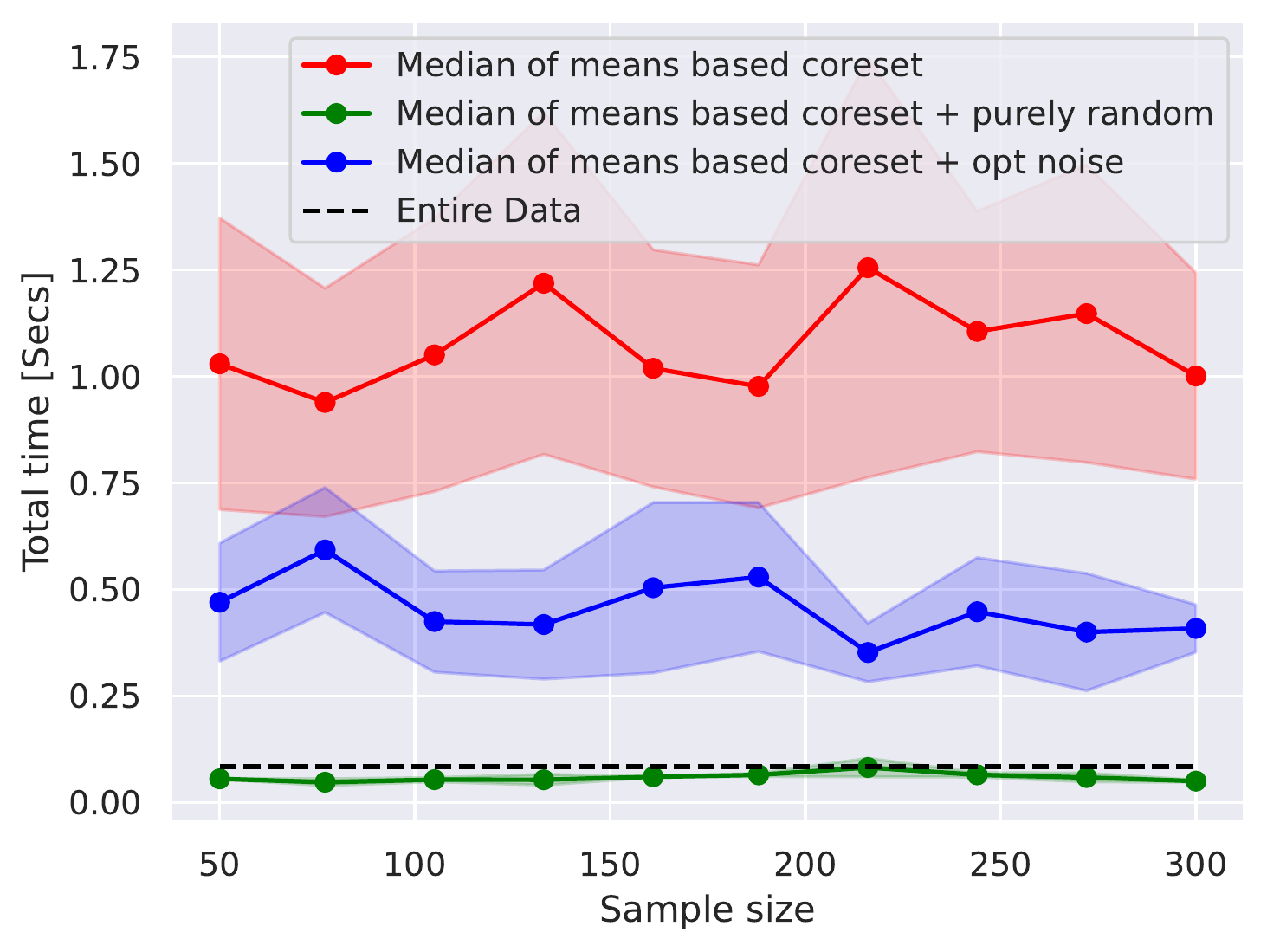}
     \label{fig:HTRU_logistic_init_median_of_means}}
       \subfigure[Results when using Sensitivity for $1$-means as our coreset inner construction]{
    \includegraphics[width=0.33\textwidth]{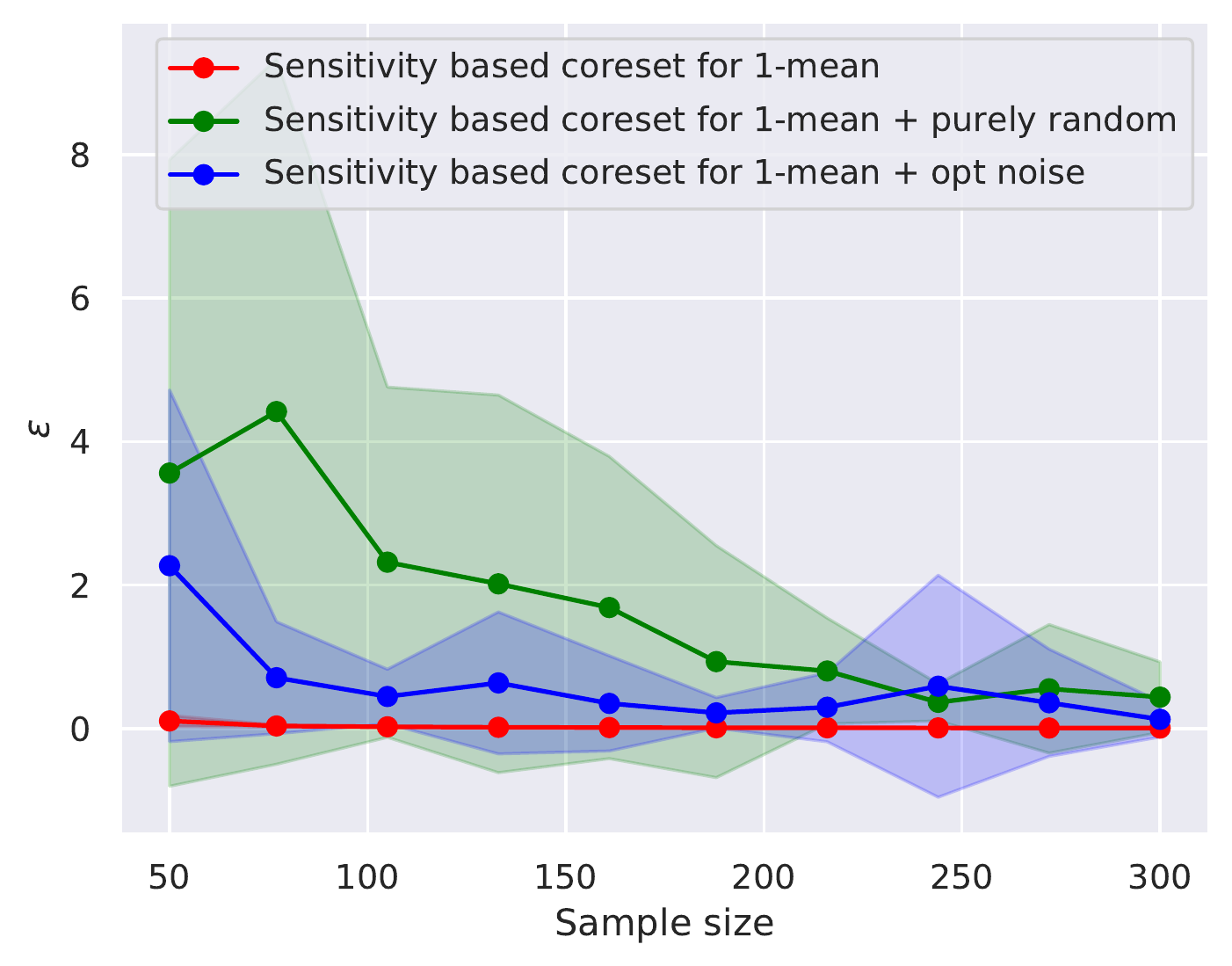}
    \includegraphics[width=0.33\textwidth]{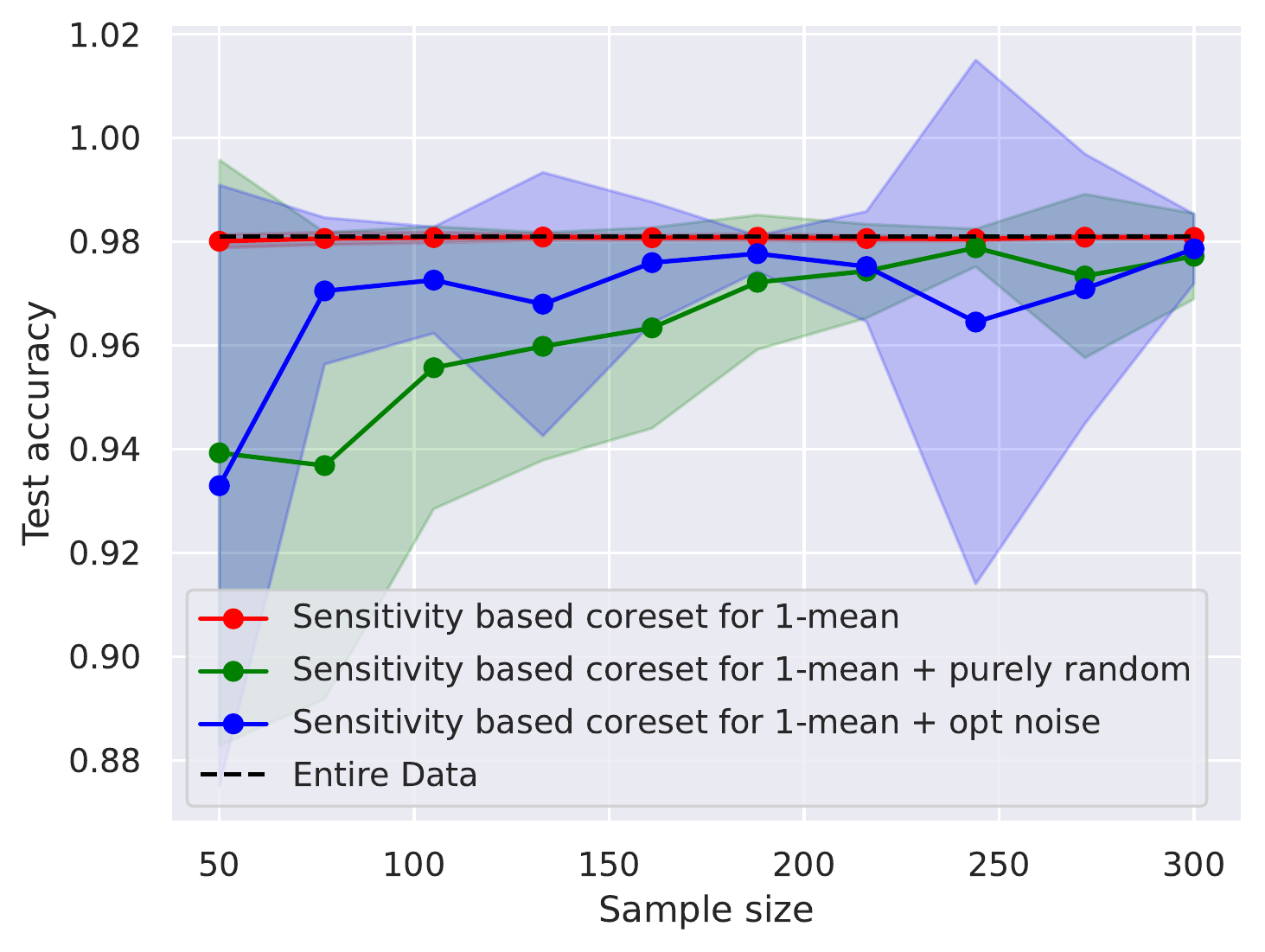}
     \includegraphics[width=0.33\textwidth]{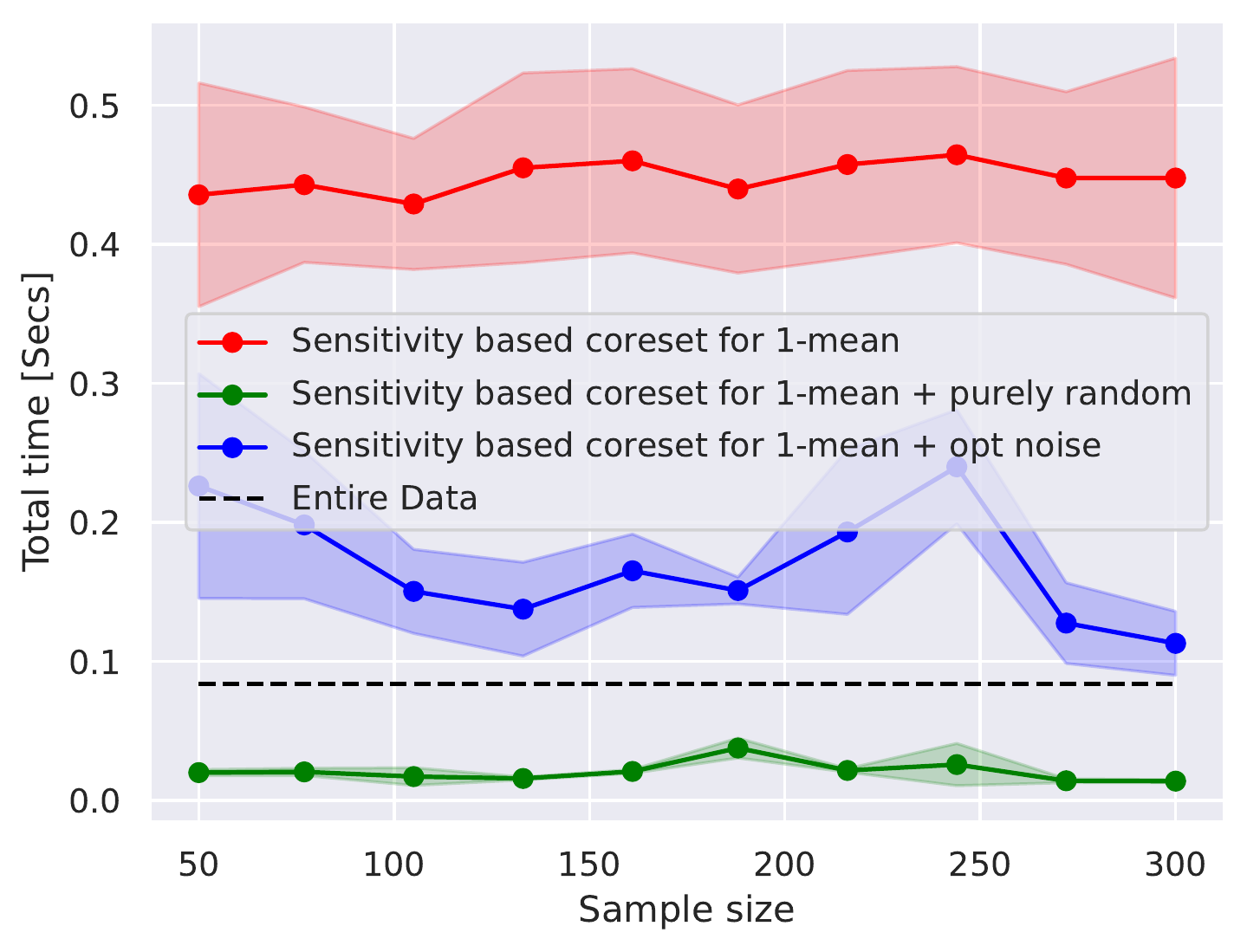}
     \label{fig:HTRU_logistic_init_median_of_means}}
     \label{fig:credit_normal}
     \caption{Evaluation of our coresets with different algorithms for choosing the next query.}
     \label{fig:differnt_choices}
 \end{figure*}

 \subsection{Experimenting with Cifar10 and TinyImageNet}

\begin{figure*}[!htb]
    \subfigure[Cifar-10]{
    \includegraphics[width=0.49\textwidth]{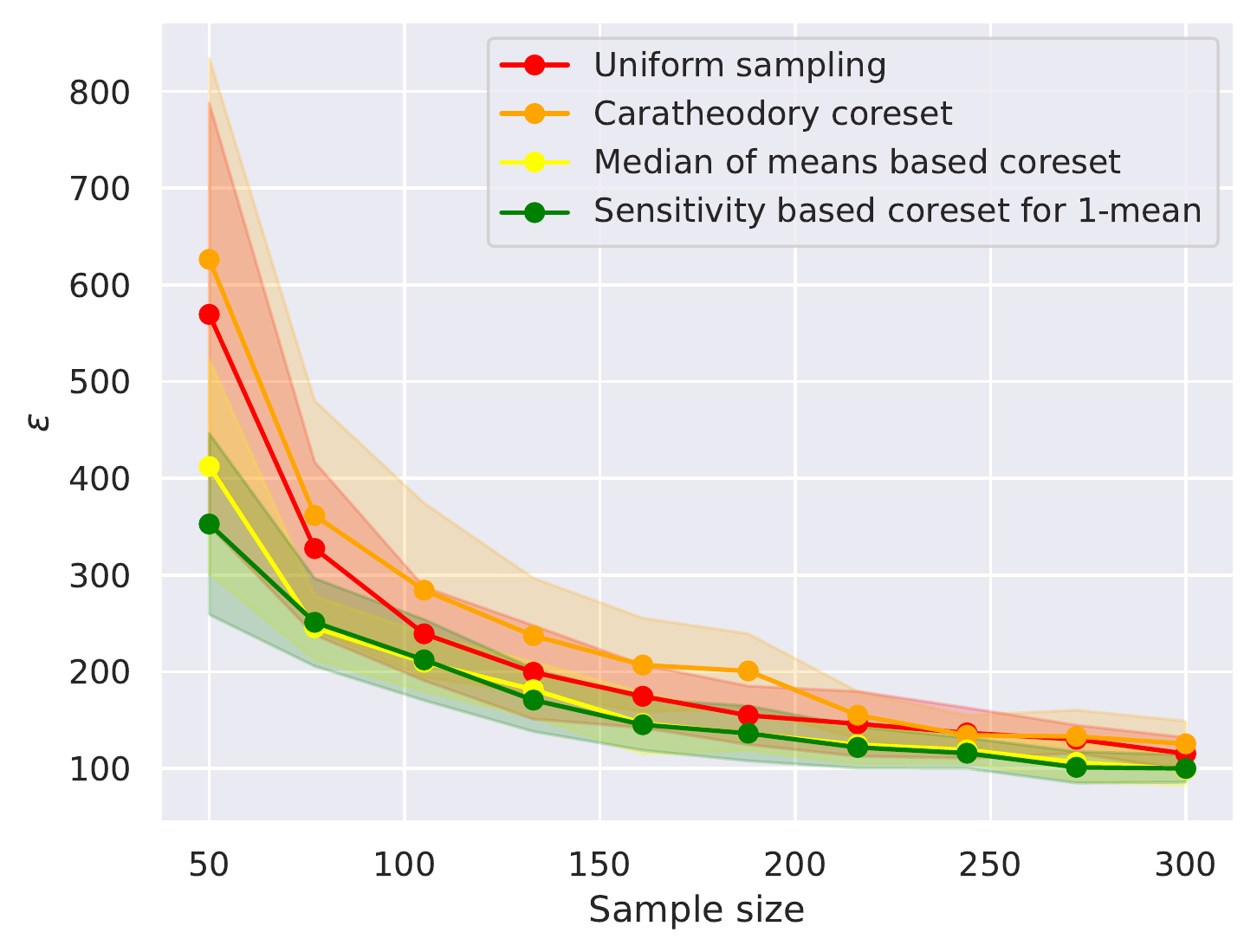}
    \includegraphics[width=0.49\textwidth]{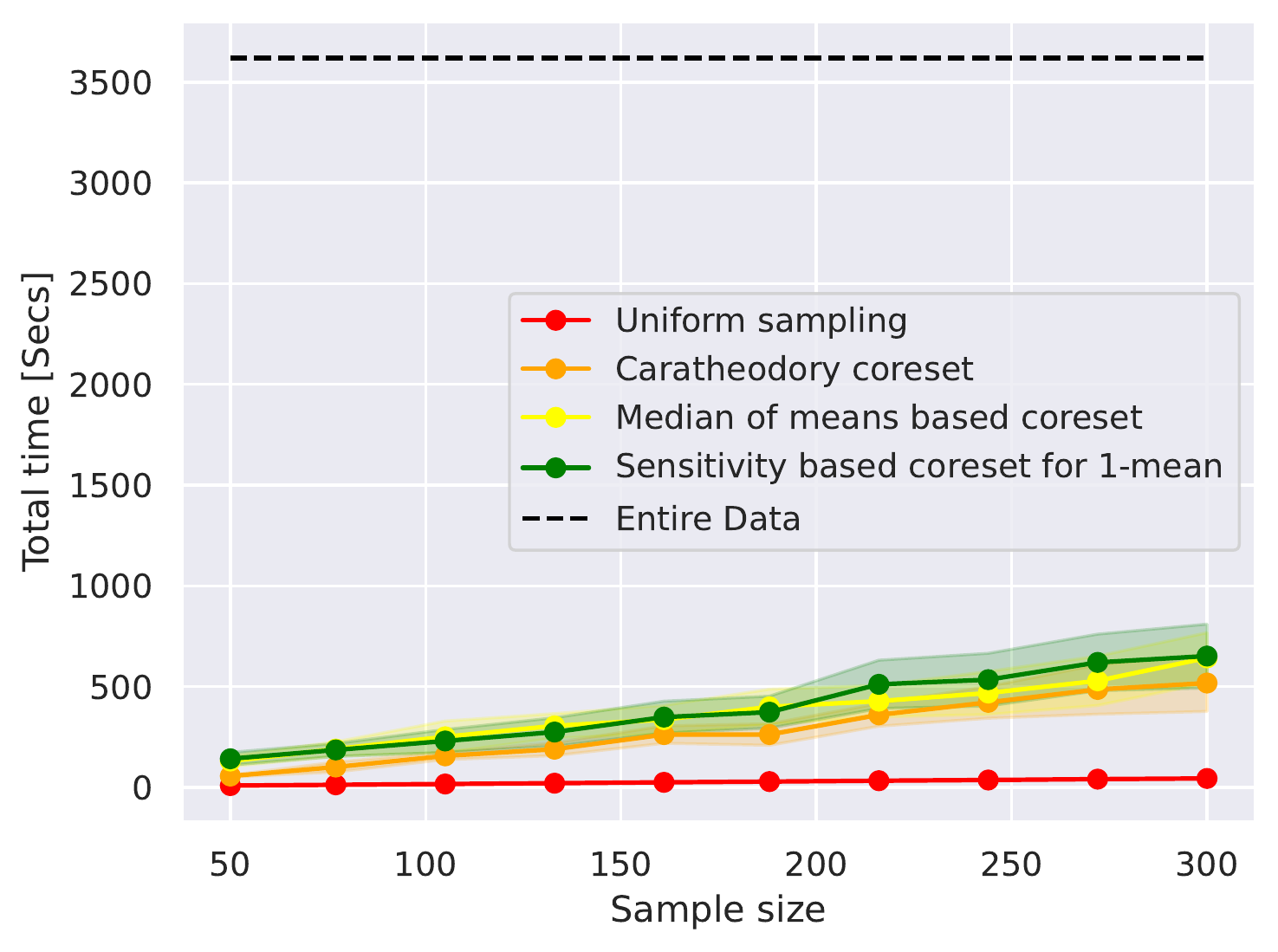}}
    \subfigure[TinyImageNet]{
    \includegraphics[width=0.49\textwidth]{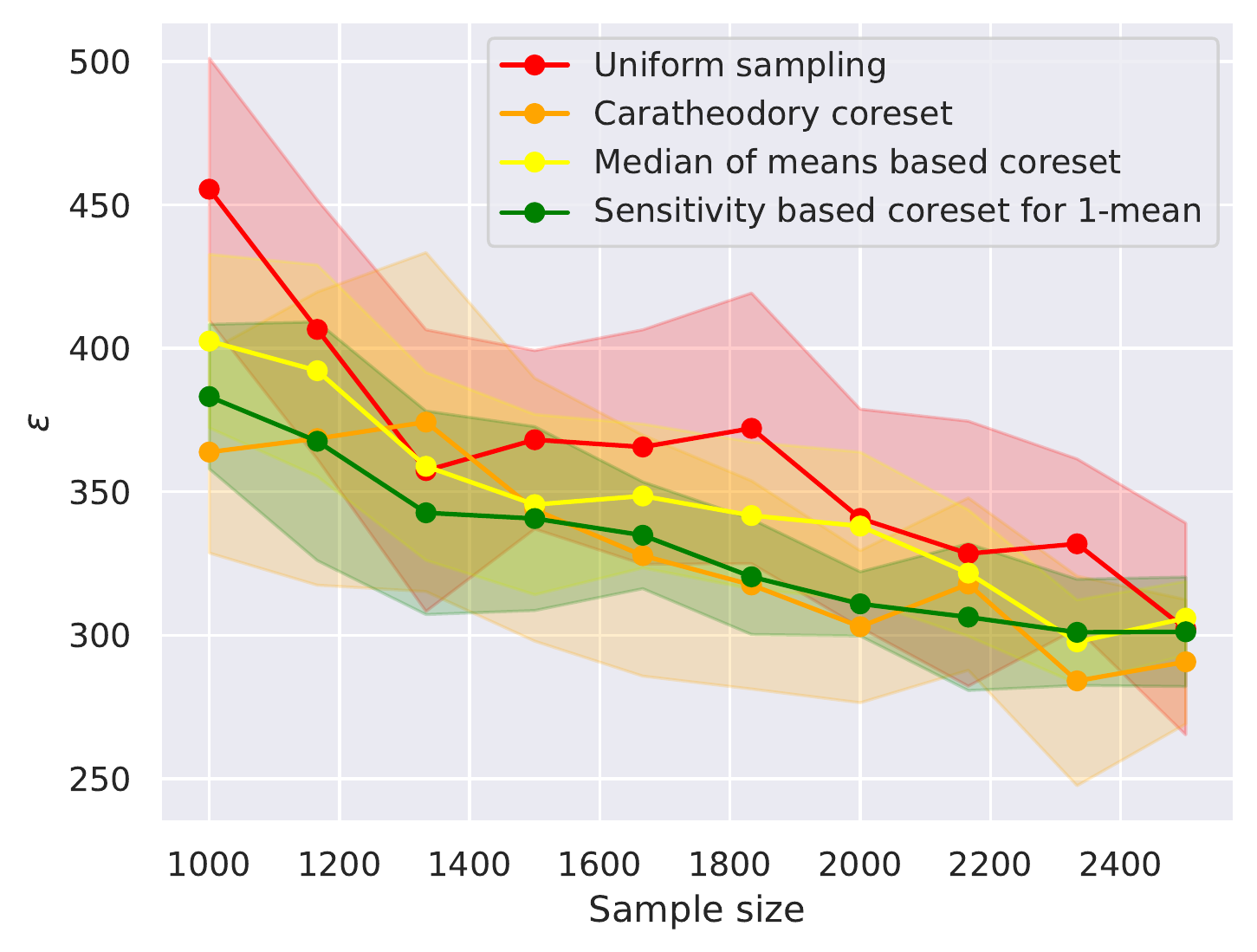}
    \includegraphics[width=0.49\textwidth]{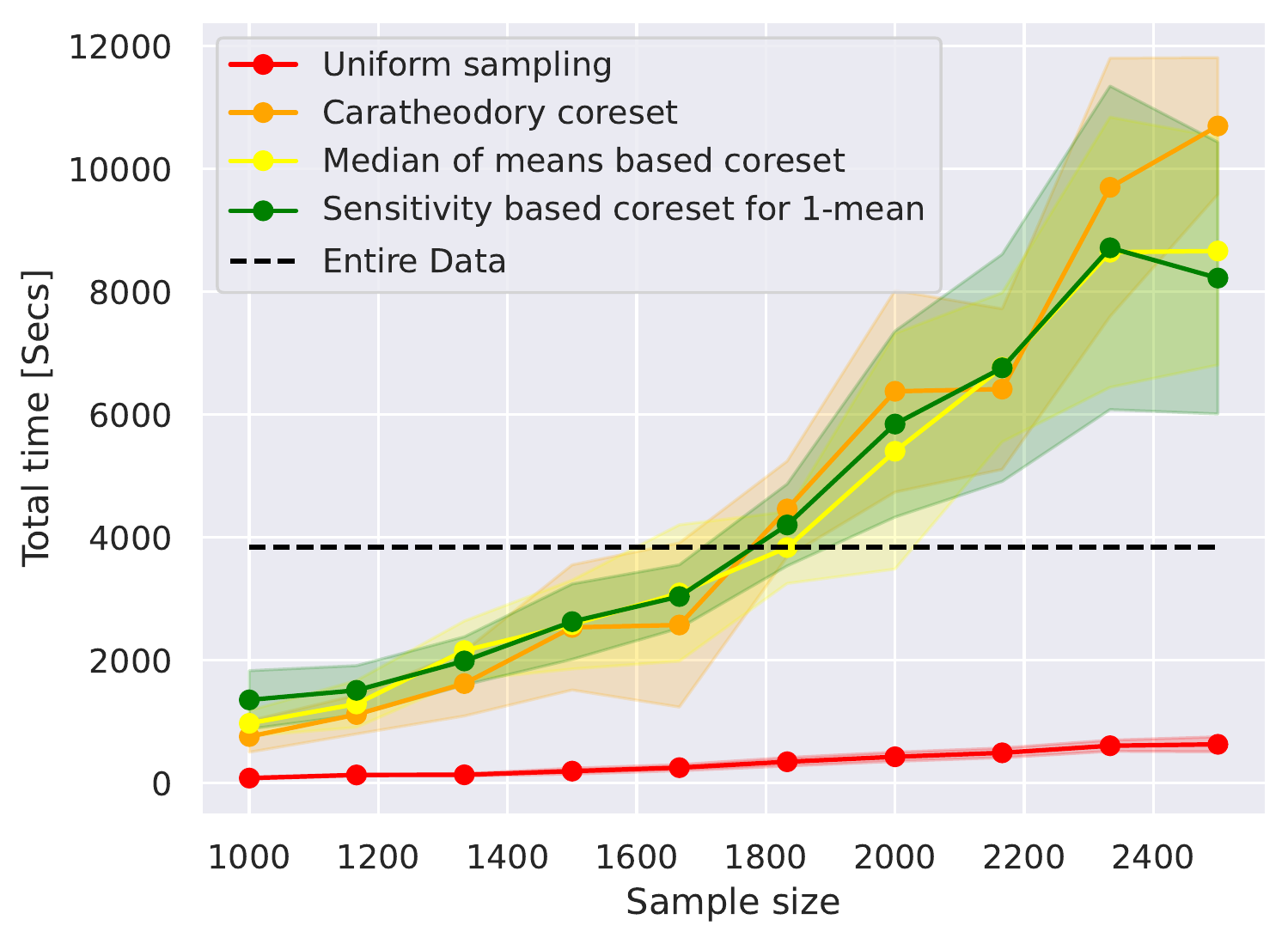}}
    \label{fig:deep_data_results}
    \caption{Evaluation of our coreset on large-scale datasets.}
\end{figure*}

 In what follows, we run our coreset paradigm on Cifar10 and TinyImageNet. For TinyImageNet data, we had to use the JL-lemma to reduce the dimensionality of the data. 
 As seen from Figure~\ref{fig:deep_data_results}, our coreset construction technique yields better coresets than uniform sampling even for large-scale datasets, where our coreset can be better than uniform sampling by at max $\approx 1.5$ times in terms of relative approximation error.

\end{document}